%% file: main.tex
\documentclass[preprint,12pt]{elsarticle}
\usepackage{comment}
\usepackage{adjustbox}
\usepackage{subfigure}
\usepackage{amsthm}
\usepackage{amsmath,amssymb,amsfonts}
\usepackage{graphicx,color}
\usepackage{textcomp}
\usepackage{xcolor}
\usepackage{hyperref}
\hypersetup{hidelinks=true}
\usepackage{algorithm,algorithmic}
\usepackage[utf8]{inputenc}
\usepackage[T1]{fontenc}
\usepackage[normalem]{ulem}
\usepackage{enumitem}
\usepackage{url}
\newcommand{\rev}[1]{{#1}}

\newcommand{\urlcolor}[1]{{\color{magenta}#1}}
\newcommand{\name}{\textit{FLex$\&$Chill}}

\journal{Advanced Engineering Informatics}
\begin{document}
\begin{frontmatter}
% \articletype{Paper} %	 e.g. Paper, Letter, Topical Review...
\title{Improving Local Training in Federated Learning via Temperature Scaling}
\author[a,b]{Kichang Lee}
\author[c]{Pei Zhang} 
\author[a]{Songkuk Kim}
\author[a]{JeongGil Ko}
\affiliation[a]{organization={School of Integrated Technology, Yonsei University}, country={Republic of Korea}}
\affiliation[b]{organization={BK21 Graduate Program in Intelligent Semiconductor Technology, Yonsei University}, country={Republic of Korea}}
\affiliation[c]{organization={Department of Electrical Engineering and Computer Science, University of Michigan}, country={USA}}

% \author{Kichang Lee$^{1,2}$\orcid{0000-0001-7276-7196}, Pei Zhang$^{3,*}$\orcid{0000-0002-8512-1615}, Songkuk Kim$^{1,*}$\orcid{0000-0003-4147-4627}, and JeongGil Ko$^{1,4,*}$\orcid{0000-0003-0799-4039}}

% \affil{$^1$School of Integrated Technology, Yonsei University, Seoul, Korea}

% \affil{$^2$BK21 Graduate Program in Intelligent Semiconductor Technology, Yonsei University, Seoul, Korea}

% \affil{$^3$Department of Electrical Engineering and Computer Science, Univeristy of Michigan, Ann Arbor, USA}

% \affil{$^4$Graduate School of Artificial Intelligence, POSTECH, Pohang, Korea}

% \affil{$^*$Corresponding Authors.}

% \email{peizhang@umich.edu, songkuk@yonsei.ac.kr, jeonggil.ko@yonsei.ac.kr}

% \keywords{Federated Learning, Temperature Scaling}

\input{0_abstract}

\end{frontmatter}
% \begin{abstract}
%Sample text inserted for demonstration. Replace with abstract text. Your abstract must give readers a brief summary of your article. Concisely describe the contents of your article and include key terms. It should be informative and accessible: indicate the general scope of the article and state the main results obtained and conclusions drawn. The abstract must be complete in itself: it must not contain undefined abbreviations and must not refer to any table, figure, reference or equation numbers. Normally the abstract text is not more than 300 words.
% \end{abstract}

\input{1_intro}
\input{2_background}

\input{3_design}
\input{4_experiment}
\input{5_discussion}
\input{6_conclusion}

%
% Each of the commands below will create an unnumbered section with the appropriate heading.
% Remove any sections that are not relevant for your article.
% All sections except suppdata will be removed if the [anonymous] option is used.
% See iopjournal-guidelines.pdf for more information.
%
\section*{Acknowledgements}
This work was supported by the Korean Ministry of Science and ICT (MSIT) and the Institute of Information \& communications Technology Planning \& Evaluation (IITP) from grants IITP-2024-2020-0-01461 (ITRC), IITP-2022-0-00420, 6G$\cdot{}$Cloud Research and Education Open Hub (IITP-2025-RS-2024-00428780), and BK21 FOUR (Fostering Outstanding Universities for Research) funded by the Ministry of Education (MOE) and National Research Foundation (NRF) of Korea.

% \data{All data used in this work is open source and can be made available upon request.}

% This section is a list of funder names and grant numbers

\bibliographystyle{plain}
\bibliography{reference}

\input{9999_appendix}
\end{document}

%% file: 0_abstract.tex
\begin{abstract}
Federated learning is inherently hampered by data heterogeneity, where non-i.i.d. training data is typical across local clients. In this paper, we explore the impact of exploiting temperature scaling during local training with respect to how it addresses the data heterogeneity issue via both mathematical and empirical analysis.
As a result, we identify that applying low-temperature values (0-1) during local training, namely \textit{Logit Chilling}, effectively expedites the model convergence and accuracy in the non-i.i.d. setting.
Considering the heterogeneous nature of federated learning, we propose \name{} that employs the \textit{Logit Chilling} and demonstrate its efficacy with extensive evaluations.
Quantitatively, from our experiments, we observe up to 6$\times$ improvement in the global federated learning model convergence time, and up to 3.37\% improvement in inference accuracy.
\end{abstract}

%% file: 1_intro.tex
%\section{Introduction}
\section{INTRODUCTION}
\label{sec:intro}
Federated learning represents a paradigm enabling the training of effective models within distributed environments without explicitly exposing local data~\cite{mcmahan2017communication}. This approach provides opportunities for communication and computation efficiency by sharing model parameters and gradients learned independently from local learning processes while suppressing raw data transmissions and potential data leakage~\cite{lee2023exploiting,lee2025mind}. Furthermore, since training operations take place in a decentralized manner, federated learning opens the possibility of parallelizing the model training process via the assistance of local client resources. As a result, federated learning presents an attractive solution for handling and exploiting the vast amounts of data generated by many mobile and Internet of Things (IoT) device entities comprising the overall system~\cite{park2023attfl,shin2024effective}.

However, in contrast to conventional centralized training methods, federated learning faces challenges associated with delayed model convergence and limited performance. More specifically, a fundamental challenge of federated learning arises from the disparity of training data used at each client device. Given this inevitable training data disparity rising from its distributed nature, federated stochastic optimization encounters difficulties in determining optimal parameters, which can lead to increased communication overhead between local devices and the central server, as well as needing additional training on (typically resource-limited) local devices~\cite{zhao2018federated}.

To address this challenge, several prior work have proposed schemes to optimize local training operations for federated learning clients or have targeted to refine the server-side model aggregation operations. For instance, FedProx~\cite{li2020federated} appends a proximal term, aiming to train models resilient to challenges posed by non-independent and non-i.i.d. data environments. SCAFFOLD~\cite{karimireddy2020scaffold} achieves expedited convergence and improved model accuracy by introducing a correction term in the local model training phase to balance the influence of each client. These operations alleviate challenges posed by the non-i.i.d. data environment, a common data distribution for federated learning clients.

%~\kc{ADD TEXT ABOUT EXISTING WORKS}
%The rationale behind this phenomenon is rooted in several reasons. One of the non-independent and identically distributed (non-i.i.d) in the local users' data which makes finding the optimal update direction in stochastic gradient descent.

%This paper proposes an orthogonally applicable, novel model training approach for federated learning, \textbf{\name{}} (\textbf{F}ederated \textbf{L}earning \textbf{EX}ploiting logit \textbf{Chill}ing), that targets to boost the convergence and model performance via temperature scaling, namely \textit{logit chilling}, during the local model training operations. 

We note that a potentially orthogonal approach to these prior efforts is to exploit the fact that a typical neural network generates a probability ($p_{i}$) for each target class from its neural network logits ($z_{i}$) by applying a softmax operation (Eq.~\ref{eq:softmax}) with a temperature $T$ value conventionally set to 1. However, it is not uncommon to exploit even higher temperature values (i.e., $T\in(1,\infty)$) in use cases such as knowledge distillation~\cite{hinton2015distilling,touvron2021training} or neural network calibration~\cite{guo2017calibration}.

\begin{equation}
%\xi(z, i, T)=p_{i} = \frac{exp(z_{i}/T)}%{\sum_{j}exp(z_{j}/T)}
p_{i} = \frac{exp(z_{i}/T)}{\sum_{j}exp(z_{j}/T)}
\label{eq:softmax}
\end{equation}

When employing a temperature $T>1$ for scaling the logits, the probability distribution across classes undergoes smoothing due to the narrowed inter-logit intervals~\cite{guo2017calibration}, making higher temperature adjustments suitable for applications such as knowledge distillation. This is possible since, despite altering the model's internal probabilities, the order of labels' probabilities remains unaffected with respect to changes in $T$. 
%By applying a temperature higher than 1 ($T\in(1,\infty)$) the probability distribution over classes is smoothed since the interval between the logits is narrowed~\kc{ADD CITATION}. Moreover, as $T$ does not change the maximum value of the probabilities, the final model prediction remains unchanged. Thus, from the perspective of information theory, applying a higher temperature provides larger entropy which is analytically identical to the amount of information while maintaining the final decision result from the model~\kc{ADD CITATION}.

While the use of diverse temperatures for neural network operations has been shown to be effective, works presented until now mainly focus on applying high temperatures \textit{after} the model is trained, making it mostly applicable for the inference operations only. For instance, since increasing $T$ reduces the model confidence, such schemes are effective in curbing the overconfidence of DNNs~\cite{guo2017calibration}.
% On the other hand, in federated learning, where disparate training data is common, \textit{increasing model confidence} of each individually trained client model (using small $T$) prior to parameter aggregation can be a more effective approach.
%
On the other hand, the effect of exploiting temperature scaling \textit{during} the training process has gathered relatively limited attention in the current literature~\cite{agarwala2020temperature}. Despite the distinctive characteristics of federated learning scenarios (e.g., data disparity and training parallelism), the benefits of employing temperature scaling during the federated training operations are left understudied.

This paper proposes a novel model training approach for federated learning, \textbf{\name{}} (\textbf{F}ederated \textbf{L}earning \textbf{EX}ploiting logit \textbf{Chill}ing), that targets to boost the model convergence and performance via temperature scaling, namely \textit{logit chilling}, during the local model training operations. This paper analyzes the efficacy and effectiveness of employing temperature scaling during the local training process in a federated learning scenario via theoretical convergence analysis and empirical evidence to provide guidelines in this unexplored territory.

% measuring both the gradient flow and data position shift within the representation space to provide guidelines in this unexplored territory.
% In this paper, we address a research question concerning the untapped benefits of employing temperatures within the range of 0 to 1 ($T \in (0,1)$) \textit{during} the federated learning process. an area of research that has received limited attention in the current literature. 

Specifically, we target to show that utilizing fractional temperature values (i.e., $T\in(0,1)$) during model training enables effective gradient propagation towards the input layer when applied to federated learning with non-\rev{i.i.d.} datasets.
% Consequently, we observed that lower temperatures facilitate more efficient data updates within the representation space when dealing with non-i.i.d. datasets, a common characteristic of federated learning systems.
From our evaluations with three datasets and three baseline federated learning models, we show that these effects accelerate the federated learning model convergence time by up to $\mathbf{6.00\times}$ and improve the inference accuracy by up to $\mathbf{3.37\%}$, despite potential concerns on model instability when training with low temperatures. Specifically, we summarize the contributions of this work in three folds as follows: 

% On the contrary, logit chilling scales the value of the logit with the temperature $T\in(0,1)$ which results in a polarized probability distribution. We revealed that logit chilling significantly boosts the speed of convergence and pushes the limit of final model performance.
% \kc{ADD TEXT}
\begin{itemize}[leftmargin=*]
    \item To the best of our knowledge, this is the first work that assesses the impact of integrating temperature scaling \textit{during} the training process in a federated learning scenario.

    \item We offer theoretical explanations and empirical evidence that \rev{demonstrate} the performance enhancements of exploiting \textit{logit chilling} in federated learning scenarios.

    \item We introduce a novel federated learning model training approach, \name{}, showcasing the application of low-temperature usage during the training process.
\end{itemize}

The remainder of this work is structured as follows. The following section will present related work in this domain to position our work among previous work. Section~\ref{sec:design} presents the design of \name{} along with deep analysis \rev{of} its operations. We will present experimental results using three datasets in Section~\ref{sec:experiment} and \rev{suggest} interesting discussion points in Section~\ref{sec:discussion}. Finally, we conclude the paper in Section~\ref{sec:conclusion}. We note that an open source implementation of the proposed scheme can be accessed at \urlcolor{\url{https://github.com/eis-lab/temperature-scaling}}.

%% file: 2_background.tex
%\section{Background and Related Work}
\section{RELATED WORK}
\label{sec:background}
%In this section, we provide background information with a discussion on relevant previous work required to better understand the rationale behind our proposed approach.
%\noindent{\textbf{Temperature scaling.}
\subsection{Temperature Scaling}
In machine learning, a model's output probability distribution serves diverse purposes. Primarily, classification models offer a probability for each class, showcasing the model's confidence in its predictions. This model-generated certainty measure allows consideration of the likelihood of each class prior to making a final classification decision. Moreover, this distribution offered by the model helps in understanding the uncertainty of a model on its decisions~\cite{park2023self,pearce2021understanding}. % For instance, even if the model assigns a specific prediction a high probability, this information allows an understanding of the prediction's reliability.

\begin{figure}[t!]
    \centering
    \includegraphics[width=.7\linewidth]{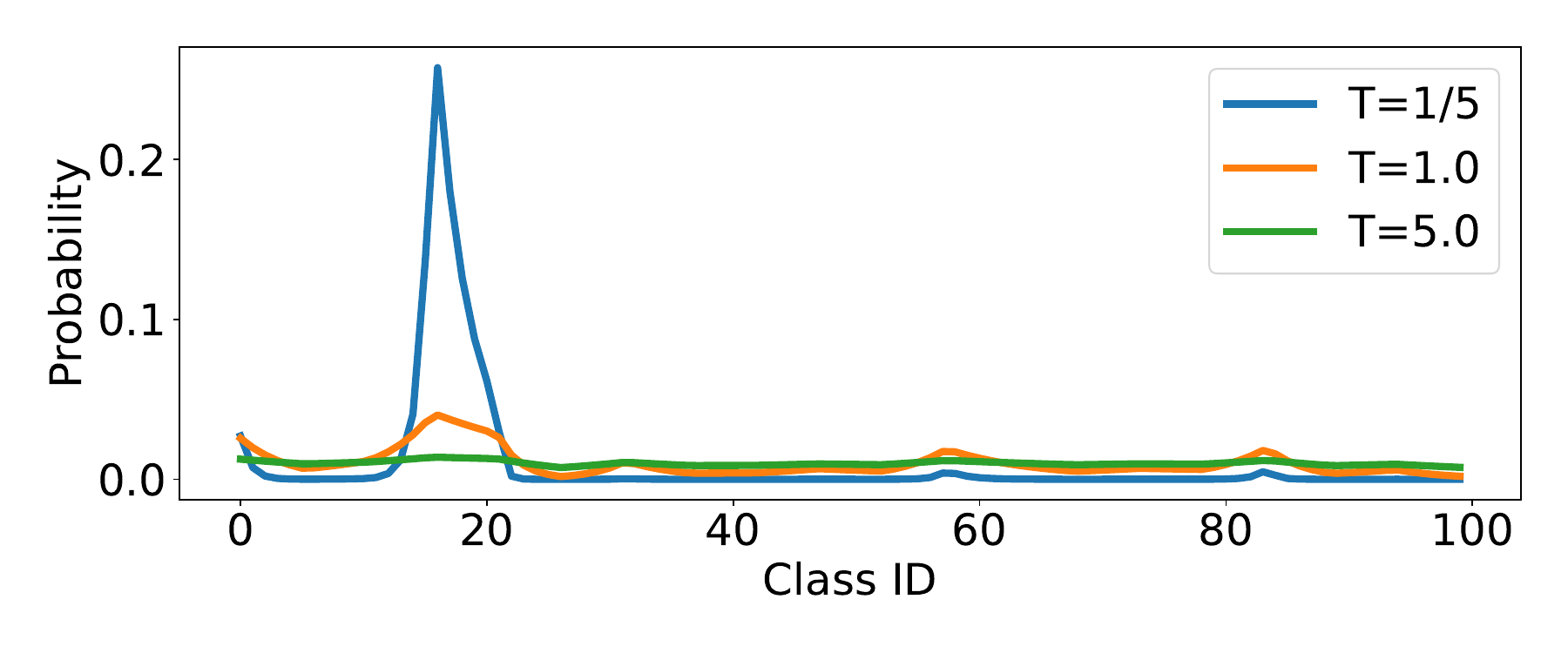}
    \caption{Effect of varying temperature $T$ on the output distribution of the softmax function, illustrating how lower $T$ sharpens class probabilities and higher $T$ produces smoother, more uniform distributions. Best viewed in color.}
    \label{fig:temperaturedemo}
\end{figure}

For neural networks, computing label probabilities from logits often involves softmax operations, which is a special case ($T=1$) of a Boltzmann distribution as in Eq.~\ref{eq:softmax}. Here, the temperature parameter ($T>0$) controls the concentration of probability towards the class with the largest logit and manages how the categorical probability is dispersed over possible classes~\cite{jang2016categorical}. Figure~\ref{fig:temperaturedemo} shows how the per-class probability distribution changes for varying temperature (i.e. $T$) values. Specifically, with $T\in(1,\infty)$, the probability distribution tends to be smoother as $T$ increases (green line in Fig.~\ref{fig:temperaturedemo}). This arises from the diminished ratio between exponentials of equally scaled values. In contrast, $T\in(0,1)$ shifts the probability distribution into a point-mass, as a lower $T$ amplifies the ratio of the exponential of scaled logits, effectively polarizing the distribution and concentrating probabilities to the maximum logit value (blue line in Fig.~\ref{fig:temperaturedemo}). Consequently, model predictions are made more confidently by allocating higher probabilities to fewer classes. Therefore, temperature scaling operates as a means to adjust the probability distribution within a neural network-based system.

Temperature scaling was applied in previous works for knowledge distillation~\cite{hinton2015distilling,touvron2021training}, generative models~\cite{shih2023long,wang2020contextual}, gradient analysis~\cite{lee2024detrigger} and model calibration~\cite{guo2017calibration}. Applying temperature adjustments to the model's output probabilities, typically through a scaling factor $T$, aids in aligning the probabilities more accurately with the true likelihoods or uncertainties present in the data.

However, most prior works have focused primarily on applying temperature scaling as a post-training technique for probability calibration, leaving its effects during training largely unexplored. While pioneering research~\cite{agarwala2020temperature} investigated temperature scaling during training and showed its positive impact in model generalization, their analysis was limited to a centralized learning scenario. Furthermore, works such as FedRS propose changing local objective functions to apply temperature scaling in federated model training operations~\cite{li2021fedrs}. However, the work is rather heuristic and does not present a thorough theoretical explanation of why altering the softmax function can improve convergence and mitigate data heterogeneity.
In contrast, our work explicitly highlights (i) the effect of temperature scaling applied during the training phase rather than post-training, and (ii) its impact specifically in federated learning scenarios characterized by heterogeneous, non-i.i.d. data distributions.

\subsection{Federated Learning}
Recently, federated learning has emerged as a promising paradigm in machine learning, particularly in scenarios where data privacy, security, and the decentralization of model training are core concerns. Unlike traditional centralized training approaches, federated learning enables model training across distributed devices or servers while keeping the data localized; thus, addressing privacy challenges associated with centralized data aggregation. This decentralized learning paradigm involves iterative model training where local updates occur on individual devices, and only aggregated model updates are shared with the central server. Specifically, federated learning aims to solve the following optimization problem:
\vspace{-1ex}
$$
\min_{w} F(w) = \sum_{k=1}^{K} p_k F_k(w),
$$
where $K$ denotes the total number of clients participating in training, $w$ represents the global model parameters to be optimized, $p_k \geq 0$ is the weight corresponding to the $k$-th client (typically proportional to the size of the local dataset), and $F_k(w)$ denotes the local objective function defined as follows:
$$
F_k(w) = \frac{1}{|D_k|}\sum_{(x,y)\in D_k} \ell(w; x, y),
$$
where $D_k$ is the local dataset on client $k$, and $\ell(w; x, y)$ represents the loss function evaluating the prediction of the model $w$ for data sample $(x,y)$. The key challenge of federated learning lies in optimizing the global objective $F(w)$ under the constraint that the clients' data remains distributed and private, each client trains locally without directly sharing its raw data.

Despite its benefits and active research, federated learning poses several challenges. A body of work has tackled the issue of communication overhead stemming from the exchange of frequent model parameter updates. A number of previously proposed works have proposed schemes to effectively abstract the model parameters in various forms to relieve the communication and networking overhead~\cite{park2023attfl}. Furthermore, federated learning systems are prone to facing issues arising from data heterogeneity caused by the diversity or differences in the characteristics, distributions, or formats of data spread across distributed federated learning clients. Previous works have attempted to address this challenge by addressing relevant issues such as variations in feature/input variations~\cite{yu2021fed2} and spatio-temporal variations~\cite{liu2020client}.

% \begin{figure}[t]
%     \centering
%     \includegraphics[width=0.95\linewidth]{grad_norm.pdf}
%     \caption{\kc{x-axis: probability, y-axis: norm of gradient}}
%     \label{fig:feasibility_1}
% \end{figure}

\vspace{-2ex}

%% file: 3_design.tex
\section{DESIGN OF FLEX\&CHILL}
\label{sec:design}
\rev{We begin discussing our framework design by examining the effect of temperature scaling (also referred to as logit scaling), providing both theoretical and empirical justification for our approach. On the theoretical side, we analyze how temperature scaling influences training dynamics in federated learning, with a particular focus on gradient amplification and its implications for convergence. Empirically, we investigate how applying temperature scaling alters the data distribution and decision boundaries within the representation space. Building on these insights, we introduce \name{}, a novel local training strategy tailored to enhance the efficiency and performance of federated learning systems.}
% We initiate discussions on our system design by analyzing the impact of temperature scaling, also noted as logit scaling, providing theoretical evidence on the rationale behind our proposed approach. 
%\jk{Our preliminary study delves into two distinct perspectives: ``gradient flow'' and ``data location shift in representation space.''} 

\subsection{Theoretical Analysis}
\label{subsec:ta}
As aforementioned, conventionally, temperature scaling has primarily been focused on calibrating the probability distribution of a model's output \textit{after} the model training phase, with limited exploration into its application \textit{during} the training process itself. In this section, we provide a theoretical analysis of the effect of temperature scaling during training with respect to the convergence in a federated learning scenario. Based on this investigation, we aim to establish a foundation for comprehending the potential implications and advantages associated with the utilization of temperature scaling techniques in neural network training.

\begin{equation}
    \frac{\partial \mathcal{L}}{\partial{z_{i}}}
    = \frac{1}{T}(p_{i}-y_{i})
    = \frac{1}{T}\left(\frac{e^{z_{i}/T}}{\sum_{j}e^{z_{j}/T}} - y_{i}\right)
    % = \frac{1}{T}(\frac{e^{z_{i}}}{\sum_{j}e^{z_{j}}} - y_{i})
\label{eq:boltzman_gradient}
\end{equation}

Eq.~\ref{eq:boltzman_gradient} presents the gradient of cross-entropy for the softmax activation ($\mathcal{L}$) with varying temperature $T$ and hard target, one-hot encoded labels. \rev{For brevity, we present the final derivation result and present detailed proof in the Appendix.}
Since the gradient is multiplied by $\frac{1}{T} > 1$, for low temperatures $T < 1$, we can expect a boost in the magnitude of the gradient. Furthermore, from the perspective of the loss term $(p_{i}-y_{i})$, when the model yields an incorrect decision, the norm of the loss term significantly increases as $p_{i}\xrightarrow{}0$.  

Building upon the theoretical analysis on the convergence of federated learning by Li et al.~\cite{Li2020On}, we can derive the inequality Eq.~\ref{eq:convergence},
where $F(w^{t})$ is the loss of the neural network with the weight ($w^t$) at round $t$, Lipschitz constant $L$, the convexity constant $\mu$, and the upper bound of variance of the stochastic gradient at local devices $\sigma^{2}$. The inequality suggests that the difference between the optimal weight $w^{*}$ and the actual weight ($w_{t}$) can be minimized by iteratively updating the weight.
\begin{equation}
\begin{split}
&\mathbb{E}[F(w^{(t+1)})-F(w^{*})] \leq \\&(1-\frac{\eta \mu}{T}) \mathbb{E}[F(w^{(t)})-F(w^{*})]+\frac{L\eta^{2} \sigma^{2}}{2T^{2}}.
\end{split}
\label{eq:convergence}
\end{equation}
Here, we can identify the role of softmax temperature ($T$) regarding the convergence of federated settings. In the inequality, the temperature $T$ controls the speed of convergence. For instance, a high temperature ($T>1$) slows the convergence as it reduces the term $\frac{\eta\mu}{T}$. Furthermore, $T$ also affects the stability of convergence as it controls the variance term ($\frac{L\eta^{2} \sigma^{2}}{2T^{2}}$). $T>1$ also reduces the amplitude of the variance term, which consequently stabilizes the convergence (Proof in Appendix).
Based on these findings, we can consider the temperature as a key hyperparameter in federated learning. We note that the above discussion can be extended to more complex adaptive optimization methods. For brevity, these details are provided in the Appendix.

Besides the mathematical analysis, to provide an intuitive rationale for applying temperature scaling during local training, the low temperatures effectively penalize incorrect predictions strongly while rewarding correct ones, enabling aggressive learning focused on challenging samples. This intuition aligns with the experimental results presented in the following section. However, it is important to note that excessively aggressive updates may cause instability.

\subsection{Temperature v.s. Learning Rate}
Since $\eta$ and $T$ have a reciprocal relationship in Eq.~\ref{eq:convergence}, we can question the relationship between the learning rate ($\eta$) and the temperature ($T$). Specifically, \textit{Will decreasing or increasing the temperature be equivalent to increasing or decreasing the learning rate?} We emphasize that while temperature and learning rate exhibit \textit{similar} effects on convergence, they are \textbf{not identical}.
\begin{align}
    w_{t+1} &= w_{t} - \eta \nabla f(w_{t})\label{eq:update1}\\
            &= w_{t} - \frac{\eta}{T}\left(\frac{e^{z_{i}/T}}{\sum_{j}e^{z_{j}/T}} - y_{i}\right)\label{eq:update2}
\end{align}
Eq.~\ref{eq:update1} presents the stochastic update of a neural network parameter $w_{t}$ at step $t$ with learning rate $\eta$.
In Eq.~\ref{eq:update2}, the update rule can be written by introducing the softmax temperature parameter $T$.
%,
% , with higher temperatures leading to softer probabilities and lower temperatures amplifying differences between logits.
Dividing the learning rate $\eta$ by the temperature $T$ suggests that increasing $T$ would have a similar effect to reducing $\eta$, and vice versa.
$T$ has, however, an additional effect in calculating the Boltzmann distribution inside the parentheses.
%However, while the mathematical form in Eq.~\ref{eq:update2} implies such a relationship, we must distinguish between the two as $T$ also scales the logits $z_{i}$ during training; thus, $T$ effectively alters the sharpness of the softmax output.

$\eta$ only controls the magnitude of the parameter update, affecting the overall convergence dynamics. On the other hand, $T$ has two-fold effects: changing convergence dynamics and modulating the output distribution's sensitivity. $T$ impacts the relative importance of individual gradients and their direction rather than only uniformly scaling the update size. Therefore, although $\eta$ and $T$ appear to play interchangeable roles for some cases, the underlying mechanisms through which they influence the training process are fundamentally different. 

\subsection{Empirical Analysis}
\label{subsec:ea}
In this section, we present empirical results via three initial experiments specifically crafted to support the analytical results and showcase the implications of employing temperature scaling throughout the training phase operations in different dimensions. Through these investigations, we aim to establish a foundation for comprehending the potential implications and advantages associated with the utilization of temperature scaling techniques in neural network training. We conduct our experiments with the CIFAR10 dataset~\cite{krizhevsky2009learning} and use a 2-layered CNN architecture~\cite{mcmahan2017communication} as the baseline neural network architecture.

\vspace{1ex}
\begin{figure}[t!]
    \centering
    \includegraphics[width=.7\linewidth]{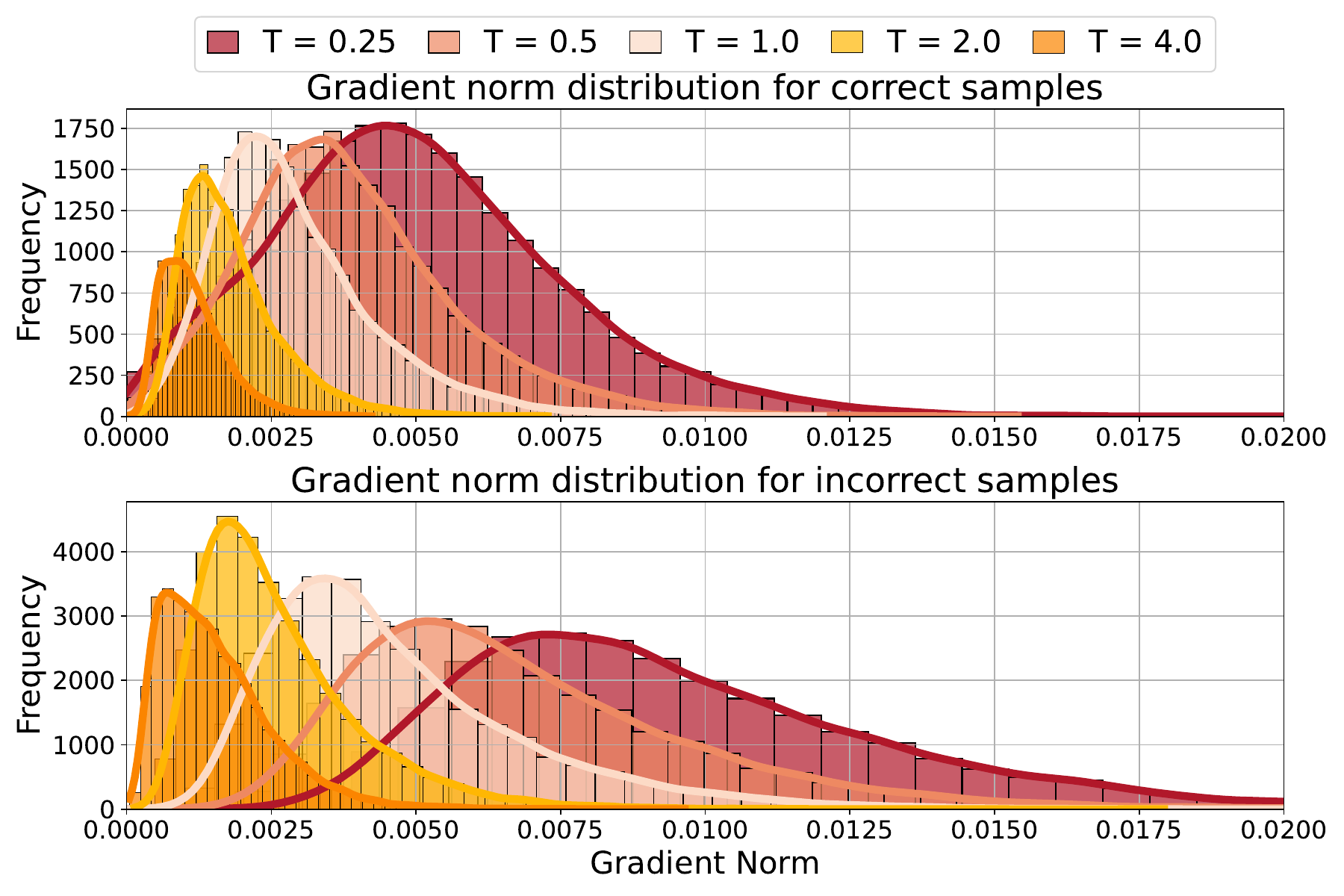}
    \caption{Distribution of gradient norm at input layer for correctly (top) / incorrectly (bottom) inferred samples with varying training temperatures.}
    \label{fig:normhist}
\end{figure}
\noindent{\textbf{Input gradient norm.}} Our first experiment focuses on assessing the norm of computed gradients within the input layer for various temperature values ($T$), and we present a sample distribution plot of the norm of input layer gradients for 50K training samples in Figure~\ref{fig:normhist}. Note that here we separate the plots for the correctly inferred samples (in the forward pass) and the incorrect ones to observe their differences. The propagation of gradients to the input layer serves as an indicative metric for the model's learning efficacy. Successful gradient propagation to the input layer implies that the model effectively utilizes and comprehends information from diverse features within the input data during the training phase~\cite{evci2022gradient}.

As shown in Figure~\ref{fig:normhist}, the mode of distribution shifts to larger values with decreasing temperatures for both correct and incorrect predictions. As mentioned, the reason behind this mode shift stems from the gradient function characteristics of softmax. As shown in Eq.~\ref{eq:boltzman_gradient}, since the gradient is multiplied by $\frac{1}{T} > 1$, for low temperatures $T < 1$, we can expect a gradient boost.

Furthermore, from the perspective of the loss term $(p_{i}-y_{i})$, when the model yields an incorrect decision, the loss term significantly increases as $p_{i}\xrightarrow{}0$. The double-folded effect accounts for a substantial mode shift with thick and long tails in the gradient distribution for lower temperatures.

\vspace{1ex}
\begin{figure}[t!]
    \centering
    \includegraphics[width=.7\linewidth]{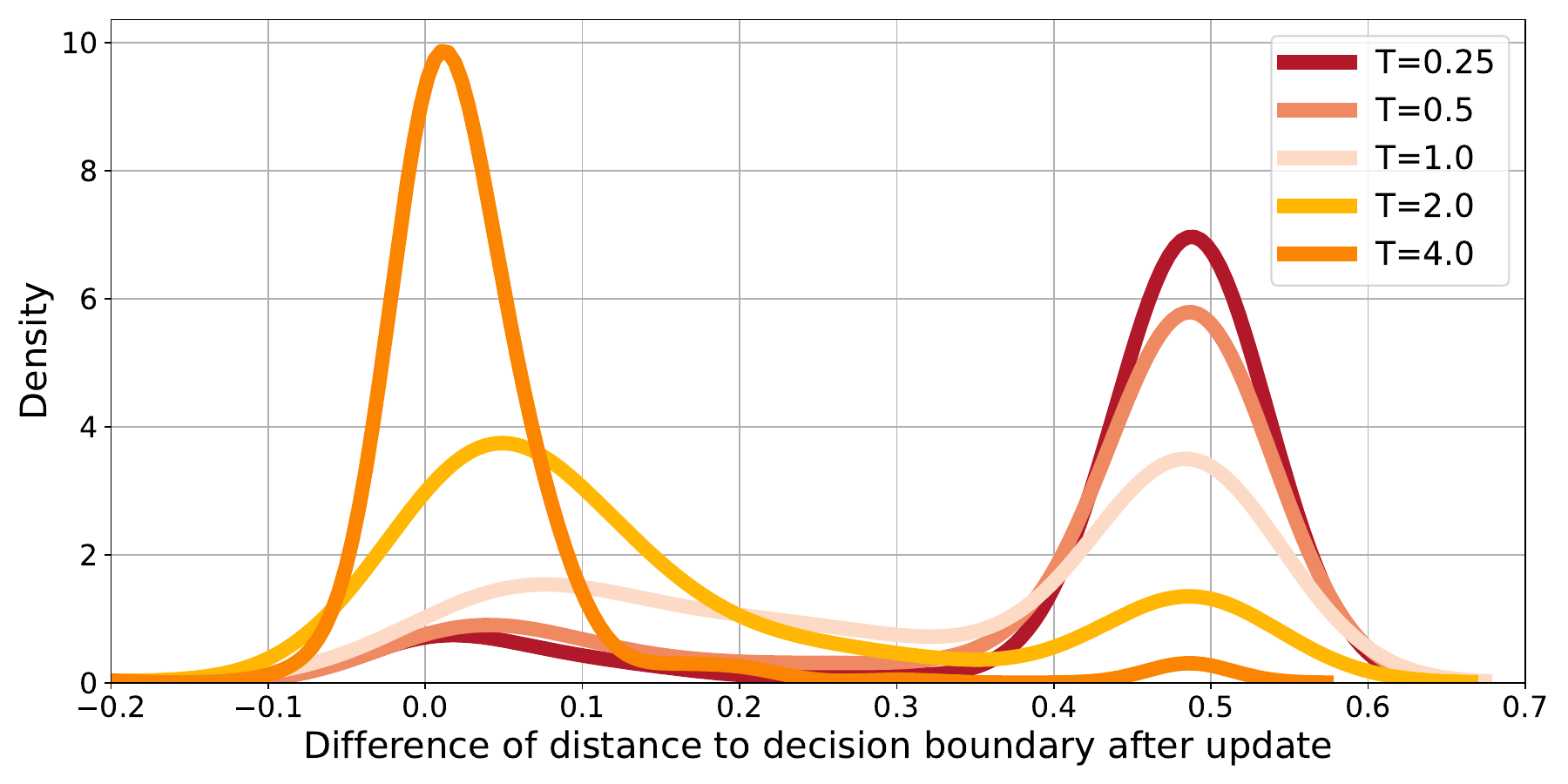}
    \vspace{-2ex}
    \caption{Distributions depicting differences between distances to the decision boundary before and after model updates for varying training temperatures. Notice that lower temperatures show a noticeable shift in estimations' positions on the representation space, suggesting their aggressiveness in modifying the model even with a small number of training samples.}
    \label{fig:feasibilitydist}
\end{figure}
\noindent{\textbf{Data position shift in the representation space.}} To better understand the impact of temperature scaling on the training process, we conducted a study into the shift of a data's position in the representation space (i.e., the distance to the decision boundary). Drawing inspiration from the concept of \textit{travel to decision boundary} as proposed by Kim et al.~\cite{kim2022curved}, we quantified the distance of a data point's inference result to the decision boundary by determining the magnitude of epsilon ($\epsilon$) value multiplied by the adversarial perturbation~\cite{goodfellow2014explaining} needed to alter the model's decision.

For this experiment, we configure an initial neural network and perform inference for 1,000 samples to identify the results' positions on the representation space. Among the 1,000, we select the ones that were misclassified, recompute their positions after performing one training operation using that specific data point, and compute the difference in positions on the representation space concerning the decision boundary. By doing so, we can observe how drastically one training operation affects the next inference output to different temperatures used in the training phase. 

Figure~\ref{fig:feasibilitydist} plots our results in the form of distributions depicting differences between distances to the decision boundary before and after model updates. When higher temperatures ($T={2, 4}$) are applied, we can notice that the changes in inference results' positions are not significant (i.e., plots skewed towards 0). Whereas, with a small $T$, we can notice more pronounced shifts in data positions within the representation space. This suggests that using a small $T$ at the softmax operations can actively steer the model towards (potentially) substantial optimization to better align with the training data.

\vspace{1ex}
\noindent{\textbf{Visualization in the representation space}}
\begin{figure*}[t!]
    \centering
    \includegraphics[width=\linewidth]{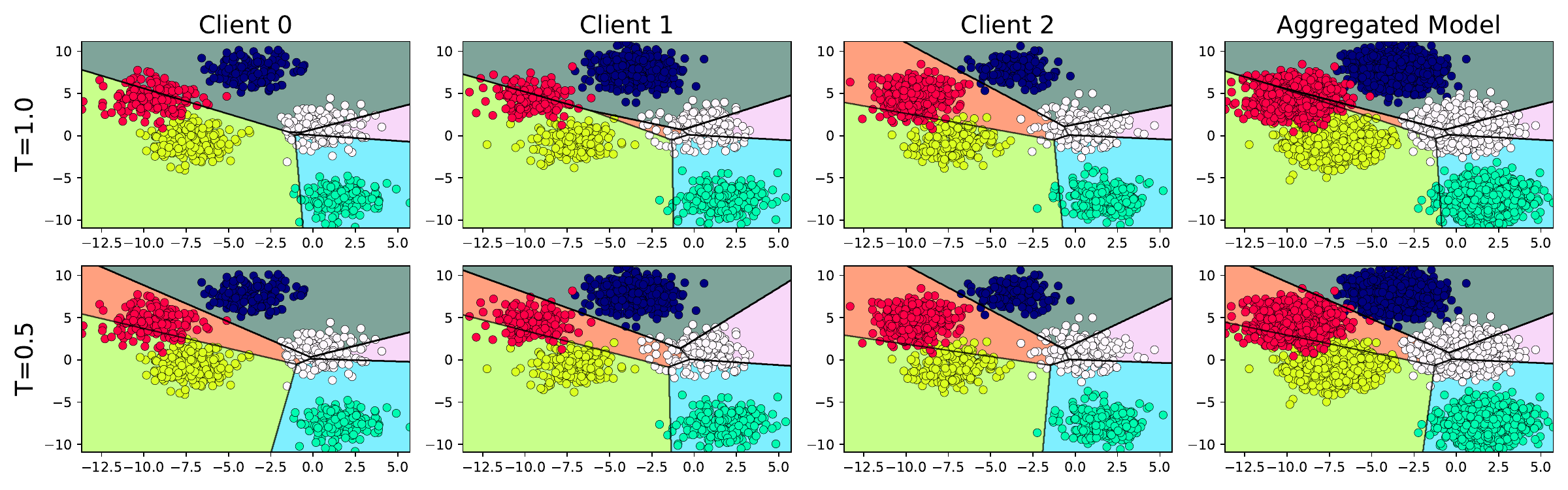}
    \vspace{-1ex}
    \caption{Example of data points in the 2D representation space with their respective classification boundaries for different federated learning clients with varying training temperatures. Best viewed in color.}
    \label{fig:toyexample}
    \vspace{-3ex}
\end{figure*}
To examine the effect of applying a low temperature during the model training phase, we conducted a simple example that classifies 2-dimensional data via logistic regression. We split the data into three different subsets and assigned them as three devices with non-i.i.d\rev{.} data. The Softmax layer was applied at different temperatures ($T$=1.0, 0.5) during the local training phase and average model weights for aggregation after 10 local updates.

Figure~\ref{fig:toyexample} illustrates the classification boundaries and samples distributed in the representation space for the three clients and the aggregated model. The results suggest that applying $T$ = 0.5 effectively reflects the local dataset characteristics compared to $T$ = 1.0. Specifically, as an example, for clients 0, 1, and the aggregate model, using $T$ = 1.0 fails to make a clear distinction between the red and light green samples; whereas, when applying $T$ = 0.5, this distinction becomes pronounced.

% The learning rate governs the trajectory of convergence by managing how aggressively the model updates parameters, whereas the temperature affects the direction of the update during training. Consequently, adjusting these two hyperparameters should be done with caution, as their interactions can yield complex and unintended effects on both convergence speed and final model performance. The impact of scaling the logits $z_{i}$ with temperature $T$ can be simply shown by training the models with varying the learning rate $\eta$ and temperature $T$ in the same proportion (i.e., maintaining a constant $\frac{\eta}{T}$).
% Furthermore, we conduct our experiments with the CIFAR10 dataset~\cite{krizhevsky2009learning} and use a 2-layered CNN architecture~\cite{mcmahan2017communication} as the baseline neural network architecture. Due to the limited space, we leave the results and details of the experiments in Appendix~\ref{sec:exp_detail}.

\subsection{\name{} and Logit Chilling}
\label{sec:proposal}

%\jk{We need to detail what \name{} actually is. Is it a framework?} \kc{I think it would be an exaggerated claim to say it is a new framework... It's more like an optimization strategy or a modification of the local training objective.}

% Based on the theoretical analysis and our empirical observations, we observe that training neural networks with lower temperatures can benefit the model convergence and performance for systems exploiting non-i.i.d. data, which is common for federated learning scenarios. We formulate this observation as an optimization strategy, termed \name{}. Specifically, \name{} exploits \textit{logit chilling}, as we `chill down' the temperature at the softmax operations, thereby altering the logit computation derived from the neural network during the training process. The focus of a \name{}-based federated learning system is to exploit logit chilling to train a client-side model with low-temperature values. As we observed through the studies above and as our evaluations will show, \name{} effectively exploits crucial information from the training samples to actively alter the model towards the new knowledge. %Naturally, this increases the individual model confidence and by aggregating such models the global model 

\rev{Based on the theoretical analysis and our empirical observations, we find that training neural networks with lower temperatures directly amplifies the gradient magnitude and accelerates convergence in the presence of non-i.i.d. data, which is a typical regime for federated learning. In particular, our analysis shows that the softmax cross-entropy gradient scales with the factor $1/T$, and the resulting convergence bound includes the term $\eta\mu/T$, indicating faster convergence when $T<1$. Complementing this, our empirical studies in the representation space (e.g., Figures~\ref{fig:feasibilitydist} and~\ref{fig:toyexample}) show that lower temperatures move samples farther from the decision boundary, especially for the misclassified points, leading to more aggressive and influential local updates at each client.}

\rev{We formulate these insights as an explicit optimization strategy, termed \name{}. Specifically, \name{} exploits \textit{logit chilling} by ``chilling down'' the temperature at the softmax layer during local training on each client, thereby sharpening the output distribution and amplifying gradients throughout the network. The focus of a \name{}-based federated learning system is to leverage these low-temperature updates so that each client can more strongly encode new knowledge into its local model, and the subsequent aggregation of these models at the server accelerates the global convergence toward a high-quality solution. As our evaluations in the following sections demonstrate that \name{} consistently achieves faster convergence and improved accuracy compared to conventional training with $T=1$ under various non-i.i.d. federated learning scenarios.}

We formulate the operations of \name{} using Algorithm~\ref{alg:system}. Here, typical federated learning operations take place at both the server and clients except for the fact that the server determines a target $T$ at each round and local model training operations are performed with $T$. Since \name{} only manipulates the local training process which inherently provides orthogonality to our proposed method, it is extremely easy to integrate with existing federated learning frameworks \rev{(e.g., FedProx~\cite{li2020federated}, SCAFFOLD~\cite{karimireddy2020scaffold}, FedBN~\cite{li2021fedbn})}.

\begin{algorithm}[ht!]
    \small
    \caption{\name{}}
    \label{alg:system}
    \textbf{Data} ($D_{1}, D_{2}, ..., D_{N}$) where $D_{i}$ is the user $i$'s local data.\\
    \textbf{Model} ($w^{1}_{t}, w^{2}_{t}, ..., w^{N}_{t}$) where $w_{t}^{i}$ is the user $i$'s model in $t$-th round.\\
    % \textbf{Input}: $w_{t}^{i}$, $D^{i}$\\
    \textbf{Parameter}: Temperature $T$
    
    \begin{algorithmic} %[1] %[1] enables line numbers
    \STATE \textbf{Server executes}:
        \STATE Initialize $w_{0}$
        \FOR{each round $t = 1, 2, ..., R$}
        \STATE $S_{t} \xleftarrow{}$ (random set of $m$ clients)
            \FOR{$k \in S_{t}$}
            \STATE $w_{t}^{k}\xleftarrow{}$ ClientUpdate($w_{t}, T$)
            \ENDFOR
            \STATE $w_{t+1}\xleftarrow{}$ Aggregation($S_{t}$)
        \ENDFOR
    
    \STATE \textbf{ClientUpdate}($w_{t}$, $T$):
        \STATE $w_{t}^{k}\xleftarrow{}w_{t}$
        \FOR{data, label $\in D_k$}
        \STATE $\mathbf{z} = w_{t}^{k}(data) / T$
        \STATE $l(\mathbf{z})=loss(\mathbf{z}; label)$
        \STATE $w_{t}^{k} \xleftarrow{}LocalUpdate(w_{t}^{k}, l(\mathbf{z}))$
        % \State $w_{t}^{k} \xleftarrow{}w_{t}^{k}-\eta \nabla l(\mathbf{z})$
        \ENDFOR
    \STATE Return $w_{t}^{k}$ to server
    \end{algorithmic}
\end{algorithm}

The low temperatures used by Logit Chilling directly influence the local training process of each federated learning client. Notably, the seamless integration of Logit Chilling and \name{} with existing federated learning frameworks is a distinct advantage, allowing for a straightforward implementation within already deployed frameworks. \rev{Furthermore, as mentioned, we note that \name{} can be used orthogonally with other efforts that accelerate federated learning training operations, such as FedProx, SCAFFOLD, and FedBN.} We provide an open-sourced implementation of \name{} at \urlcolor{\url{https://github.com/eis-lab/temperature-scaling}}.

%% file: 4_experiment.tex
\section{\rev{EVALUATION}}
\label{sec:experiment}
We evaluate the performance of \name{} using three datasets: (i) FEMNIST from the LEAF database~\cite{caldas2018leaf}, including the Extended MNIST handwriting data~\cite{cohen2017emnist}%partitioned based on the owner of the 62 digits/characters,
, (ii) the CIFAR10 dataset, and (iii) the CIFAR100 dataset~\cite{krizhevsky2009learning}. As metrics, we use the average accuracy and training loss observed at the clients, selected to observe both the performance gain and the effectiveness of \name{} in its training operations. Furthermore, we also present the number of federated learning rounds needed to achieve a pre-defined target accuracy.

\subsection{Experiment Setup}
\label{sec:exp_setup}
\noindent{\textbf{Dataset.}} We detail the datasets used in our work as follows. (i) The FEMNIST dataset comprises of 10 handwritten digits (0-9) and 52 characters (26 lowercase and 26 uppercase) images contributed by 712 users, totaling 157,132 samples. We selected a subset of 36 users, each with diverse amounts of local data to account for the non-i.i.d. environment. We followed the implementations provided by the original work~\cite{caldas2018leaf}. (ii) We use baseline CIFAR10 and CIFAR100 datasets~\cite{krizhevsky2009learning} and manually partition data over 50 federated learning users, with each holding 1,000 samples.

\begin{figure}[t]
    \centering
    \subfigure[FEMNIST]{
    \includegraphics[width=.33\linewidth]{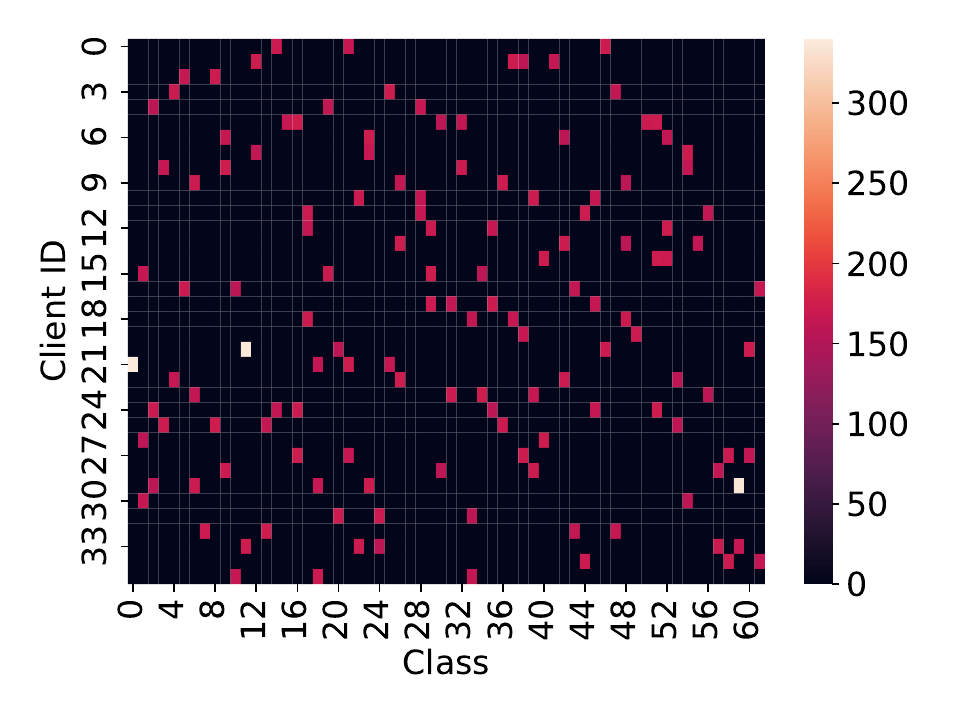}
    }
    \hspace{-3.75ex}
    \subfigure[CIFAR10]{
    \includegraphics[width=.33\linewidth]{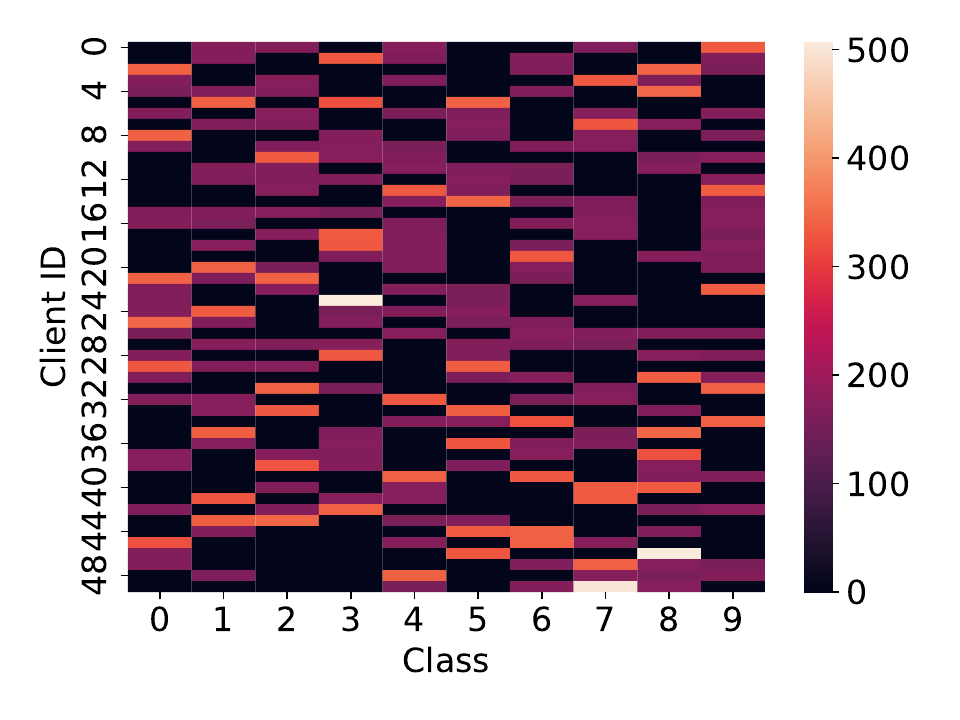}
    }
    \hspace{-3.75ex}
    \subfigure[CIFAR100]{
    \includegraphics[width=.33\linewidth]{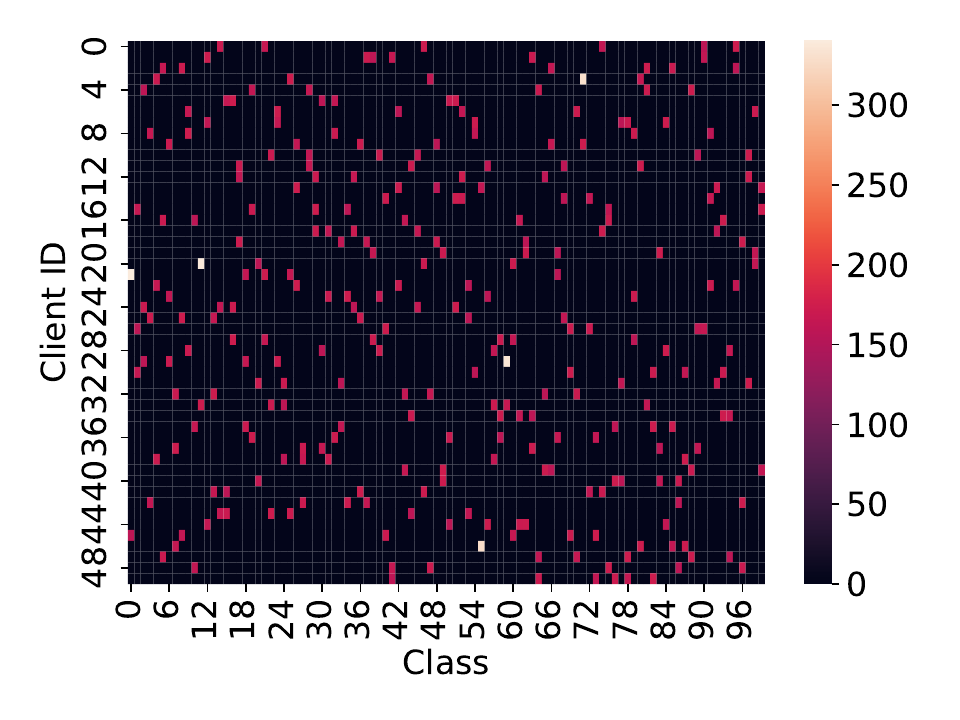}
    }
    \caption{Visualization of the distribution of training data used in FEMNIST, CIFAR10-CNN and CIFAR100-ResNet experiments. Best viewed in color.}
    \label{fig:basic_datadist}
\end{figure}
\begin{figure}[t]
    \centering
    \subfigure[Severe Heterogeneity ($\alpha$=0.1)]{
    \includegraphics[width=.3\linewidth]{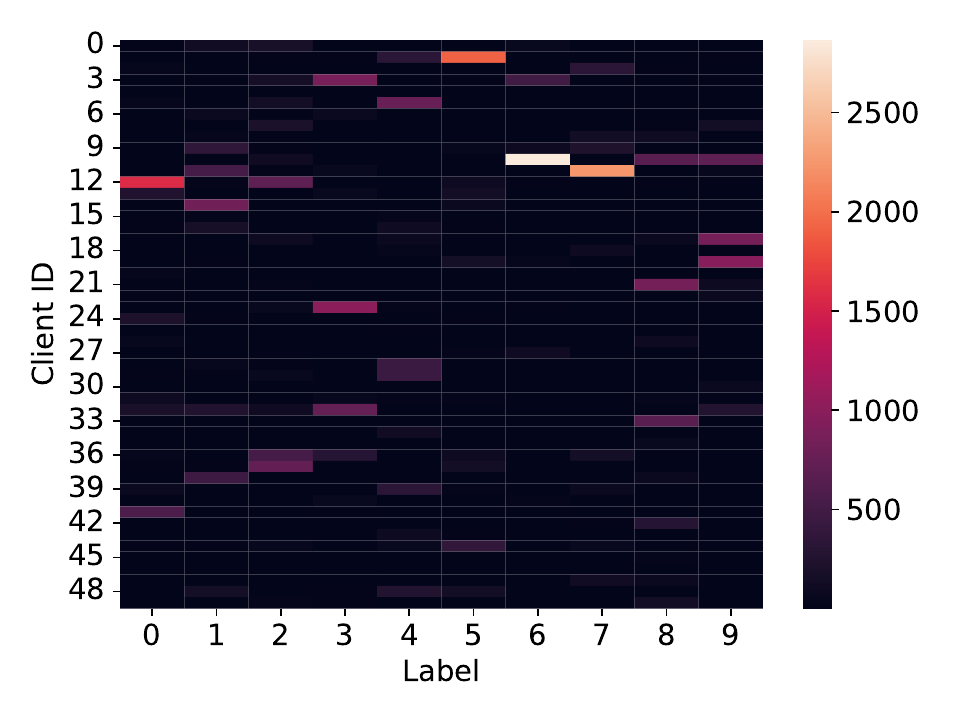}
    }
    % \hspace{-1.ex}
    \subfigure[Moderate Heterogeneity ($\alpha$=0.5)]{
    \includegraphics[width=.3\linewidth]{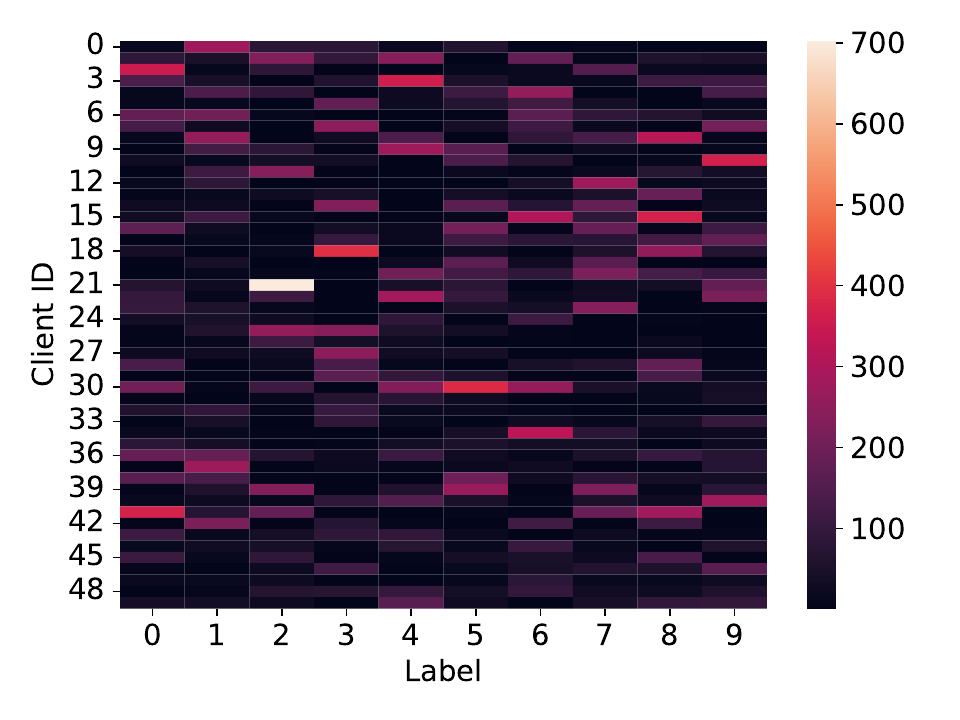}
    }
    % \hspace{-1.ex}
    \subfigure[Weak Heterogeneity ($\alpha$=1.0)]{
    \includegraphics[width=.3\linewidth]{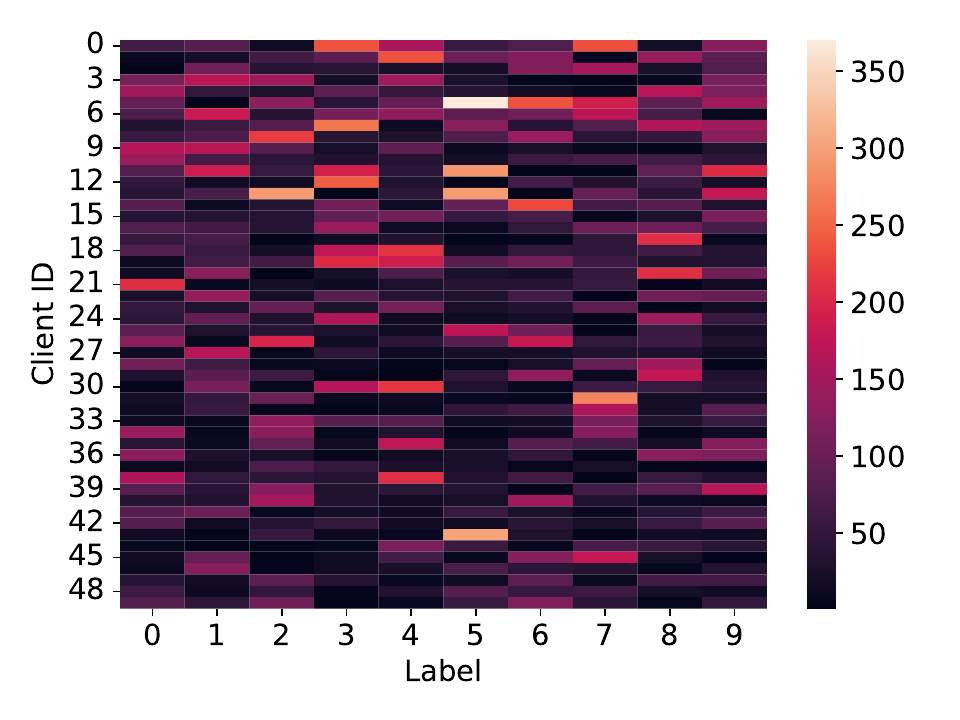}
    }
    \caption{\rev{Visualization of training data distribution partitioned by the Dirichlet-based method. Best viewed in color.}}
    \label{fig:dir_datadist}
\end{figure}

\rev{Figure~\ref{fig:basic_datadist} visualizes the data distribution used for our evaluations (c.f., Sec~\ref{sec:experiment}). In this section, we employ two complementary dataset‐splitting strategies.} First, in Sections~\ref{subsec:averageperform},~\ref{subsec:convergence}, and~\ref{subsec:stability}, we use pathological (shard-based) partitioning, a common approach for simulating real-world data heterogeneity. Here, each class’s data is divided into fixed-size shards, and each client is assigned a subset of these shards without overlap. By varying both the number of shards per client and the shard size, we induce controlled quantity skew (different total data amounts) and label skew (uneven class distributions) across clients. \rev{In image-based tasks such as CIFAR10 and CIFAR100, this shard-based assignment also implicitly introduces feature-level skew, since different clients see different subsets of the underlying intra-class variations (e.g., orientation, background, and lighting), leading to mixed heterogeneity in both label composition and input characteristics.}

\rev{Second, for all other experiments in Section~\ref{sec:experiment}, we adopt a Dirichlet-based dataset partitioning. Figure~\ref{fig:dir_datadist} shows the data distribution sampled via Dirichlet-based partitioning over the clients. Concretely, let $K$ be the number of classes and $C$ the number of clients.} We first draw, for each client $i$, a class‐proportion vector
$$
\mathbf{p}^{(i)} \sim \mathrm{Dirichlet}(\alpha, \dots, \alpha)\,,
$$
where $\alpha>0$ is the concentration parameter. Then, for each class $k$, we allocate to client $i$ exactly
$$
n_k^{(i)} = \bigl\lfloor\,|\mathcal{D}_k|\;p_k^{(i)}\bigr\rfloor
$$
samples drawn without overlap from the global pool $\mathcal{D}_k$. In this way, each client ends up with a total of $\sum_{k}n_k^{(i)}$ samples whose class composition follows $\mathbf{p}^{(i)}$. By tuning $\alpha$, we control the degree of heterogeneity: (i) A small $\alpha$ (e.g.\ $\alpha=0.1$) yields highly skewed, non‐\rev{i.i.d.} splits (clients see only a few classes); (ii) A moderate $\alpha$ (e.g.\ $\alpha=0.5$) induces moderate class imbalance; (iii) A large $\alpha$ (e.g.\ $\alpha\ge1$) approaches the \rev{i.i.d.} regime. This Dirichlet‐based method, therefore, provides a continuous spectrum of label and quantity skew, allowing us to systematically evaluate our algorithms under varying non‐\rev{i.i.d.} conditions. \rev{Similar to the shard-based setting, when applied to image datasets, Dirichlet partitioning also induces mixed skew with clients differing not only in class proportions but also in the subset of visual factors they observe; resulting in both label and feature heterogeneity.}

\rev{We note that these two widely adopted non-i.i.d. benchmarks (shard-based and Dirichlet-based partitioning) already create mixed heterogeneity patterns that combine label skew, quantity skew, and, in the case of image datasets, feature-level skew arising from intra-class variability.}

\vspace{2ex}

\noindent{\textbf{Model.}} For each dataset, we utilized distinct model architectures, taking into account the specific challenges and characteristics inherent to each dataset. For FEMNIST, an MLP model featuring four linear layers was employed. We term this combination as ``FEMNIST-DNN'' in the rest of our evaluations. In the case of the CIFAR10 and CIFAR100 datasets, experiments were conducted using a two-layer CNN model~\cite{mcmahan2017communication} and a ResNet18~\cite{he2016deep}, respectively. These two cases are denoted as ``CIFAR10-CNN'' and ``CIFAR100-ResNet18.'' We note that the impact of temperature scaling during the local training process is a model/dataset agnostic phenomenon. 
\begin{table}[h!]
\centering
\begin{adjustbox}{width=.4\linewidth,center}
\begin{tabular}{clc}
\hline
Layer    & \multicolumn{1}{c}{Details}                                       & Repetition \\ \hline
FC 1     & \begin{tabular}[c]{@{}l@{}}Linear(784, 512)\\ ReLU()\end{tabular} & $\times 1$ \\ \hline
FC 2     & \begin{tabular}[c]{@{}l@{}}Linear(512, 256)\\ ReLU()\end{tabular} & $\times 1$ \\ \hline
FC 3     & \begin{tabular}[c]{@{}l@{}}Linear(256, 128)\\ ReLU()\end{tabular} & $\times 1$ \\ \hline
FC 4     & Linear(256, 62)                                                   & $\times 1$ \\ \hline
\end{tabular}
\end{adjustbox}
\caption{Model architectures details of DNN used to train the FEMNIST dataset}
\label{tab:mlp}
\end{table}

\begin{table}[h!]
\centering
\begin{adjustbox}{width=.5\linewidth,center}
\begin{tabular}{clc}
\hline
Layer  & \multicolumn{1}{c}{Details}                                                                                            & Repetition \\ \hline
Conv 1 & \begin{tabular}[c]{@{}l@{}}Conv2d(3, 10, k=(5, 5), s=(1, 1))\\ ReLU()\\ MaxPool2d(k=(2, 2))\end{tabular} & $\times 1$ \\ \hline
Conv 2 & \begin{tabular}[c]{@{}l@{}}Conv2d(10, 20, k=(5, 5),s=(1, 1))\\ ReLU()\\ MaxPool2d(k=(2, 2))\end{tabular} & $\times 1$ \\ \hline
FC 1   & \begin{tabular}[c]{@{}l@{}}Linear(500, 256)\\ ReLU()\end{tabular}                                                      & $\times 1$ \\ \hline
FC 2   & Linear(256, 10)                                                                                           & $\times 1$ \\ \hline
\end{tabular}
\end{adjustbox}
\caption{Model architectures details of 2-layered CNN used to train the CIFAR10 dataset}
\label{tab:cnn}
\end{table}

Table~\ref{tab:mlp} and \ref{tab:cnn} show the detailed model architecture used for the experiments, respectively. For the ResNet18 model used to train the CIFAR100 dataset, we exploited PyTorch implementation~\cite{NEURIPS2019_9015} by only replacing the final classification head from \texttt{Linear(512, 1000)} to \texttt{Linear(512, 100)}. It is important to note that low temperatures may cause unstable convergence in large models (e.g., ResNet, ViT). We recommend incorporating normalization layers to mitigate this effect. We further note that investigating methods to minimize instability caused by low temperatures with respect to different model architectures or components is an interesting direction for future work.

In real‐world deployments, clients often employ heterogeneous model architectures tailored to their specific applications and resource constraints. In this work, however, we restrict our evaluation to a homogeneous setting in which all clients share the same model architecture, thereby abstracting away variations in computational and communication capabilities. Extending our approach to support architectural heterogeneity and diverse client resources is an important avenue for future research, as it would more closely reflect practical federated learning scenarios.

\vspace{2ex}

\noindent{\textbf{Experiment Details.}} Each case underwent local training for 300 federated training rounds, utilizing FedAvg~\cite{mcmahan2017communication} as the baseline federated learning framework. FedAvg is a widely used framework for federated learning. In each round, the server selects 10 participating clients, employing Stochastic Gradient Descent (SGD) as the optimizer with a learning rate of 0.001~\cite{ruder2016overview}. Each client trained its model for 10 local epochs, employing a batch size of 16, and utilized cross-entropy loss as the loss function. Unless specified otherwise, the experiment was conducted using the same hyperparameters as mentioned in this section.
Table~\ref{tab:hyperparams} shows the default hyperparameters used for the experiments. We note that we applied a learning rate decay of $10^{-5}$ to enhance training stability. Although the effect was minimal and did not alter the trend of the main results, we recommend its application.
\begin{table}[h!]
\centering
\begin{adjustbox}{width=.7\linewidth,center}
\begin{tabular}{|c|cccc|}
\hline
Hyperparameters        & \multicolumn{1}{c|}{FEMNIST} & \multicolumn{1}{c|}{CIFAR10} & \multicolumn{1}{c|}{CIFAR100} & CKA Eval \\ \hline
total clients          & \multicolumn{1}{c|}{36}          & \multicolumn{2}{c|}{50}                                                   & 10             \\ \hline
communication rounds   & \multicolumn{3}{c|}{300}                                                                                     & 100            \\ \hline
optimizer              & \multicolumn{4}{c|}{SGD}                                                                                                      \\ \hline
learning rate          & \multicolumn{3}{c|}{0.001}                                                                                   & 0.005          \\ \hline
local epoch            & \multicolumn{3}{c|}{10}                                                                                      & 10             \\ \hline
participants per round & \multicolumn{3}{c|}{10}                                                                                      & 10             \\ \hline
batch size             & \multicolumn{3}{c|}{16}                                                                                      & 32             \\ \hline
shard size | $\alpha$  & \multicolumn{1}{c|}{N/A}         & \multicolumn{2}{c|}{200}                                                  & $\alpha$=0.5   \\ \hline
loss function          & \multicolumn{4}{c|}{Cross Entropy}                                                                                            \\ \hline
\end{tabular}
\end{adjustbox}
\caption{Hyperparameters used for Section~\ref{sec:experiment}. $\alpha$ denotes the importance of the Dirichlet distribution.}
\label{tab:hyperparams}
\end{table}

Finally, for the experiments, we exploited a server with four GeForce RTX 2080 Ti GPUs, Intel(R) Xeon(R) Silver 4210 CPU @ 2.20GHz, and 64GB RAM.

% \subsection*{Data distribution}
% \label{sec:basic_datadist}

\subsection{Average Global Model Accuracy and Loss}
\label{subsec:averageperform}
\begin{figure*}[t!]
    \centering
    \subfigure[FEMNIST-DNN]{
    \includegraphics[width=0.3\linewidth]{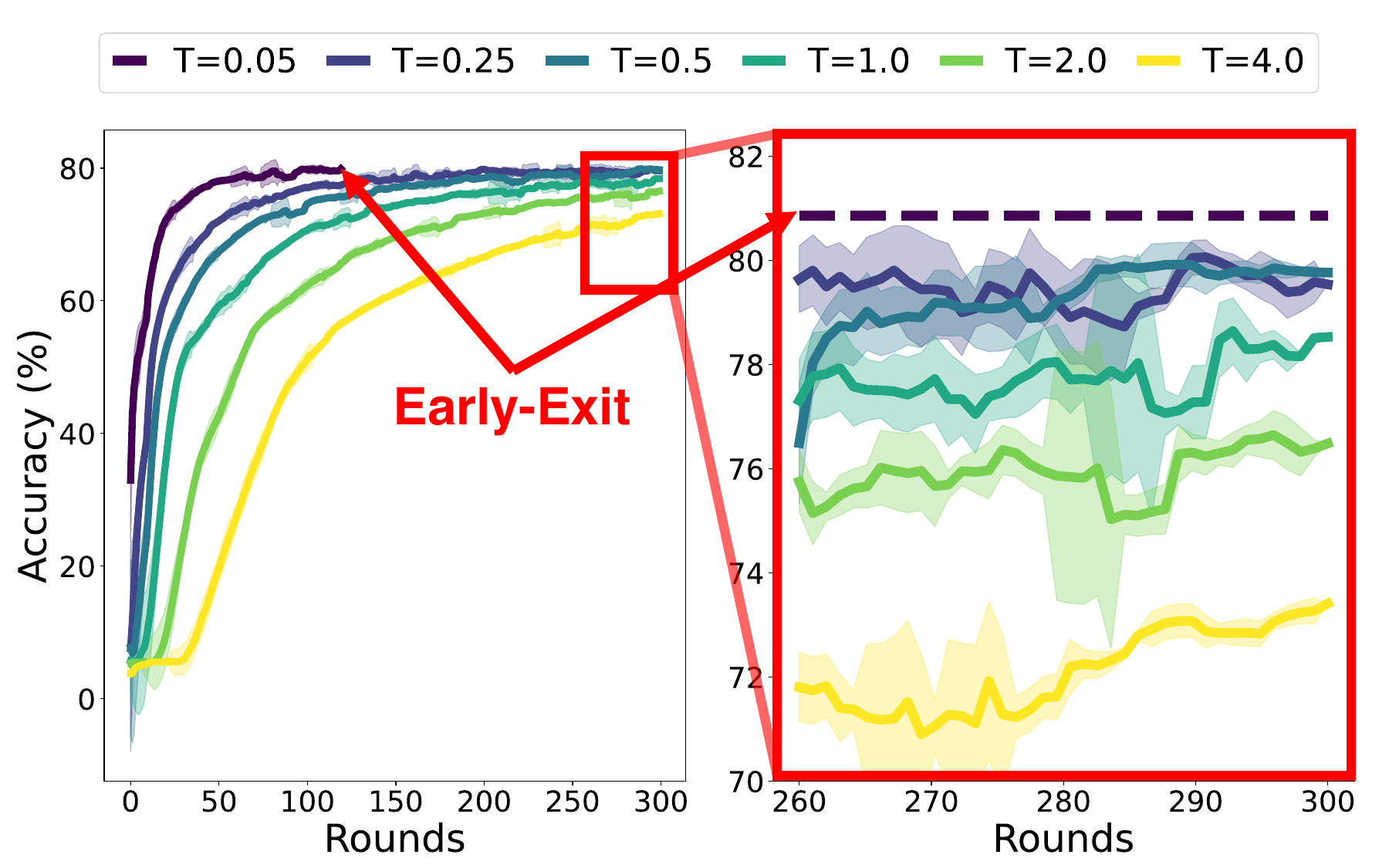}
    }
    \subfigure[CIFAR10-CNN]{
    \includegraphics[width=0.3\linewidth]{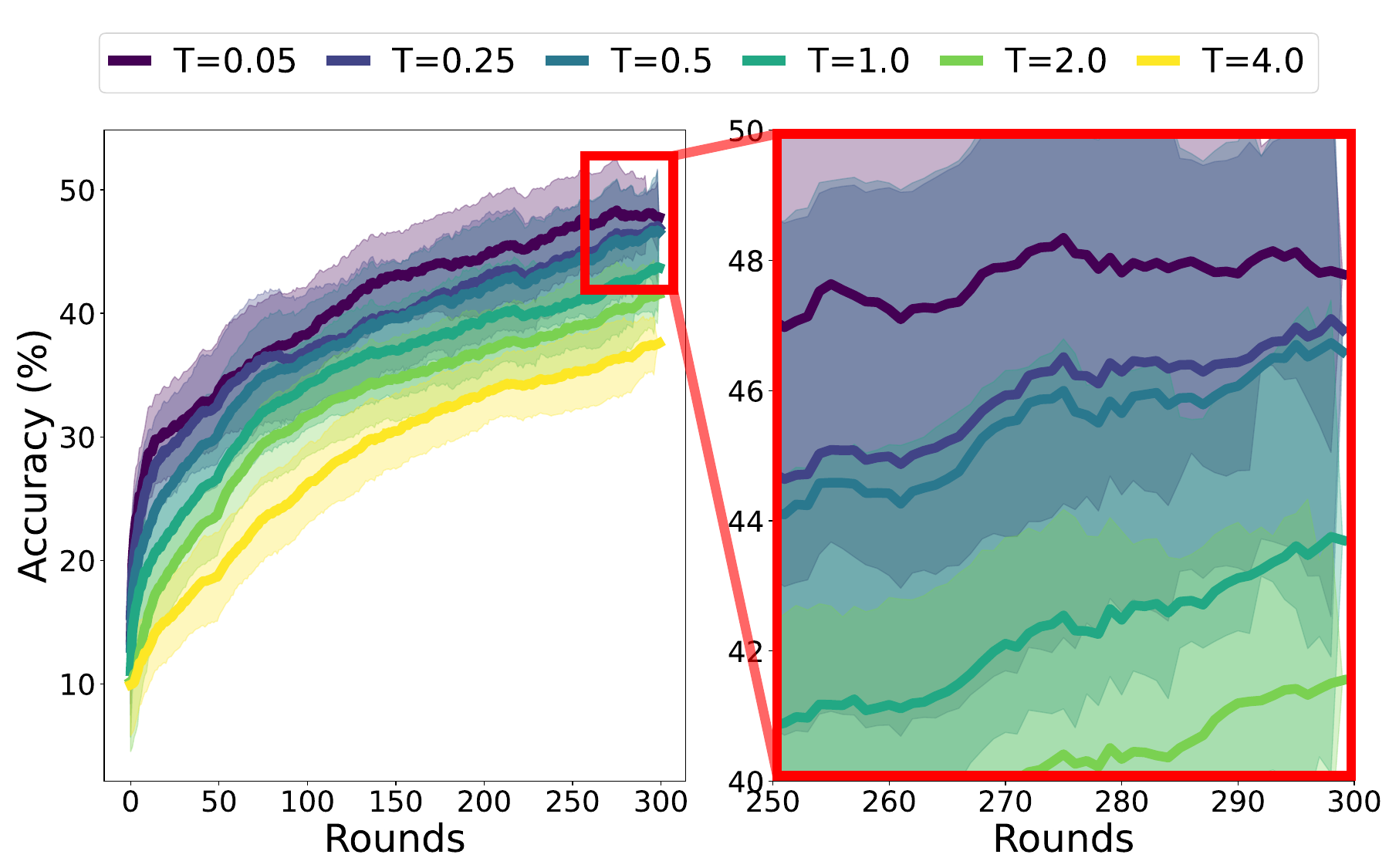}
    }
    \subfigure[CIFAR100-ResNet18]{
    \includegraphics[width=0.3\linewidth]{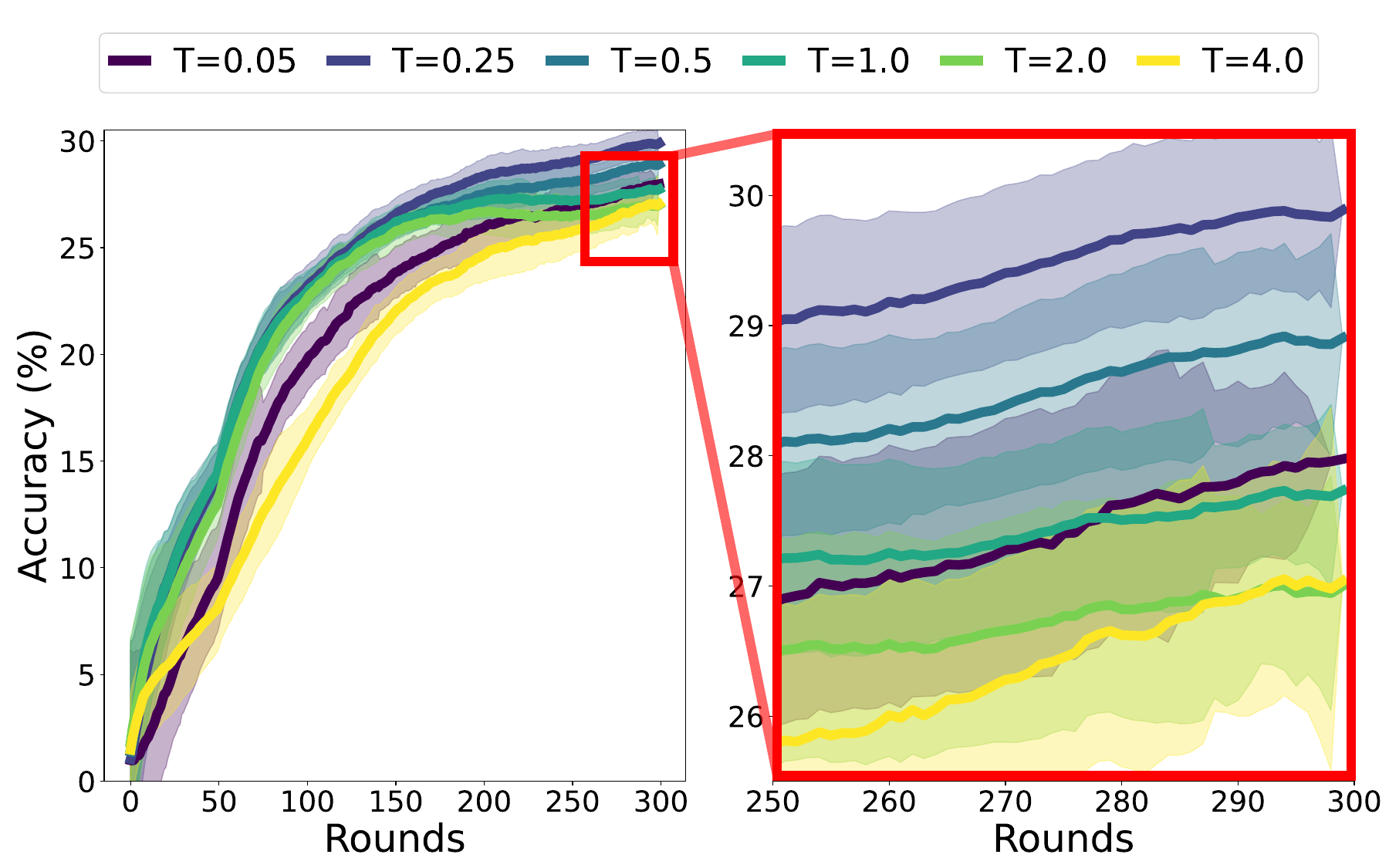}
    }
    \caption{Average global model accuracy for different temperatures used in training. Overall, lower temperatures show higher average model accuracy for federated learning systems. The shaded areas indicate the standard deviation across different runs. \rev{Note that the FEMNIST–DNN configuration exhibited relatively low variance compared to the other settings, largely due to its simpler architecture and lower overall complexity.}}
    \label{fig:acc}
\end{figure*}

\begin{figure*}[t!]
    \centering
    \subfigure[FEMNIST-DNN]{
    \includegraphics[width=0.3\linewidth]{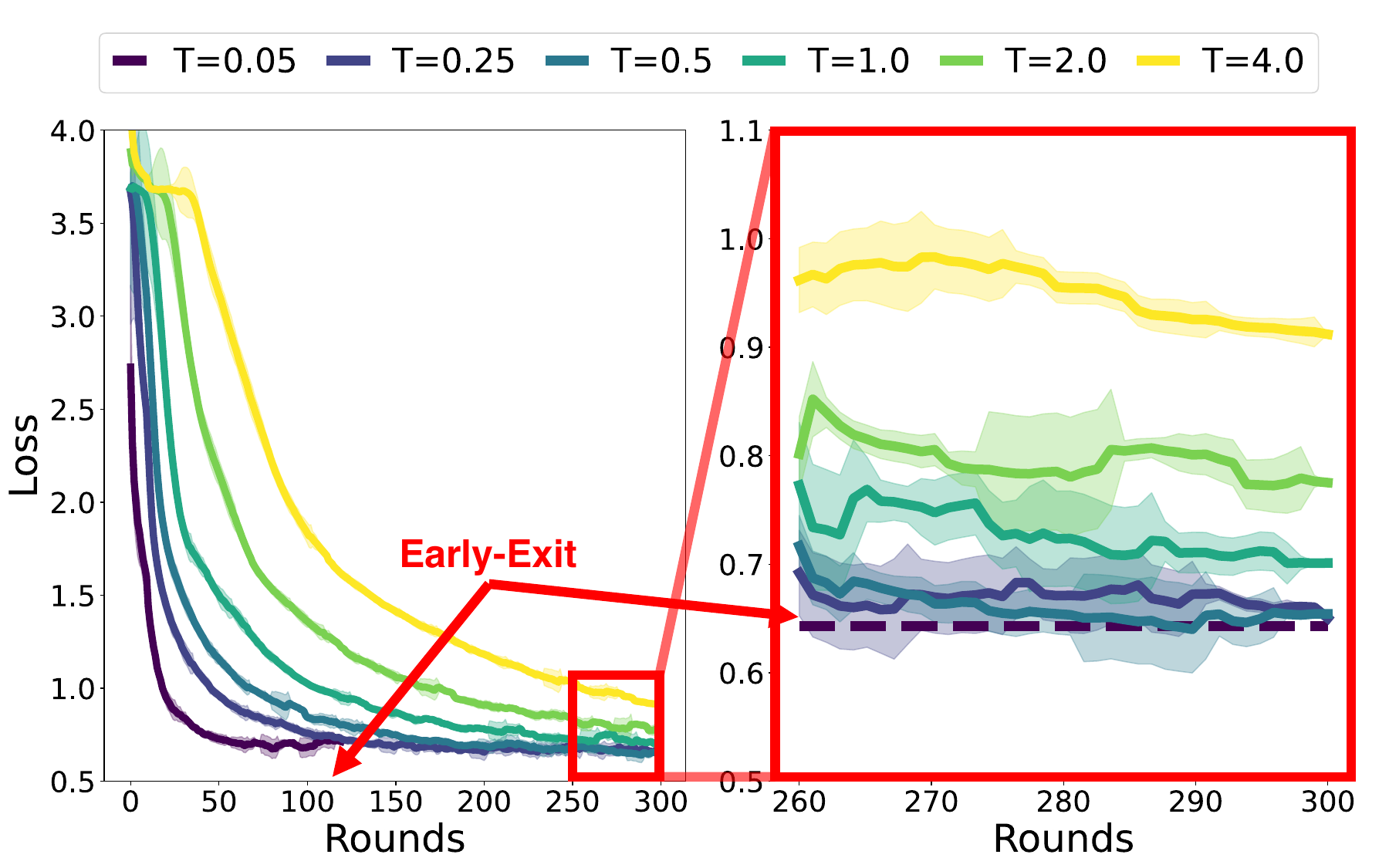}
    }
    \subfigure[CIFAR10-CNN]{
    \includegraphics[width=0.3\linewidth]{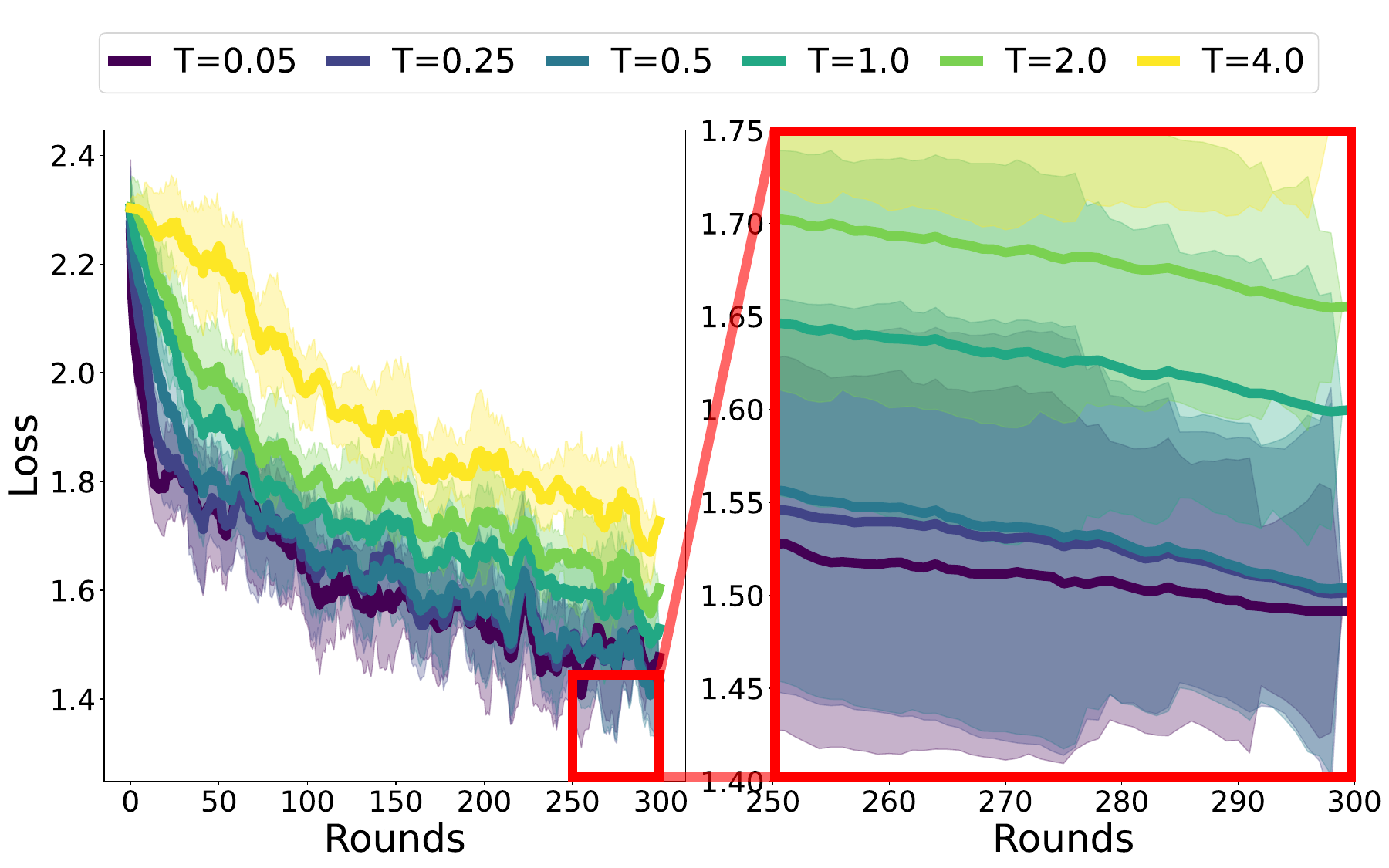}
    }
    \subfigure[CIFAR100-ResNet18]{
    \includegraphics[width=0.3\linewidth]{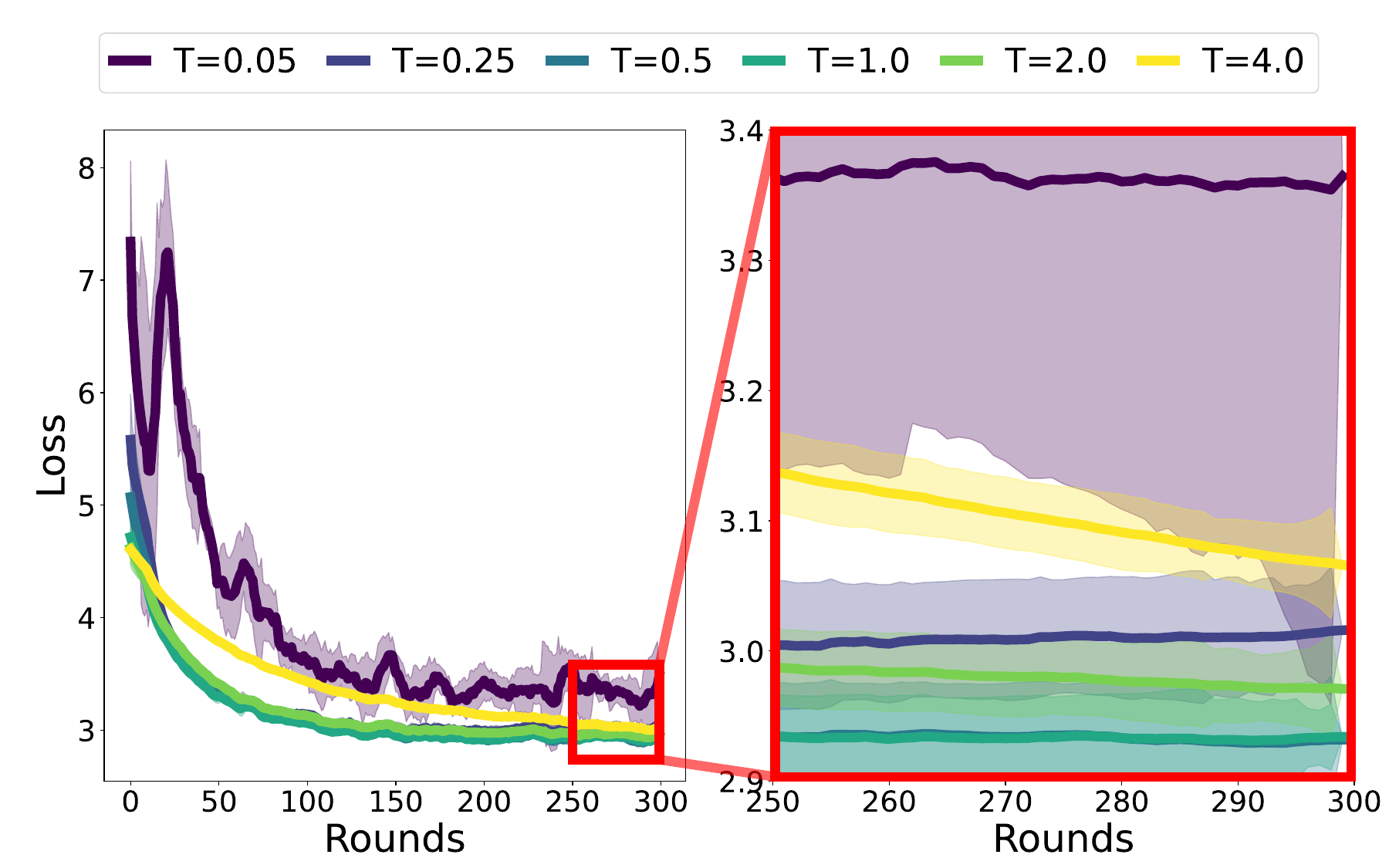}
    }
    \caption{Average global model loss for different temperatures used in training. Overall, lower temperatures show a lower average model loss for federated learning systems. The shaded areas indicate the standard deviation across different runs.}
    \label{fig:loss}
\end{figure*}
We report the average global model accuracy obtained for models trained with different temperatures in Figure~\ref{fig:acc}. For all three datasets (and the three respective models used for each test case) we see an increasing accuracy trend observed for lower temperatures. Note that for the FEMNIST-DNN case with $T$=0.05, we noticed that the accuracy converges after 120 rounds; thus, perform an early exit for this case. One interesting aspect we can notice is that the accuracy for the $T$=0.05 case, which is the lowest temperature we tested for, does not show the best performance for the CIFAR100-ResNet18 experiment. While the $T$=0.05 case shows the best performance of FEMNIST-DNN and CIFAR10-CNN, for CIFAR100-ResNet18, we see an accuracy worse than the $T$=0.5 case. This suggests that training with the lowest possible temperature value is not always ideal, rather this result highlights the importance of identifying a proper training temperature as we will further discuss in Section~\ref{sec:discussion}. A closer zoom into the plots suggests that the best-performing configuration can change over the training round and can vary for different datasets, but overall, we can notice that exploiting a temperature value lower than 1 in the training process can help improve the classification accuracy.

The average accuracy for all federated learning clients after 300 training rounds presented in Table~\ref{tab:overall} confirm our observations, indicating that with $T$=0.05, the CIFAR10 case shows a 3.37\% improvement in accuracy compared to the $T$=1 case. Even for other cases, the use of a low-temperature value shows promising results in terms of accuracy. We emphasize once more that this does not mean that the accuracy will always perform better at the lowest temperature. Rather, our results suggest the need to carefully explore the fractional temperature space in the federated learning training process. We note that the performance for model performance and convergence speed tends to be similar regardless of federated learning hyperparameters (i.e., batch size, local epoch, number of participants).

Next, we examine the validation loss of the global model. In Figure~\ref{fig:loss} we present the average loss observed at each federated learning training round. Again, we note that we perform early exit for FEMNIST-DNN at 120 rounds when $T$=0.05 due to quick model convergence. We can notice here that the use of lower temperatures in the training phase generally helps to quickly lower the validation loss during the local training phase for federated learning. At the same time, when observing the CIFAR100-ResNet18 plots, we can see that the loss for $T$=0.05 does not exhibit stable convergence, serving as a reason behind the lower accuracy reported for CIFAR100-ResNet18.

From Table~\ref{tab:overall} where we report the average validation loss after 300 training rounds (120 for the FEMNIST-DNN @ $T$=0.05), we can make similar observations. Especially, for FEMNIST-DNN, we see a six-fold improvement in loss compared to $T$=1. For the CIFAR100-ResNet18 case, however, we notice that the lowest loss convergence is at $T$=0.25 rather than 0.05.
%This observation emphasizes the need to properly consider and explore fractional temperature values for client-side model training. 
As noted in Section~\ref{sec:design}, this observation emphasizes the need to properly consider and explore fractional temperature values for client-side model training to balance between the boosting convergence and the instability during local training or the risk of overfitting.

\subsection{Federated Learning Convergence}
\label{subsec:convergence}
\begin{table*}[t]
\begin{adjustbox}{width=\linewidth,center}
\begin{tabular}{|c|ccc|ccc|ccc|}
\hline
                  & \multicolumn{3}{c|}{FEMNIST-DNN}                                                                        & \multicolumn{3}{c|}{CIFAR10-CNN}                                                                         & \multicolumn{3}{c|}{CIFAR100-ResNet18}                                                                   \\ \hline
Temperature ($T$) & \multicolumn{1}{c|}{Accuracy (\%)}          & \multicolumn{1}{c|}{Loss}           & Num. of Round       & \multicolumn{1}{c|}{Accuracy (\%)}          & \multicolumn{1}{c|}{Loss}           & Num. of Round        & \multicolumn{1}{c|}{Accuracy (\%)}          & \multicolumn{1}{c|}{Loss}           & Num. of Round        \\ \hline
$T=4.0$           & \multicolumn{1}{c|}{73.60 (-5.06)}          & \multicolumn{1}{c|}{0.893}          & 299 (0.44×)         & \multicolumn{1}{c|}{37.58 (-3.76)}          & \multicolumn{1}{c|}{1.724}          & 299 (0.84×)          & \multicolumn{1}{c|}{28.05 (-0.20)}          & \multicolumn{1}{c|}{2.984}          & 288 (0.69×)          \\ \hline
$T=2.0$           & \multicolumn{1}{c|}{77.20 (-1.46)}          & \multicolumn{1}{c|}{0.749}          & 205 (0.64×)         & \multicolumn{1}{c|}{40.35 (-0.99)}          & \multicolumn{1}{c|}{1.638}          & 270 (0.93×)          & \multicolumn{1}{c|}{27.89 (-0.36)}          & \multicolumn{1}{c|}{2.918}          & 288 (0.69×)          \\ \hline
$T=1.0$           & \multicolumn{1}{c|}{78.66 (0.00)}           & \multicolumn{1}{c|}{0.684}          & 132 (1.00×)         & \multicolumn{1}{c|}{41.34 (0.00)}           & \multicolumn{1}{c|}{1.594}          & 251 (1.00×)          & \multicolumn{1}{c|}{28.25 (0.00)}           & \multicolumn{1}{c|}{2.918}          & 200 (1×)             \\ \hline
$T=0.5$           & \multicolumn{1}{c|}{79.78 (+1.13)}          & \multicolumn{1}{c|}{0.654}          & 88 (1.50×)          & \multicolumn{1}{c|}{41.52 (+0.18)}          & \multicolumn{1}{c|}{1.551}          & 161 (1.56×)          & \multicolumn{1}{c|}{29.77 (+1.52)}          & \multicolumn{1}{c|}{\textbf{2.875}} & 178 (1.12×)          \\ \hline
$T=0.25$          & \multicolumn{1}{c|}{79.50 (+0.85)}          & \multicolumn{1}{l|}{0.650}          & 57 (2.32×)          & \multicolumn{1}{c|}{43.41 (+2.07)}          & \multicolumn{1}{c|}{1.532}          & 132 (1.90×)          & \multicolumn{1}{c|}{\textbf{30.63 (+2.38)}} & \multicolumn{1}{c|}{2.997}          & \textbf{152 (1.32×)} \\ \hline
$T=0.05$          & \multicolumn{1}{c|}{\textbf{80.86 (+2.20)}} & \multicolumn{1}{c|}{\textbf{0.643}} & \textbf{22 (6.00×)} & \multicolumn{1}{c|}{\textbf{44.71 (+3.37)}} & \multicolumn{1}{c|}{\textbf{1.528}} & \textbf{106 (2.37×)} & \multicolumn{1}{c|}{27.05 (-2.72)}          & \multicolumn{1}{c|}{4.228}          & 239 (0.74×)          \\ \hline
\end{tabular}
\end{adjustbox}
\caption{Average classification accuracy, model test loss, and number of federated learning training rounds needed to achieve the maximum accuracy of the $T$=4 cases for each dataset-model combination. 
%Quantitative results suggest the effectiveness of exploiting low temperatures for model training in federated learning both in terms of model accuracy and training efficiency.
}
\label{tab:overall}
\end{table*}

% \rev{Finally, in the final columns for each test case in Table~\ref{tab:overall}, we present the number of federated learning training rounds needed to achieve a preset target accuracy for different temperature values used for model training. This quantity serves as our notion of ``convergence speed'' and provides an understanding of how efficiently the local training processes operate under different temperatures. Specifically, we configure the target accuracy to be the maximum test accuracy achieved within 300 training rounds with $T=4$, which gives a single, model-agnostic threshold across all configurations. As an example, for the FEMNIST-DNN and CIFAR10-CNN cases, the maximum accuracy with $T=4$ is reached at the 299$^{\text{th}}$ round, while CIFAR100-ResNet18 attains its $T=4$ maximum at after 288 rounds. We then compare the number of rounds needed for each temperature setting to reach the corresponding $T=4$ target, and present normalized values in Table~\ref{tab:overall} with respect to the standard setting $T=1.0$.}

\rev{Finally, in the final columns for each test case in Table~\ref{tab:overall}, we present the number of federated learning training rounds needed to achieve a preset target accuracy for different temperature values used for model training. This quantity serves as our notion of ``convergence speed'' and provides an understanding of how efficiently the local training processes operate under different temperatures. Specifically, we define the target accuracy as follows: for each dataset–model configuration, we first run the training with $T=4$ for 300 rounds, and record the highest test accuracy that the global model (after server aggregation) achieves within those 300 rounds. The highest accuracy obtained by the global model trained with $T=4$ serves as the target accuracy for all temperature settings. For example, in the FEMNIST-DNN and CIFAR10-CNN experiments, the global model trained with $T=4$ reaches its maximum accuracy at round 299, whereas for CIFAR100-ResNet18, the corresponding maximum occurs at round 288.}

\rev{Next, for each temperature value $T \in \{0.05, 0.25, 0.5, 1.0, 2.0, 4.0\}$, we measure how many rounds the global model needs to reach this $T=4$ target accuracy. To facilitate comparison across temperatures, Table~\ref{tab:overall} reports the \textit{convergence speed} for each temperature, expressed as a speed-up factor normalized by the number of rounds required by the standard setting $T=1.0$. Thus, a value of 1.0$\times$ corresponds to the speed of training with $T=1.0$, values larger than 1.0$\times$ indicate faster convergence (fewer rounds needed), and values smaller than 1.0$\times$ indicate slower convergence.}

\rev{Results in Table~\ref{tab:overall} suggest that lower temperatures substantially reduce the number of rounds required to reach this target accuracy. For FEMNIST-DNN, the standard configuration $T=1.0$ requires 132 rounds to reach the $T=4$ target, whereas $T=0.05$ achieves the same target in only 22 rounds, corresponding to an approximately 6$\times$ speedup. For CIFAR10-CNN, $T=1.0$ needs 251 rounds, while $T=0.05$ reaches the target accuracy in 106 rounds (about 2.37$\times$ faster). For CIFAR100-ResNet18, the most effective low-temperature setting is $T=0.25$, which reaches the target accuracy in 152 rounds compared to 200 rounds for $T=1.0$ ($\sim$1.32$\times$ faster). These concrete gains reflect the aggressive training dynamics induced by low-temperature scaling, with high-temperature training tending to conservatively absorb information, whereas lower temperatures amplify gradients and drive larger, more influential model updates when given the same local data. In federated settings where each client holds a relatively small dataset and must effectively convey its local knowledge through aggregation, this aggressiveness allows \name{} to reach high-performing global models in significantly fewer rounds. Additional experiments with varying training hyperparameters (e.g., numbers of participating clients, local batch size, and local epochs), reported in the Appendix, show that this convergence-speed advantage of low-temperature training persists as the training hyperparameters change, even though the exact speedup factor varies with the degree of data heterogeneity and local update configuration.}

% Finally, in the last columns for each test case in Table~\ref{tab:overall} we present the number of federated learning training rounds needed to achieve a preset target accuracy for different temperature values used for model training. This number provides an understanding of how efficient the local training processes are when different temperature values are applied. Specifically, in this experiment, we configure the target accuracy to be the maximum accuracy achieved within 300 training rounds with $T$=4. For example, for the FEMNIST-DNN and CIFAR10-CNN cases, the maximum accuracy for $T$=4 was achieved at the 299$^{th}$ round, while CIFAR100-ResNet18 achieved such accuracy after 288 rounds. 

% From the results in Table~\ref{tab:overall} we can notice that with $T$=0.05, the FEMNIST-DNN case achieves the target accuracy after only 22 training rounds. We see similar trends of low temperatures positively affecting the training efficiency throughout our results. This is from the general aggressiveness of training behaviors when applying lower temperature training. Given a single training sample, while high temperature-based training conservatively absorbs the available information, lower temperatures are more progressive in making model changes. Given the small datasets used for local training and the need to effectively deliver its knowledge to the server for aggregation, we see \name{} as an effective way to train client-side models in the federated learning process.

\subsection{Model stability of low-temperature training during federated learning aggregation}
\label{subsec:stability}
\begin{figure}[t!]        
        \centering
        \includegraphics[width=.75\linewidth]{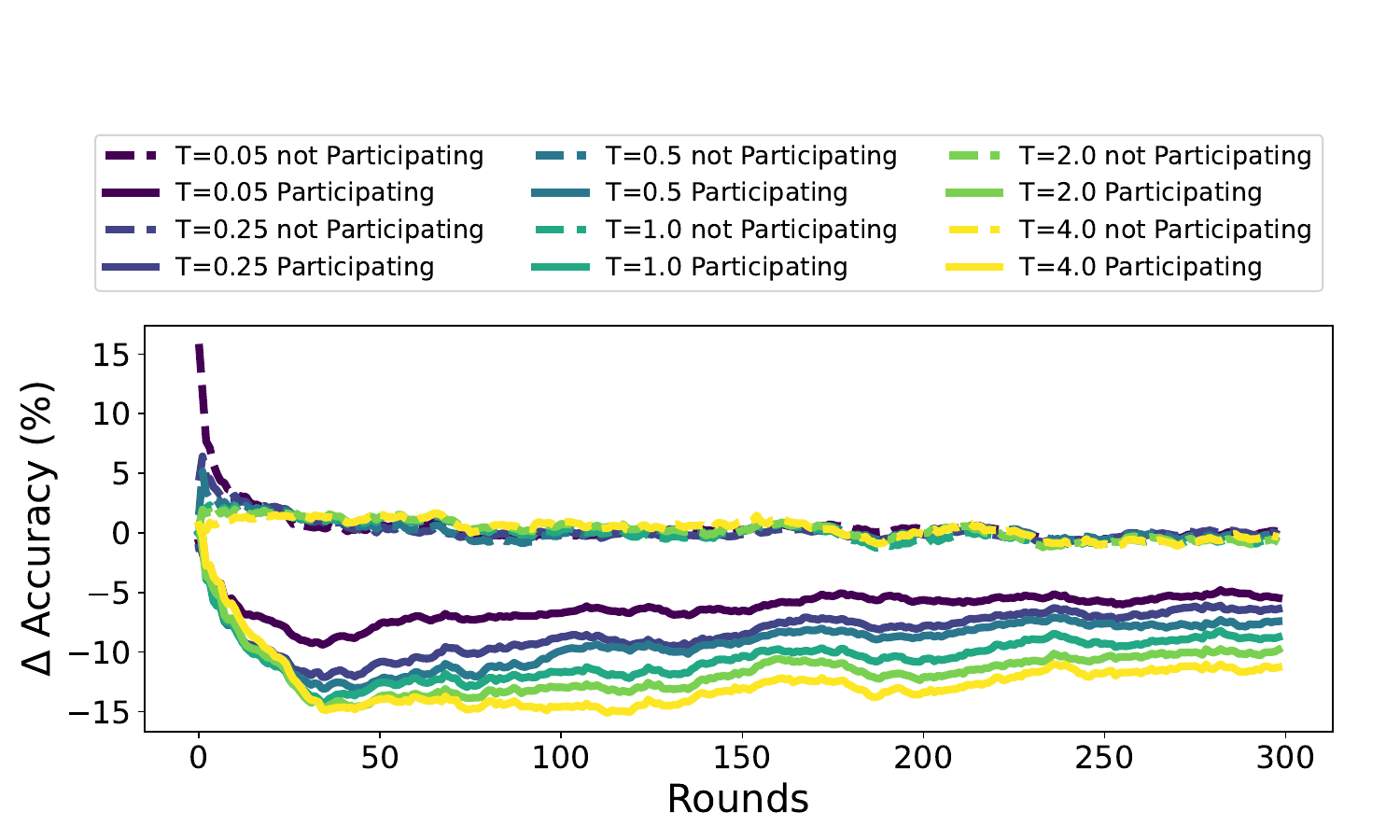}
        \caption{Accuracy drop observed before and after server-side model aggregation. The aggregation process of client models trained with lower temperatures better retains the individual knowledge.}
        \label{fig:trained-acc}
\end{figure}
Figure~\ref{fig:trained-acc} plots the changes in accuracy observed for both participating and non-participating clients at each federated learning round. Specifically, once the model is aggregated, each client computes its accuracy with both the pre-update and the post-update model and computes their difference. Note that pre-update models for clients participating in the round would have updated their local models (very recently) with their own data.

As discussed in Section~\ref{subsec:ea}, low temperatures expedite local training even with a small amount of training data. Nevertheless, the effect of low-temperature training on the model generalization and aggregation stability of federated learning has not yet been thoroughly studied. To inspect the stability of federated learning aggregation for models trained with low temperatures, we measured the performance change of each client-side model before and after the server-side aggregation round. % 2

Thus, the pre-model performance will be high. However, since each model is trained aggressively over heterogeneous data, the aggregation process could puddle up this knowledge and the aggregated model may lose information. Ideally, with their local information properly aggregated, the performance drop will be minimal. Non-participating clients on the other hand, would hope to have an improved updated model of the federated learning process. 

As Figure~\ref{fig:trained-acc} shows the models trained with low temperatures retain the knowledge better than the medium and hot models for participating clients (solid plots groups at the bottom). In the early rounds, the models with high temperatures show less performance drop, but the fast learning speed of the low-temperature model amortizes the performance drop, which results in faster learning of the aggregated model. Furthermore, we can also notice that the aggregation process does not interfere with the performance of non-participating clients (grouped on top), implying that federated learning with low-temperature training preserves the generalization power.

\subsection{Impact of Dataset Dispersion and Federated Learning on \name{}}
We now examine the impact of \name{} on different model training configurations. Specifically, using the CIFAR10-CNN test case, we examine two additional configurations: (i) centralized learning with i.i.d. data and (ii) federated learning with i.i.d. data. Note that the federated learning network is identical to our previous experiments except for the dataset distribution and was tested for 300 rounds. By comparing these results with the non-i.i.d. federated learning results, we target to understand the impact of \name{} on federated learning and dataset disparity in greater detail. % 1

As Table~\ref{tab:iid-effect} shows, applying \name{}, thus a lower $T$ does not induce a positive impact on the performance of centralized learning scenarios. In fact, for centralized learning, controlling $T$ for the training process does not show a positive impact. We see the same phenomena for the federated learning system with i.i.d. samples as well, suggesting that the performance enhancements we see in Table~\ref{tab:overall} are not an effect of the federated learning process itself. % 2

\begin{table}[t!]
    \centering
        \begin{adjustbox}{width=.7\linewidth,center}
        \begin{tabular}{|c|cc|cc|cc|}
        \hline
                          & \multicolumn{2}{c|}{Centralized - i.i.d.}                                                                        & \multicolumn{2}{c|}{Federated - i.i.d.}                                                                         & \multicolumn{2}{c|}{Federated - non-i.i.d.}                                                                   \\ \hline
        $T$ & \multicolumn{1}{c|}{Acc. (\%)}          	& \multicolumn{1}{c|}{Loss} & \multicolumn{1}{c|}{Acc. (\%)}    & \multicolumn{1}{c|}{Loss}     & \multicolumn{1}{c|}{Acc. (\%)} & \multicolumn{1}{c|}{Loss}           \\ \hline
        4.0           & \multicolumn{1}{c|}{58.05}          & \multicolumn{1}{c|}{1.152}    & \multicolumn{1}{c|}{48.22}          	& \multicolumn{1}{c|}{1.473}        & \multicolumn{1}{c|}{37.58}     & \multicolumn{1}{c|}{1.724}          \\ \hline
        2.0           & \multicolumn{1}{c|}{63.27}          & \multicolumn{1}{c|}{1.041}    & \multicolumn{1}{c|}{51.90}          	& \multicolumn{1}{c|}{1.403}        & \multicolumn{1}{c|}{40.35}     & \multicolumn{1}{c|}{1.638}          \\ \hline
        1.0           & \multicolumn{1}{c|}{65.18}          & \multicolumn{1}{c|}{1.044}    & \multicolumn{1}{c|}{54.66}            & \multicolumn{1}{c|}{1.325}        & \multicolumn{1}{c|}{41.34}     & \multicolumn{1}{c|}{1.594}          \\ \hline
        0.5           & \multicolumn{1}{c|}{64.19}          & \multicolumn{1}{c|}{1.030} 	& \multicolumn{1}{c|}{53.19}            & \multicolumn{1}{c|}{1.320}    	& \multicolumn{1}{c|}{41.52}     & \multicolumn{1}{c|}{1.551}          \\ \hline
        0.25          & \multicolumn{1}{c|}{63.88}          & \multicolumn{1}{c|}{1.092}    & \multicolumn{1}{c|}{54.64}            & \multicolumn{1}{c|}{1.265}    	& \multicolumn{1}{c|}{43.41} 	 & \multicolumn{1}{c|}{1.532} \\ \hline
        0.05          & \multicolumn{1}{c|}{60.98} 			& \multicolumn{1}{c|}{1.227}  	& \multicolumn{1}{c|}{47.60} 			& \multicolumn{1}{c|}{1.701} 		& \multicolumn{1}{c|}{44.71}     & \multicolumn{1}{c|}{1.528}          \\ \hline
        \end{tabular}
        \end{adjustbox}
        \caption{Average inference accuracy and model test loss for CIFAR10-CNN with different configurations. The impact of \name{} is prominent for federated learning scenarios with non-i.i.d. datasets, whereas its impact is less than else-wise.}
        \label{tab:iid-effect}
        \vspace{-1ex}
\end{table}

\begin{figure}[t!]
    \centering
    \includegraphics[width=.6\linewidth]{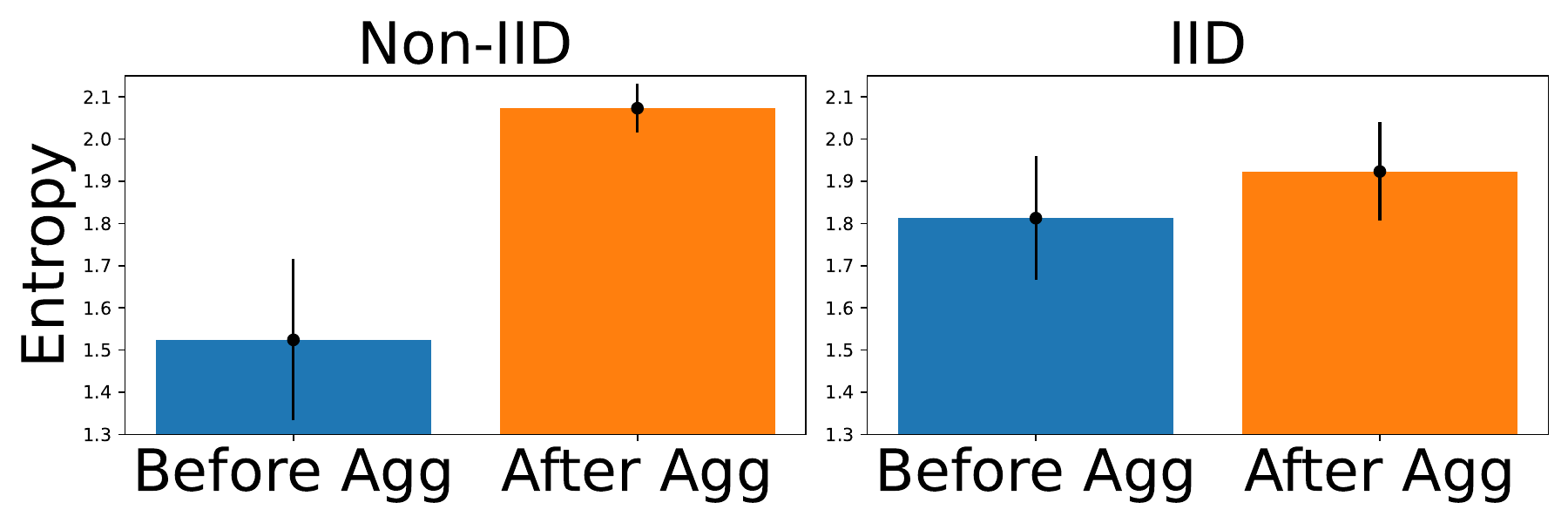}
    \caption{\rev{Average entropy before aggregation (Before Agg) and after aggregation (After Agg) for different data disparity configurations.}}
    \vspace{-1ex}
    \label{fig:entropy}
\end{figure}

In federated learning, where non-i.i.d. environments are common, the loss of information during model aggregation can be a significant factor contributing to performance degradation. Figure~\ref{fig:entropy} shows the entropy of the model before and after server-side aggregation in two different settings: non-i.i.d. and i.i.d. data distribution. In the federated learning setup, the model's entropy increases after aggregation, indicating that the probability distribution produced by the model has become \textit{smoother}, with the smoothing effect more pronounced in non-i.i.d. environments. By applying a lower $T$, we carefully argue that the smoothed probability distribution can be sharpened, which appears to enhance both convergence and final performance. 

We do point out that holding i.i.d. datasets in federated learning will show overall higher performance compared to non-i.i.d. configurations (regardless of $T$), but is not practical to assume when deploying federated learning systems. Rather, as we discussed, the impact of exploiting low training temperatures is beneficial when dealing with non-i.i.d. datasets.

\begin{figure}[t!]
    \centering
    \subfigure[\vspace{-2ex}Severe heterogeneity. ($\alpha$=0.1)]{
    \includegraphics[width=.6\linewidth]{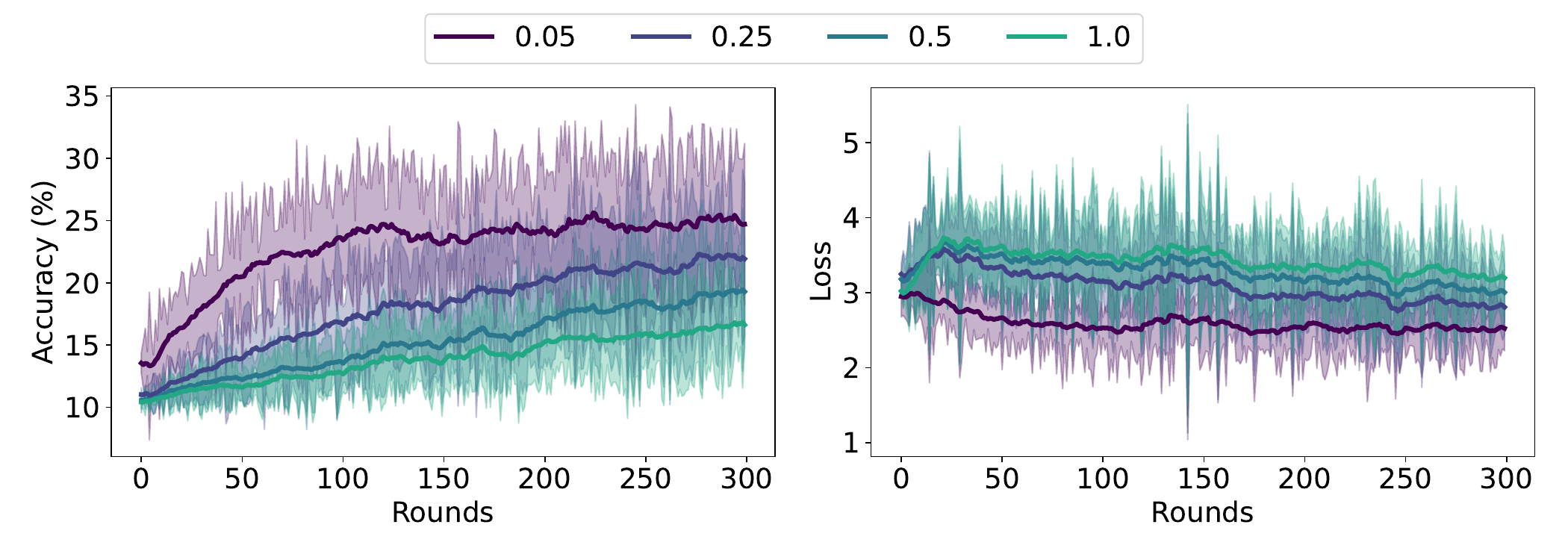}
    }
    \vspace{-2ex}
    \subfigure[\vspace{-2ex}Moderate heterogeneity. ($\alpha$=0.5)]{
    \includegraphics[width=.6\linewidth]{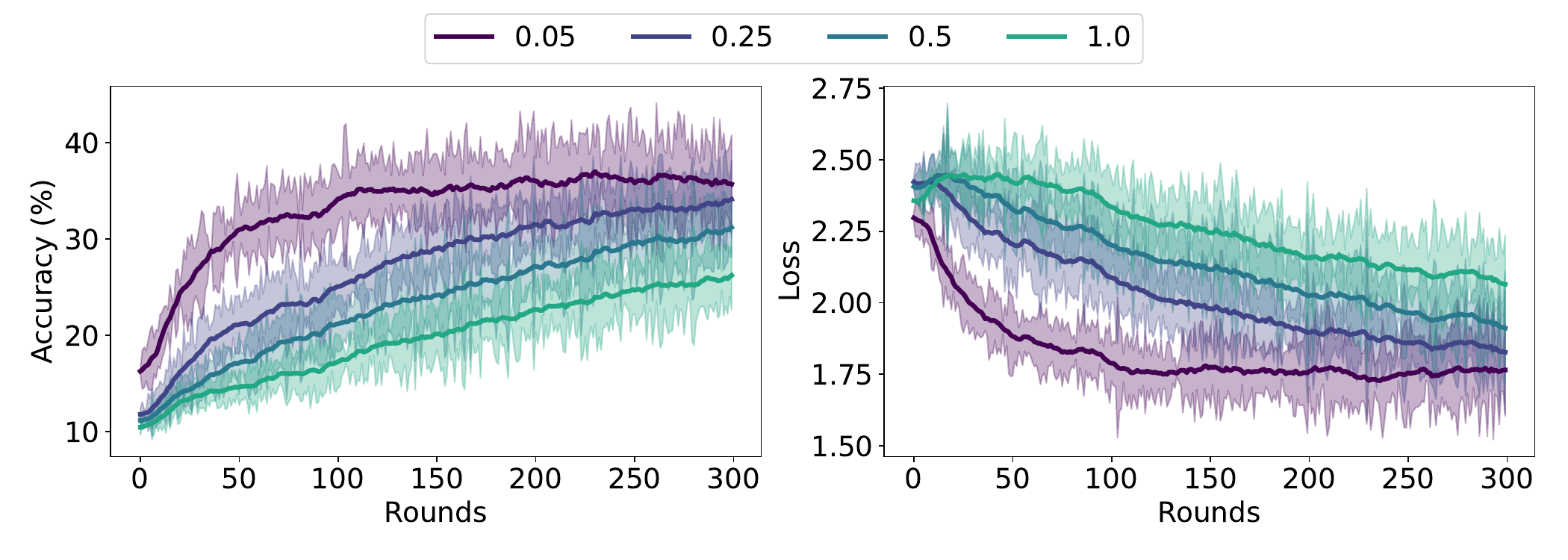}
    }
    \vspace{-2ex}
    \subfigure[\vspace{-2ex}Weak heterogeneity. ($\alpha$=1.0)]{
    \includegraphics[width=.6\linewidth]{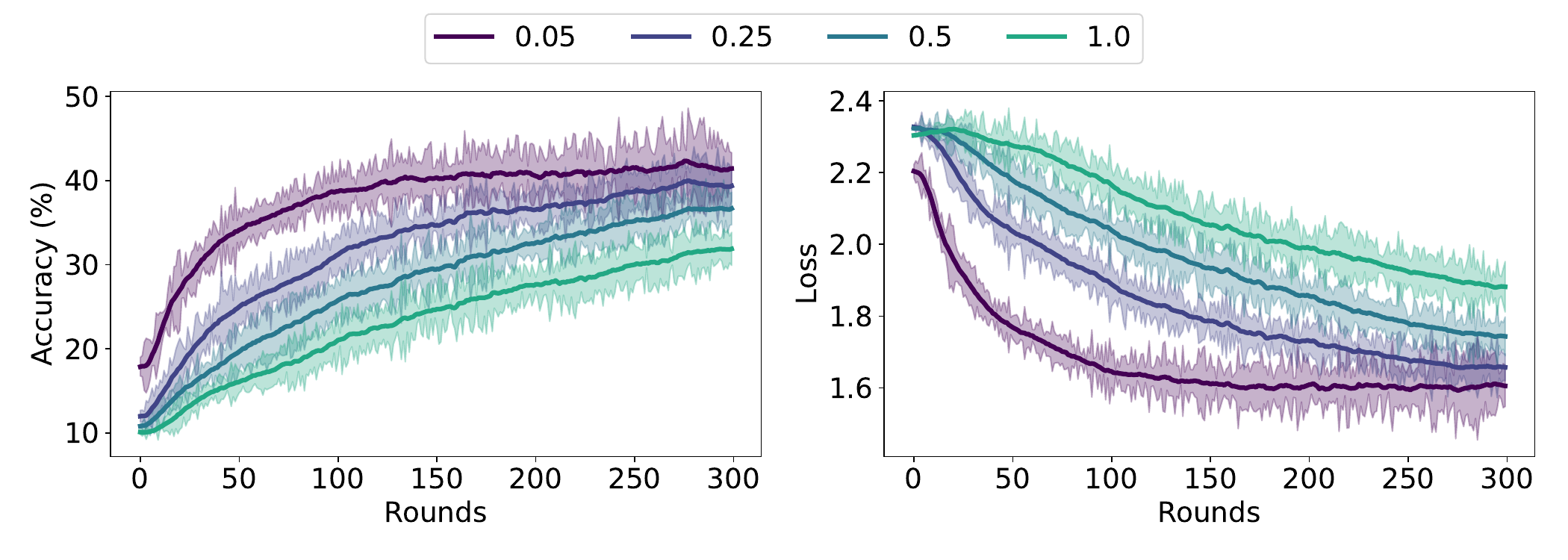}
    }
    \caption{Test accuracy and test loss over federated learning rounds with varying the degree of data disparity and the training temperature on the CIFAR10 dataset. Relatively moderate data disparity shows better training quality with stability. The shaded areas indicate the standard deviation across different runs.}
    \label{fig:noniid}
\end{figure}

To assess the robustness of \name{} under varying degrees of data heterogeneity, we partition the dataset using a Dirichlet distribution~\cite{hsu2019measuring}. A lower Dirichlet parameter \(\alpha\) corresponds to a higher degree of heterogeneity, while a higher \(\alpha\) indicates a more homogeneous setting. We consider \(\alpha = \{0.1, 0.5, 1.0\}\) to represent severe, moderate, and weak heterogeneity levels, respectively. Figure~\ref{fig:noniid} presents the global model accuracy over 300 training rounds. As depicted in the figure, a lower temperature consistently leads to superior model performance across all settings.  

\subsection{Feature Space Similarity}
\label{sec:featuresim}
\begin{figure*}[t!]
    \centering
    \includegraphics[width=\linewidth]{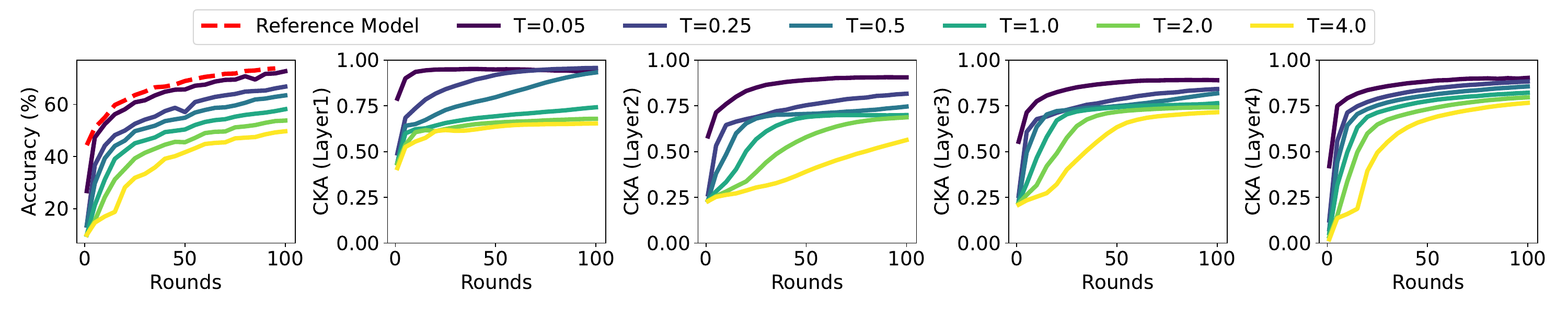}
    \vspace{-3ex}
    \caption{Test accuracy of the global model and feature space similarity in each layer between reference model trained with i.i.d. dataset and the global model trained with non-i.i.d. dataset with different temperatures applied during the local training.}
    \label{fig:cka_reference}
\end{figure*}

\begin{figure}[t!]
    \centering
    \subfigure[$T$=0.05]{
    % \includegraphics[width=\linewidth, trim={7cm 0cm 7cm 1cm}, clip]{CKA_0.05.pdf}
    % }\\
    \includegraphics[width=\linewidth]{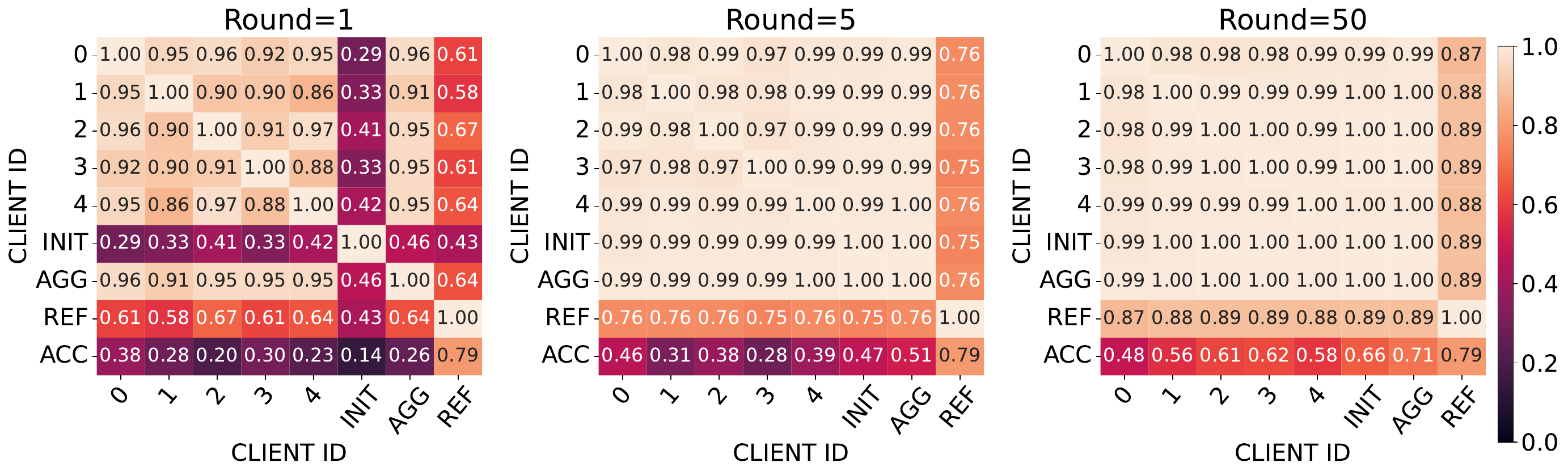}
    }\\
    % \subfigure[$T$=1.0]{
    % \includegraphics[width=\linewidth, trim={7cm 0cm 7cm 1cm}, clip]{CKA_1.0.pdf}
    % }
    \subfigure[$T$=1.0]{
    \includegraphics[width=\linewidth]{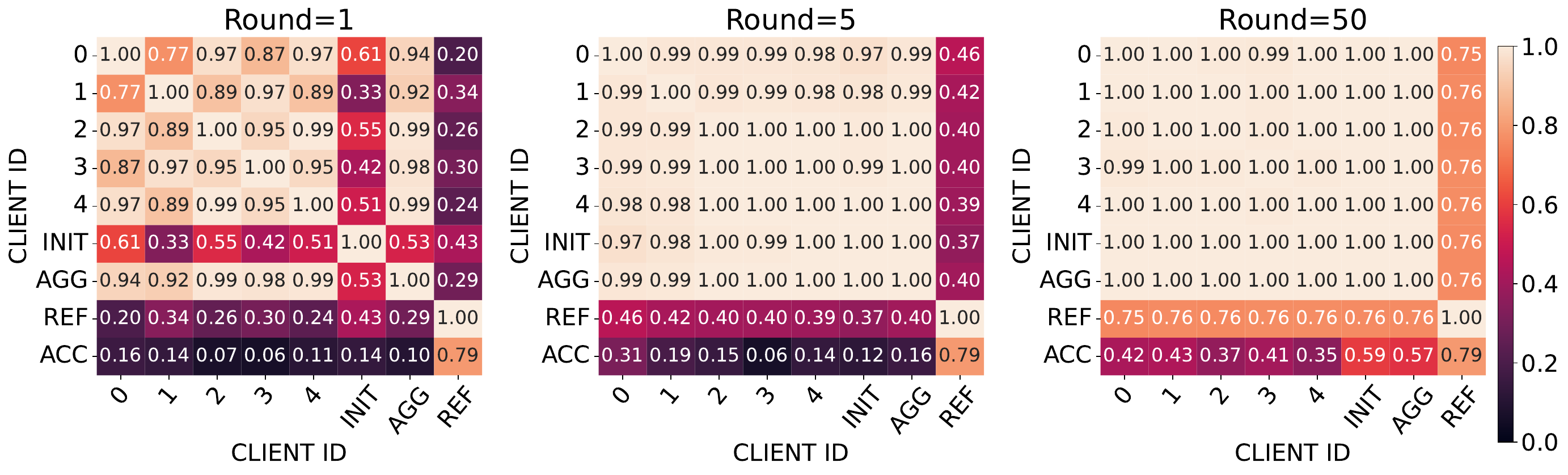}
    }
    \caption{Client-wise feature space similarity matrix measured in penultimate layers by using CKA and inference accuracy in different temperatures (0.05, 1.0) and federated rounds (1, 5, 50). (INIT: Initial dispatched model in the given round, AGG: Aggregated global model, REF: Reference model, ACC: Accuracy)}
    \label{fig:CKA_simmat}
\end{figure}

To better understand the mechanism and rationale behind the performance of \name{}, we analyze the similarity of the feature space between global (aggregated) models trained with different temperatures against a reference model trained in an \textit{i.i.d.} data configuration. Specifically, we employ the 
Centered Kernel Alignment (CKA)~\cite{kornblith2019similarity} for computing the similarity of feature spaces, a widely used approach in evaluating model divergence in federated learning~\cite{collins2022fedavg, luo2021no, shin2024effective}. 
CKA is a similarity measure between two different neural networks by measuring the similarity of the network representation for the given input data. Let $X$ and $Y$ be the features generated from two different neural networks, we can calculate the CKA by Eq~\ref{eq:cka}. As we are assuming a federated learning scenario, the clients share the same neural network architecture. Thus, the dimensions of the two features from each corresponding layer are identical, allowing for a direct comparison to be possible (detailed discussion provided in Appendix). % \textcolor{red}{\sout{(detailed discussion provided in Appendix)}}
We note that in our experiments, we exploited the linear CKA to measure the similarity of the feature spaces between two neural networks. 
\begin{equation}
    \begin{split}
    CKA(X,Y) &= \frac{||Y^{T}X||^{2}_{F}}{||X^{T}X||_{F}||Y^{T}Y||_{F}}\\
                       % &= \frac{\sum^{p1}_{i=1}\sum^{p2}_{j=1}\lambda^{i}_{X}\lambda^{j}_{Y}<\mathbf{u}^{i}_{X},\mathbf{u}^{j}_{Y}>^{2}}{\sqrt{\sum^{p1}_{i=1}(\lambda^{i}_{X})^{2}}\sqrt{\sum^{p2}_{j=1}(\lambda^{j}_{Y})^{2}}}
    \end{split}
    \label{eq:cka}
\end{equation}
Note that a limitation of CKA measures is that they only report network similarity without considering training quality. Thus, we use a well-trained reference model from an i.i.d. environment to observe CKA variations of the penultimate and classification layers between the global models and the reference model.

We note that the reference model is trained with $T$=1, while comparison models (e.g., global models from federated learning) are trained in non-i.i.d. environments with different $T$. In the experiments, we used a 2-layered CNN and the CIFAR-10 dataset as in \rev{Section~\ref{sec:exp_setup}}, with the dataset distributed among 10 clients using a Dirichlet distribution with $\alpha=0.5$, and all clients participated in the federated training process. % \textcolor{red}{\sout{For detailed discussions on the experimental setup and CKA, please refer to Appendix~\ref{sec:CKA}.}}

\begin{table}[t!]
\centering
\begin{adjustbox}{width=0.8\linewidth,center}
\begin{tabular}{|c|c|c|c|c|c|}
\hline
Temperature ($T$)                                                            & Accuracy (\%)  & Layer 1 CKA     & Layer 2 CKA     & Layer 3 CKA     & Layer 4 CKA     \\ \hline
4.00                                                                         & 49.72          & 0.6539          & 0.5632          & 0.7149          & 0.7661          \\ \hline
2.00                                                                         & 53.82          & 0.6792          & 0.6872          & 0.7417          & 0.7963          \\ \hline
1.00                                                                         & 58.19          & 0.7419          & 0.6997          & 0.7644          & 0.8214          \\ \hline
0.50                                                                         & 63.48          & 0.9322          & 0.7459          & 0.8181          & 0.8559          \\ \hline
0.25                                                                         & 66.85          & \textbf{0.9569} & 0.8161          & 0.8418          & 0.8830          \\ \hline
0.05                                                                         & \textbf{72.76} & 0.9390          & \textbf{0.9059} & \textbf{0.8898} & \textbf{0.9026} \\ \hline
\begin{tabular}[c]{@{}c@{}}Reference Model\\ ($T$=1.00, i.i.d.)\end{tabular} & 73.86          & 1.0000          & 1.0000          & 1.0000          & 1.0000          \\ \hline
\end{tabular}
\end{adjustbox}
\caption{Final accuracy and CKA values of 2 layer CNN measured on the CIFAR10 dataset in Figure~\ref{fig:cka_reference}.}
\label{tab:cka_reference}
\end{table}

As the accuracy and CKA plots for different layers in Figure~\ref{fig:cka_reference} and Table~\ref{tab:cka_reference} show, introducing the non-i.i.d. environment itself drops the model accuracy (73.86\% $\xrightarrow{}$ 58.19\%) at $T$=1, with the CKA of the penultimate/classification layers between the two models remaining at 0.7644/0.8214 after 100 rounds. Nevertheless, with $T$=0.05, despite the dataset disparity, the global model achieved a comparable accuracy of 72.76\% with relatively higher CKA values (0.8898/0.9026) in both the penultimate and classification layers. We point out that improved gradient propagation to input layer via applying low temperatures help models to learn similar local features (i.e., layers 1, 2), affecting the following layers (i.e., layers 3, 4) in the early stage of the learning process. As a result, applying low temperatures during local training effectively improves the overall accuracy performance in a non-i.i.d. data distribution setting and helps models learn similar feature spaces compared to the ideal i.i.d. data configuration.

We now aim to better understand how local model updates progress with different $T$. For brevity of the paper, we only report the CKA at the penultimate layer. Specifically, Figure~\ref{fig:CKA_simmat} plots the model accuracy (ACC) and measured CKA at the penultimate layer of a client's local model with (i) the initially distributed model (INIT), (ii) the aggregated model (AGG), (iii) the ideal reference model trained with i.i.d\rev{.} data distribution (REF), and (iv) other client models while $T$=0.05 and 1.0. In Round 1, clients' local models trained with $T$=1 show high feature space similarity against each other while failing to learn the features of the ideal reference model. With $T$=0.05, the CKA between local models is relatively low while showing high similarity with the reference model. After five rounds (and beyond), local models trained with both $T$=1.0 and $T$=0.05 showed high similarity amongst themselves and the initial/aggregated models, but the $T$=0.05 case continuously shows higher similarity with REF and reports higher accuracy. Hence, we argue that adapting low temperatures allows local models to learn effective and diverse features to converge the federated learning process quickly.

\subsection{\name{} with Advanced FL Schemes}
\label{sec:advschemes}
\begin{figure}[t!]
    \centering
    \subfigure[FedProx]{
    \includegraphics[width=.7\linewidth]{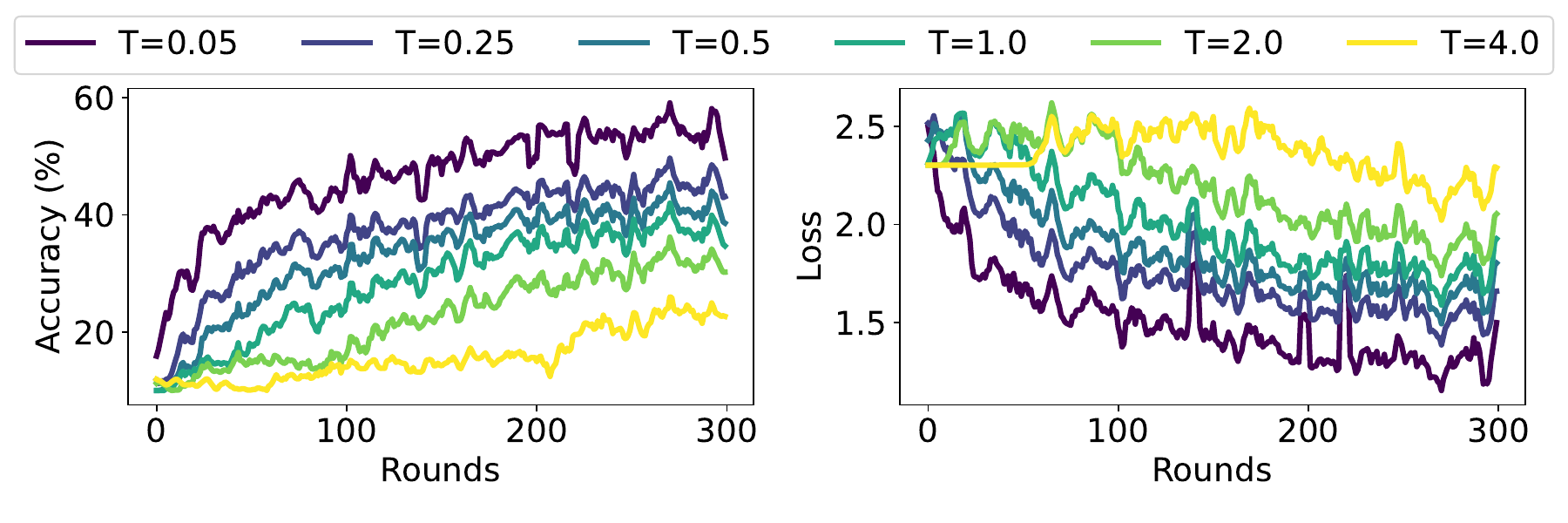}
    }
    \subfigure[SCAFFOLD]{
    \includegraphics[width=.7\linewidth]{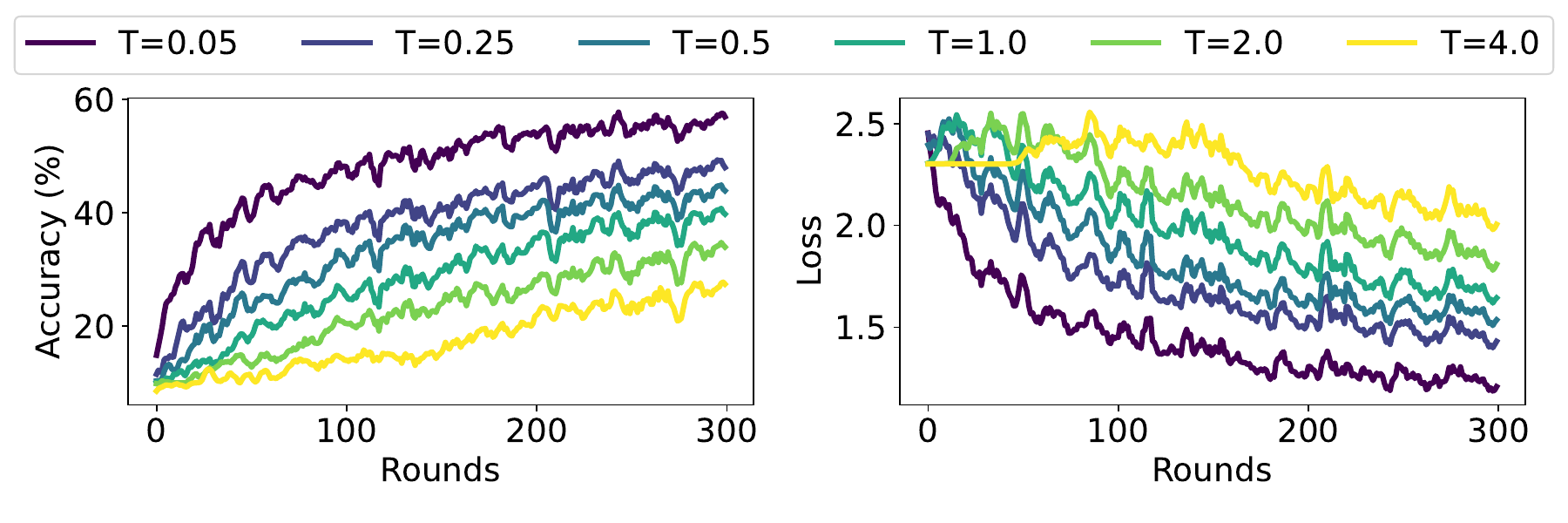}
    }
    \subfigure[\rev{FedBN}]{
    \includegraphics[width=.7\linewidth]{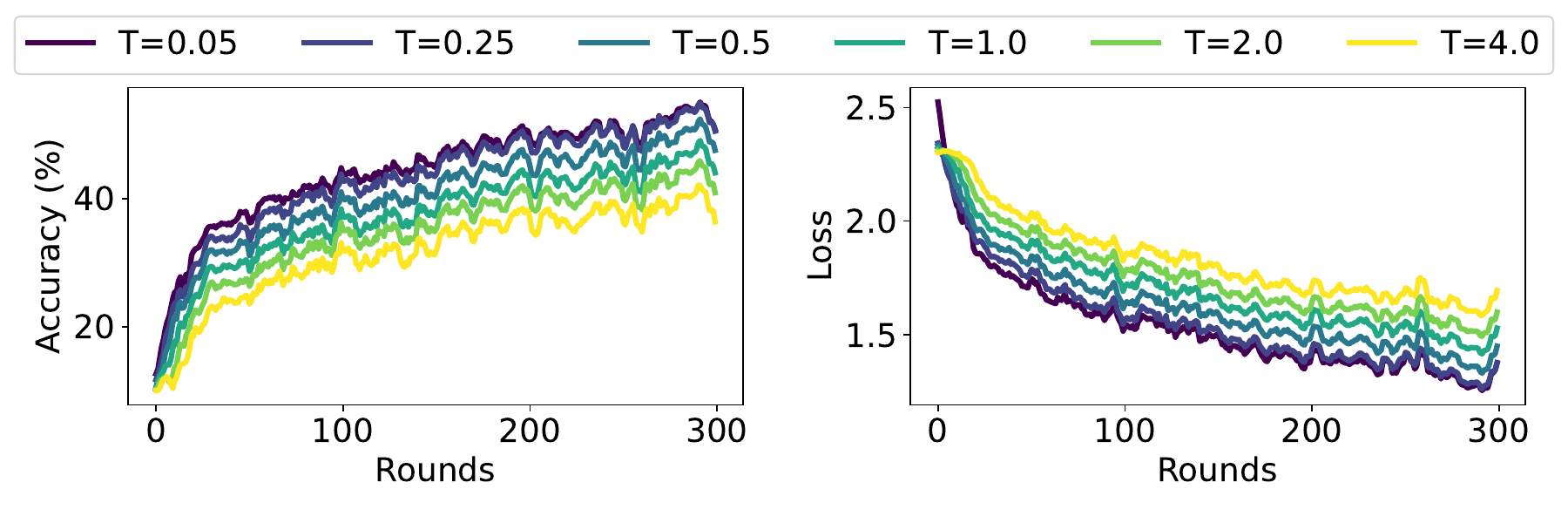}
    }
    \caption{Test accuracy and test loss on the CIFAR10 dataset with varying training temperatures with advanced federated learning schemes. Overall, models trained with lower temperatures showed superior performance.}
    \label{fig:fedprox_scaffold}
\end{figure}

\begin{table*}[t]
\begin{adjustbox}{width=\linewidth,center}
\begin{tabular}{|c|ccc|ccc|ccc|}
\hline
                  & \multicolumn{3}{c|}{SCAFFOLD}
                  & \multicolumn{3}{c|}{FedProx}
                  & \multicolumn{3}{c|}{FedBN} 
                  \\ \hline
Temperature ($T$)
& \multicolumn{1}{c|}{Accuracy (\%)} 
& \multicolumn{1}{c|}{Loss} 
& Num. of Round
& \multicolumn{1}{c|}{Accuracy (\%)} 
& \multicolumn{1}{c|}{Loss} 
& Num. of Round
& \multicolumn{1}{c|}{Accuracy (\%)}
& \multicolumn{1}{c|}{Loss}
& Num. of Round
\\ \hline

$T=4.0$
& \multicolumn{1}{c|}{32.32 (-11.60)} & \multicolumn{1}{c|}{1.935} & 300 (0.39×)
& \multicolumn{1}{c|}{29.95 (-13.41)} & \multicolumn{1}{c|}{1.989} & 300 (0.76×)
& \multicolumn{1}{c|}{43.77 (-6.60)} & \multicolumn{1}{c|}{1.562} & 292 (0.61×)
\\ \hline
$T=2.0$
& \multicolumn{1}{c|}{38.40 (-5.52)} & \multicolumn{1}{c|}{1.729} & 222 (0.55×)
& \multicolumn{1}{c|}{37.42 (-5.94)} & \multicolumn{1}{c|}{1.739} & 300 (0.76×)
& \multicolumn{1}{c|}{47.32 (-3.05)} & \multicolumn{1}{c|}{1.478} & 212 (0.84×)
\\ \hline

$T=1.0$
& \multicolumn{1}{c|}{43.92 (+0.00)} & \multicolumn{1}{c|}{1.562} & 121 (1.00×)
& \multicolumn{1}{c|}{43.36 (+0.00)} & \multicolumn{1}{c|}{1.555} & 227 (1.00×)
& \multicolumn{1}{c|}{50.37 (+0.00)} & \multicolumn{1}{c|}{1.397} & 178 (1.00×)
\\ \hline

$T=0.5$
& \multicolumn{1}{c|}{47.90 (+3.98)} & \multicolumn{1}{c|}{1.464} & 95 (1.27×)
& \multicolumn{1}{c|}{46.99 (+3.63)} & \multicolumn{1}{c|}{1.457} & 202 (1.12×)
& \multicolumn{1}{c|}{53.81 (+3.44)} & \multicolumn{1}{c|}{1.312} & 143 (1.24×)
\\ \hline
$T=0.25$
& \multicolumn{1}{c|}{52.02 (+8.10)} & \multicolumn{1}{c|}{1.324} & 45 (2.69×)
& \multicolumn{1}{c|}{50.78 (+7.42)} & \multicolumn{1}{c|}{1.356} & 143 (1.59×)
& \multicolumn{1}{c|}{55.95 (+5.58)} & \multicolumn{1}{c|}{1.244} & 88 (2.02×)
\\ \hline

$T=0.05$
& \multicolumn{1}{c|}{\textbf{60.10 (+16.18)}} & \multicolumn{1}{c|}{\textbf{1.127}} & \textbf{13 (9.31×)}
& \multicolumn{1}{c|}{\textbf{61.15 (+17.79)}} & \multicolumn{1}{c|}{\textbf{1.106}} & \textbf{49 (4.63×)}
& \multicolumn{1}{c|}{\textbf{56.31 (+5.94)}}  & \multicolumn{1}{c|}{\textbf{1.235}} & \textbf{82 (2.17×)}
\\ \hline
\end{tabular}
\end{adjustbox}
\caption{\rev{Average classification accuracy, model test loss, and number of federated learning training rounds needed to achieve the maximum accuracy of the $T$=4 cases for FedProx, SCAFFOLD, and FedBN.}}
\label{tab:scaffold_fedprox_temperature}
\end{table*}

Finally, as discussed in Section~\ref{sec:proposal}, \name{} is designed as an orthogonal enhancement applicable to a wide range of federated learning algorithms. To further demonstrate this property, we additionally trained the 2-layer CNN with \rev{FedBN~\cite{li2021fedbn}} alongside FedProx~\cite{li2020federated} and SCAFFOLD~\cite{karimireddy2020scaffold}, which represent more advanced optimization schemes compared to naive FedAvg. Figure~\ref{fig:fedprox_scaffold} \rev{and Table~\ref{tab:scaffold_fedprox_temperature}} report the global model accuracy over 300 federated learning rounds across different temperature settings.

\rev{As detailed earlier in Table~\ref{tab:cnn}, the model used in our main experiments do not include batch normalization; thus, we use a variant of the 2-layer CNN with batch normalization when evaluating FedBN. This architectural difference leads to slight numerical deviations compared to other experiments. Nevertheless, the overall trend remains consistent with lower temperatures (0.05, 0.25, 0.5) outperforming the standard or higher temperatures (1.0, 2.0, 4.0) across all three algorithms (i.e., FedProx, SCAFFOLD, and FedBN).} This result further highlights the orthogonality and broad applicability of \name{} across diverse federated optimization strategies.

\subsection{Experiments on Transformer Architecture}
\begin{figure}[t!]
    \centering
    \includegraphics[width=.7\linewidth]{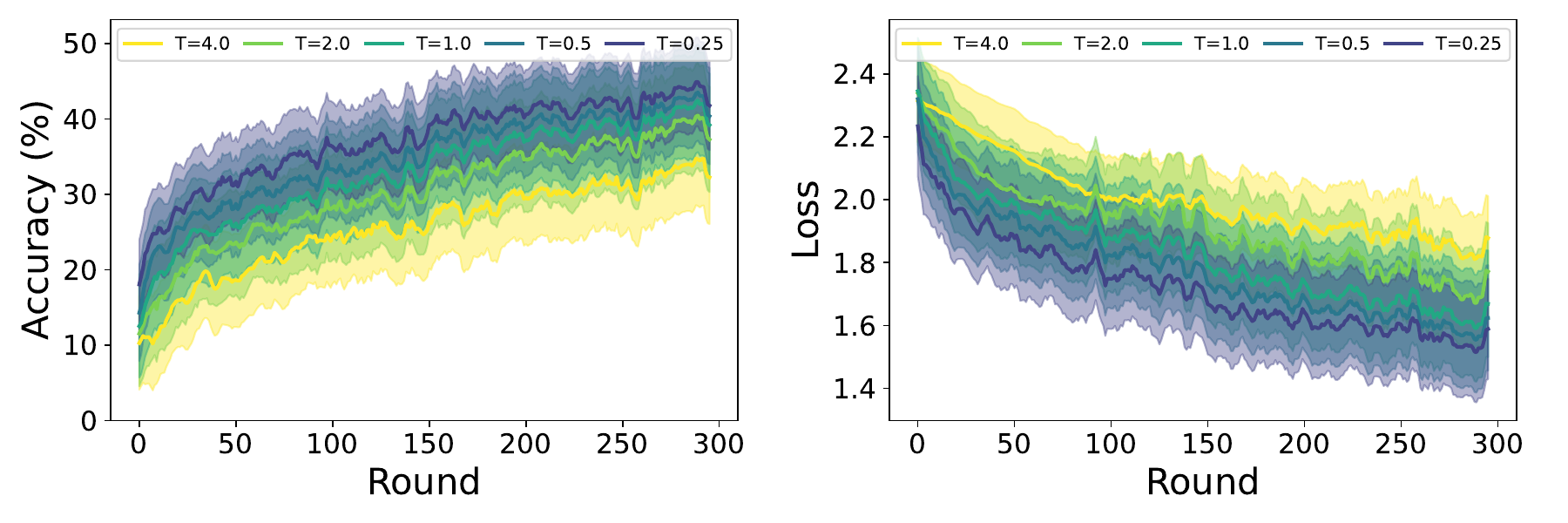}
    \caption{Test accuracy and test loss over federated learning rounds with varying the training temperature on the Vision Transformer and CIFAR10 dataset configuration. The shaded areas indicate the standard deviation across different runs.}
    \vspace{-1ex}
    \label{fig:vit}
\end{figure}
\begin{table}[t!]
\centering
\begin{adjustbox}{width=0.8\linewidth,center}
\begin{tabular}{clc}
\hline
Layer                        & \multicolumn{1}{c}{Details}                                                                                                & Repetition \\ \hline
Patch Embedding              & \begin{tabular}[c]{@{}l@{}}Conv2d(3, 256, k=(4, 4), s=(4, 4))\\ Flatten()\\ Add CLS token + Positional Embedding\end{tabular} & $\times 1$  \\ \hline
Transformer Encoder Layer    & \begin{tabular}[c]{@{}l@{}}Multi‐head Attention (d\_model=256,\\ nhead=4)\\ MLP (dim\_feedforward=1024,\\ activation=GELU,\\ dropout=0.1)\end{tabular} & $\times 3$  \\ \hline
LayerNorm                    & LayerNorm(normalized\_shape=256)                                                                                                & $\times 1$  \\ \hline
Classification Head (FC)     & Linear(256, 10)                                                                                                                 & $\times 1$  \\ \hline
\end{tabular}
\end{adjustbox}
\caption{Model architecture details of the ViT used for CIFAR-10 classification}
\label{tab:vit}
\end{table}

We assess our method’s applicability to transformer architectures by running FedAvg on a Vision Transformer (ViT)\rev{~\cite{dosovitskiy2020image}} trained on CIFAR-10. Five temperature values ($T\in\{0.25, 0.5, 1.0, 2.0, 4.0\}$) are compared to evaluate their effects on convergence speed and final accuracy. Table~\ref{tab:vit} details the ViT configuration used, and Figure~\ref{fig:vit} shows test accuracy and loss over communication rounds. As with our convolutional experiments, lower temperatures accelerate convergence and improve final accuracy in the transformer setting. This behavior arises because ViTs, as with most classification models, use a softmax output layer and cross-entropy loss~\cite{lee2024tazza}, so temperature scaling similarly sharpens probability distributions and amplifies meaningful gradients.

\subsection{Experiments on Time-Series Dataset}
\begin{figure}[t!]
    \centering
    \includegraphics[width=.7\linewidth]{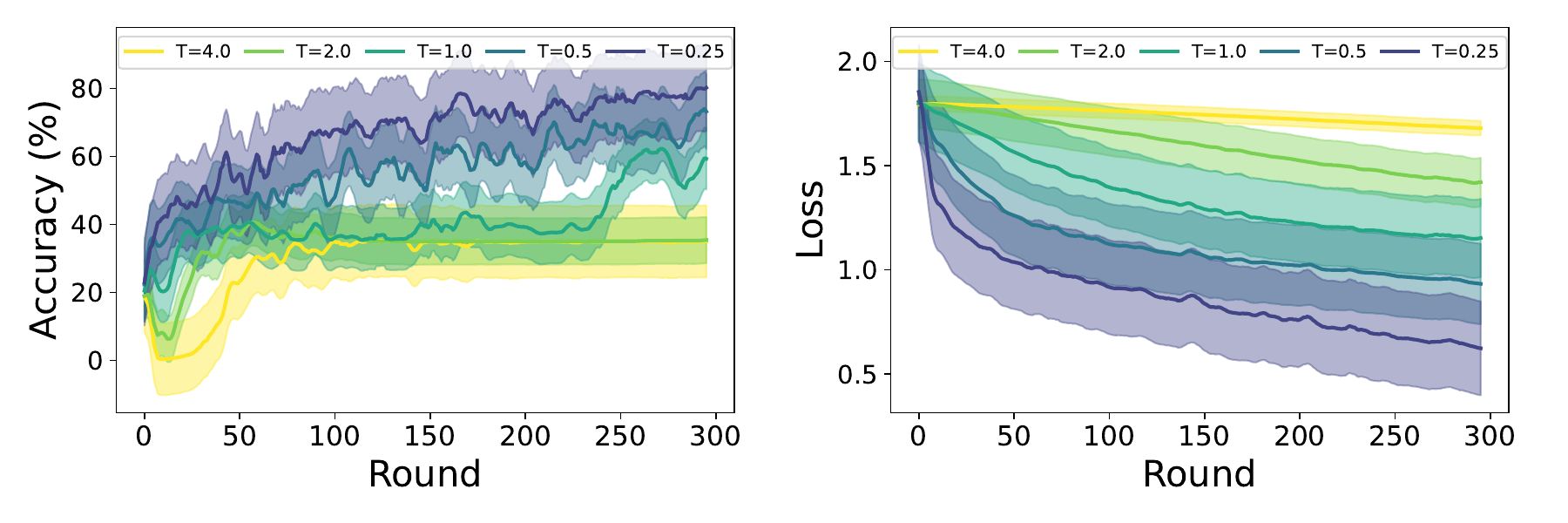}
    \caption{Test accuracy and test loss over federated learning rounds with varying the training temperature on the 1D-CNN and UCI-HAR dataset configuration. The shaded areas indicate the standard deviation across different runs.}
    \vspace{-1ex}
    \label{fig:har}
\end{figure}
\begin{table}[t!]
\centering
\begin{adjustbox}{width=.7\linewidth,center}
\begin{tabular}{clc}
\hline
Layer                    & \multicolumn{1}{c}{Details}                                                                                          & Repetition \\ \hline
Conv Block 1             & \begin{tabular}[c]{@{}l@{}}Conv1d(1, 16, k=7, s=1, p=3)\\ BatchNorm1d(16)\\ ReLU()\\ MaxPool1d(k=2)\end{tabular}       & $\times 1$  \\ \hline
Conv Block 2             & \begin{tabular}[c]{@{}l@{}}Conv1d(16, 32, k=5, s=1, p=2)\\ BatchNorm1d(32)\\ ReLU()\\ MaxPool1d(k=2)\end{tabular}      & $\times 1$  \\ \hline
Conv Block 3             & \begin{tabular}[c]{@{}l@{}}Conv1d(32, 64, k=5, s=1, p=2)\\ BatchNorm1d(64)\\ ReLU()\\ MaxPool1d(k=2)\end{tabular}      & $\times 1$  \\ \hline
Conv Block 4             & \begin{tabular}[c]{@{}l@{}}Conv1d(64, 128, k=3, s=1, p=1)\\ BatchNorm1d(128)\\ ReLU()\\ AdaptiveAvgPool1d(output=1)\end{tabular} & $\times 1$  \\ \hline
Classification Head (FC) & Linear(128, 6)                                                                                                       & $\times 1$  \\ \hline
\end{tabular}
\end{adjustbox}
\caption{Model architecture details of the 1D CNN HAR classification.}
\label{tab:cnn1d}
\end{table}

We assess our method’s applicability to IoT- and mobile-oriented federated scenarios\rev{~\cite{seo2025fedunet, xiao2025federated, xing2022efficient}} 
by running FedAvg on a 1-dimensional CNN trained on an Inertial Measurement Unit (IMU) sensor dataset (UCI-HAR)~\cite{ucihar}. 
\rev{This time-series setting is closely related to recent federated learning studies on human activity recognition and multitask time-series classification~\cite{lee2025gmt}, which employ contrastive learning or feature-based distillation to enhance representation quality under heterogeneous data~\cite{xiao2025federated, xing2022efficient}. 
These approaches primarily redesign the training objective (e.g., by adding contrastive or distillation losses), while keeping the underlying softmax temperature fixed. 
In contrast, \name{} is orthogonal to such methods as it modifies the local training dynamics by applying low-temperature scaling at the softmax layer, and can, in principle, be combined with these feature-centric objectives.}

Five temperature values ($T\in\{0.25, 0.5, 1.0, 2.0, 4.0\}$) are compared to evaluate their effects on convergence speed and final accuracy. 
Table~\ref{tab:cnn1d} details the 1D-CNN configuration used, and Figure~\ref{fig:har} shows test accuracy and loss over communication rounds. 
As with our image-based experiments, lower temperatures accelerate convergence and improve final accuracy in the time-series sensor data classification task.

%% file: 5_discussion.tex
\section{\rev{DISCUSSION}}
\label{sec:discussion}

Based on our experiences in designing \name{}, we outline a set of interesting discussion points that point towards meaningful future research directions.
\vspace{0.1 in}

\noindent{\textbf{$\bullet$ Identifying the optimal temperature.}} Our study exhibits the effectiveness of exploiting lower temperatures as a way to enhance federated learning performance. However, since the softmax layer training temperature can be considered a hyperparameter, optimal temperatures can vary depending on the applications and model architectures. Thus, like other hyperparameters used in model training (e.g., batch size, local epoch), identifying an optimal temperature value is challenging to analytically solve, being beyond the scope of this paper.

%Thus, the development of an algorithm for determining the optimal temperature for a given configuration for future work.

Nevertheless, based on our experiences, we can offer heuristics for selecting temperatures regarding the model capacity and task complexity.
Large capacity models with relatively easy tasks (e.g., low-resolution data, small number of labels to classify) often exhibit unstable convergence with low temperatures. We hypothesize that this phenomenon arises from lower temperatures compelling models to rapidly adapt to local data, which can potentially bias the model towards local data and disrupt the federated learning process. Thus, we carefully conclude our heuristics in two-fold. (i) Balancing the model capacity and task complexity is crucial to best leverage the temperature scaling during training; (ii) Large capacity models tend to work well with relatively high temperatures since they tend to suffer from instability with a too-low temperature.
\vspace{0.1 in}

\noindent{\textbf{$\bullet$ Dynamic per-sample temperature scaling.}} While our paper focuses on understanding the advantages of maintaining a \textit{constant low temperature} for the training process, this may not be the ideal approach when targeting maximum performance. In reality, varying temperatures can be applied in the training process with respect to their significance~\cite{shin2022fedbalancer, zeng2021mercury}. That would mean that the training process can exploit lower temperatures for the samples that are ``harder'' to learn, while higher temperatures are used for those easier to understand. Alternatively, a more course-grain approach of dynamically adjusting the temperature on a per-federated learning round-basis can be another meaningful approach to adapt towards changing local data characteristics.
\vspace{0.1 in}

% \noindent{\textbf{$\bullet$ Supporting heterogeneous federated learning.}} While many federated learning frameworks assume homogeneous devices, recent work points out that this is not always the case~\cite{PARK23fedhm}. In a federated learning network consisting of devices with heterogeneous computing/network capabilities, the issue of ``stragglers'', which take longer to complete local training, becomes important to address\cite{shin2024effective}. Leveraging insights gleaned from our empirical observations, applying lower temperatures for potential stranglers can be an interesting solution. This would require a prior analysis on the conditions expected at each device and a global temperature configuration consensus. We find this as an interesting direction of future research.
\noindent{\textbf{$\bullet$ Supporting computation heterogeneity.}} While many federated learning frameworks assume homogeneous devices, recent work points out that this is not always the case~\cite{PARK23fedhm}. In a federated learning consisting of devices with heterogeneous computing and network capabilities, the issue of ``stragglers,'' which take longer to complete local training, becomes important to address~\cite{shin2024effective}. Leveraging insights from our empirical observations, applying lower temperatures for potential stragglers can be an interesting solution; it would require prior analysis of each device’s expected conditions and a global temperature configuration consensus. We find this an interesting direction of future research. Future work could profile each client’s compute and network resources to assign and negotiate personalized temperatures, lower on less constrained devices to speed convergence, and relatively higher on powerful clients to reduce overhead.

\noindent{\rev{\textbf{$\bullet$ Developmental Tendencies and Challenges.}}}
\rev{We identify several developmental tendencies and challenges that can guide future research. Beyond using a fixed temperature, dynamic or client-adaptive temperature scheduling represents a natural next step, particularly as our results indicate that low-temperature training preserves client-specific knowledge more effectively (as also suggested by the smaller post-aggregation accuracy drop in Figure~\ref{fig:trained-acc}). This makes \name{} a promising foundation for integration with personalization frameworks and heterogeneous model deployments. At the same time, important challenges remain, including determining the optimal temperature for different model capacities and task complexities, ensuring stability for large networks under very low temperatures, and understanding performance limits under extreme non-i.i.d. scenarios such as fully disjoint label distributions. Addressing these open issues will further clarify the scope and robustness of temperature-based training in federated learning.}

%% file: 6_conclusion.tex
\section{\rev{CONCLUSION}}
\label{sec:conclusion}
This research introduces a novel training approach for federated learning systems, \name{}, aimed at enhancing convergence and improving model performance by exploiting a temperature scaling method termed \textit{logit chilling} in federated learning's local model training operations. Specifically, our work revolves around the implications of employing fractional temperature values ($T\in(0,1)$) \textit{during} model training. Our findings, derived from an extensive set of evaluations using three datasets and three widely applied neural networks for federated learning, demonstrate that exploiting the \name{} approach in the client-side model training process holds the potential to expedite model convergence by inducing more pronounced changes despite a limited number of local samples. Consequently, we show that by doing so, we can also achieve an overall higher inference accuracy. While the impact of lower temperature varies with respect to the neural network architecture and dataset, our evaluations reveal up to 6.00$\times$ improvement in federated learning model convergence speed and 3.37\% improvement in inference accuracy.
We see this work as a first step into devising a more fine-tuned (and optimal) training mechanism for federated learning systems.

%% file: 9999_appendix.tex
% \section{Appendix}
\section*{Appendix}
%\vspace{1ex}

\vspace{1ex}
\section*{$\bullet$ Gradient of \rev{Softmax with Varying Temperature}}
%\label{subsec:proof1}
\begin{proof}
    Let cross-entropy loss for softmax as $\mathcal{L} = -\sum_{j=1}^{C}y_{j}\log{(p_{j})}$ where $C$ is the number of classes and $y_{j}$ and $p_{j}$ corresponds to the hard label (i.e., onehot-encoded label) and probability for given class. Here, the probability $p_{j}$ is computed by exploiting the softmax with varying temperatures which can be formalized as follows.
    \begin{equation*}
        p_{j}=\frac{e^{z_{j}/T}}{\sum_{i=1}^{C}{e^{z_{i}/T}}}
    \end{equation*}
    Here, $T$ is the temperature, and $z_{j}$ is the logit used as the input of the softmax function.
    
    \begin{align*}
    \frac{\partial \mathcal{L}}{\partial z_{i}}&=-\sum_{j=1}^{C}\frac{\partial y_{j}\log{(p_{j})}}{\partial z_{i}} \\
    &= \sum_{j=1}^{C}y_{j}\frac{\partial \log{(p_{j})}}{\partial z_{i}}\\
    &= -\sum_{j=1}^{C}y_{j}\frac{\partial \log{(p_{j})}}{\partial z_{i}} \\
    &= -\sum_{j=1}^{C}y_{j}\frac{1}{p_{j}}\frac{\partial p_{j}}{\partial z_{i}}\\
    &= -\frac{y_{i}}{p_{i}}\frac{\partial p_{i}}{\partial z_{i}}-\sum_{j\neq i}^{C}\frac{y_{j}}{p_{j}}\frac{\partial p_{j}}{\partial z_{i}} \\
    &= -\frac{y_{i}}{p_{i}}\frac{1}{T}p_{i}(1-p_{i})-\sum_{j\neq i}^{C}\frac{y_{j}}{p_{j}}(-p_{j}p_{i})\\
    &= -\frac{y_{i}}{T}+\frac{1}{T}y_{i}p_{i}+\frac{1}{T}\sum_{j\neq i}^{C}y_{j}p_{i} \\
    &= -\frac{y_{i}}{T}+\frac{1}{T}\sum_{j=1}^{C}y_{j}p_{i}\\
    &= -\frac{y_{i}}{T}+\frac{1}{T}p_{i}\sum_{j=1}^{C}y_{j}\\
    &= \frac{1}{T}(p_{i}-y_{i})\\
   \therefore \frac{\partial\mathcal{L}}{\partial z_{i}} &=\frac{1}{T}(p_{i}-y_{i}) \\
    &=\frac{1}{T}(\frac{e^{z_{i}/T}}{\sum_{j}e^{z_{j}/T}} - y_{i})
    \end{align*}
\end{proof}
\break

\vspace{1ex}
\section*{$\bullet$ Convergence Analysis}
\label{subsec:proof2}
\subsection*{Stochastic Optimization: SGD}
We first start the convergence analysis with basic \textsc{SGD} optimizer \cite{dean2012large}, incorporating both temperature scaling $T$ and learning rates $\eta$.

\medskip
\noindent{}\textbf{Assumption 1. $L$-smoothness}: local loss function $F_{k}(w)$ is L-smooth.
\begin{equation*}
     ||\nabla F_{k}(w_{1})- \nabla F_{k}(w_{2})||\leq L||w_{1}-w_{2}||
\end{equation*}
\textbf{Assumption 2. $\mu$-strong convexity}: loss function $F(w)$ is $\mu$-strong convex.
\begin{equation*}
     F(w_{1}) \geq F(w_{2}) + \nabla F(w_{2})^{\top}(w_{1}-w_{2}) + \frac{\mu}{2}||w_{1}-w_{2}||
\end{equation*}
\textbf{Assumption 3. Bounded Variance}: We assume that the gradient of the local loss function ($\nabla F_{k}(w)$) has finite bounded variance which means that the difference between the local gradient and the global gradient is bounded.

1. Local update

\begin{equation*}
    w_{k}^{(t+1)} = w_{k}^{(t)} - \eta \nabla F_{k}(w_{k})^{(t)}
\end{equation*}

2. Global model update

\begin{equation*}
    w^{(t+1)} = \sum_{k=1}^{K}p_{k}w_{k}^{(t+1)}
\end{equation*}

3. Difference between local and global update

\begin{align*}
    &\mathbb{E}[F(w^{(t+1)})-F(w^{*})]\leq \\
    &(1-\eta \mu) \mathbb{E}[F(w^{(t)})-F(w^{*})]+\frac{L\eta^{2} \sigma^{2}}{2}
\end{align*}
$\therefore$ $t \xrightarrow{} \infty$, $F(w^{t}) \xrightarrow{} F(w^{*})$

4. Incorporating temperature scaling

Recall that we can rewrite the local gradient as follows (see Section~\ref{sec:design}-\ref{subsec:ta}): 
\begin{align*}
    \nabla F_{k}(w) = \frac{1}{T}\nabla F_{k}(w;T=1)
\end{align*}

\begin{align*}
    \therefore \ & \mathbb{E}[F(w^{(t+1)})-F(w^{*})] \leq \\
    &(1-\eta \mu/T) \mathbb{E}[F(w^{(t)})-F(w^{*})]+\frac{L\eta^{2} \sigma^{2}}{2T^{2}} \blacksquare.
\end{align*}

\noindent{\textbf{Quantitative Trade‐off between Temperature and Learning Rate.}}

Starting from
\begin{align*}    
    &\mathbb{E}[F(w^{(t+1)}) - F(w^{*})] \;\le\;\\&\underbrace{\bigl(1 - \tfrac{\eta\mu}{T}\bigr)}_{\rho}\, \mathbb{E}[F(w^{(t)}) - F(w^{*})] + \frac{L\,\eta^{2}\,\sigma^{2}}{2\,T^{2}},
\end{align*}
we identify:
\begin{itemize}
  \item \emph{Contraction factor}: $\rho = 1 - \eta\mu/T$. To ensure $\rho<1$, one needs
    \[
      0 < \eta < \frac{T}{\mu}.
    \]
    A larger $T$ (for fixed $\eta$) yields $\rho$ closer to 1, i.e., slower exponential decay.
  \item \emph{Steady‐state error term}: $\;\frac{L\,\eta^{2}\,\sigma^{2}}{2\,T^{2}}\,$, which scales like $(\eta/T)^{2}$.
\end{itemize}

Hence after $t$ rounds,
\[
    \mathbb{E}[F(w^{(t)}) - F(w^{*})]
    \approx \rho^{t}\,[F(w^{(0)}) - F(w^{*})]
    + O\!\bigl((\eta/T)^{2}\bigr).
\]
Noting $\rho^{t}\approx\exp(-t\,\eta\mu/T)$, the \emph{effective learning rate} is $\alpha = \eta/T$.

\medskip
\noindent{}\textbf{Optimal Effective Stepsize and Temperature Scheduling}
Define $\alpha = \eta/T$. Then:
\[
  \rho = 1 - \alpha\,\mu,\quad
  \text{steady‐state error} = O(\alpha^{2}).
\]
Balancing convergence speed and variance suggests a diminishing $\alpha$, e.g.
\[
  \alpha_{t} = \frac{1}{\mu\,t},
  \quad\Longrightarrow\quad
  \eta_{t} = \frac{T}{\mu\,t}.
\]
Such a schedule ensures both rapid early descent and vanishing steady‐state error.

\noindent{\textbf{Summary.}}
\begin{itemize}
  \item Both contraction speed and variance error depend on the ratio $\eta/T$.
  \item Higher temperature $T$ for fixed $\eta$ slows convergence and reduces variance term.
  \item A time‐decaying stepsize $\eta_{t}\propto T/t$ attains the best trade‐off.
\end{itemize}

\subsection*{Adaptive Optimization: Adam}
\label{subsec:adam}

We now extend the above convergence analysis to the widely used \textsc{Adam} optimizer \cite{kingma2014adam}, incorporating both temperature scaling $T$ and adaptive per‐coordinate learning rates.
\medskip

\textbf{Assumption 4. Bounded Gradients \& Second Moments.}  For all $k,t$ and coordinates $i$,
\[
    |\nabla_i F_{k}(w^{(t)})| \le G,\quad
    \mathbb{E}\bigl[\nabla_i F_{k}(w^{(t)})^2\bigr] \le G^2.
\]
Let $\beta_{1},\beta_{2}\in[0,1)$ be the \textsc{Adam} decay rates, and $\epsilon>0$ a small constant.

\medskip

\noindent 1. \emph{Local Adam update} on client $k$ at round $t$:
\begin{align*}
    m_{k}^{(t)} &= \beta_{1}\,m_{k}^{(t-1)} \;+\; (1-\beta_{1})\,\nabla F_{k}(w_{k}^{(t)};T),\\
    v_{k}^{(t)} &= \beta_{2}\,v_{k}^{(t-1)} \;+\; (1-\beta_{2})\,\bigl[\nabla F_{k}(w_{k}^{(t)};T)\bigr]^{2},\\
    \hat m_{k}^{(t)} &= \frac{m_{k}^{(t)}}{1-\beta_{1}^{t}},\quad
    \hat v_{k}^{(t)} = \frac{v_{k}^{(t)}}{1-\beta_{2}^{t}},\\
    w_{k}^{(t+1)}
    &= w_{k}^{(t)}
      \;-\;\eta\;\frac{\hat m_{k}^{(t)}}{\sqrt{\hat v_{k}^{(t)}} + \epsilon}.
\end{align*}

\noindent 2. \emph{Global aggregation:
\[
    w^{(t+1)} \;=\; \sum_{k=1}^{K} p_{k}\,w_{k}^{(t+1)}.
\]}

\medskip
Under Assumptions 1–4 and for a suitably chosen constant stepsize $\eta$, one can show~\cite{reddi2019convergence} that
\begin{align*}
    \frac{1}{T}\,\nabla F_{k}(w) \Longrightarrow
    \frac{1}{T}\,\mathbb{E}\bigl[\|\nabla F(w^{(t)})\|_1\bigr]
    \;\le\;
    O\!\Bigl(\tfrac{1}{\sqrt{T\,t}}\Bigr)
    + O\!\Bigl(\tfrac{1}{T\,t}\Bigr).
\end{align*}
More precisely, after $T_{\max}$ total gradient steps (across all clients and rounds), the averaged gradient norm satisfies
\[
  \frac{1}{T_{\max}} \sum_{t=1}^{T_{\max}}
  \mathbb{E}\bigl[\|\nabla F(w^{(t)})\|_{1}\bigr]
  \;\le\;
  \frac{D_1}{\sqrt{T\,T_{\max}}}
  + \frac{D_2}{T\,T_{\max}},
\]
for constants $D_1,D_2$ depending on $G,\beta_{1},\beta_{2},\epsilon,L$ and the initialization.

\medskip
\noindent{\textbf{Trade‐offs under Adam \& Temperature Scaling}}
\begin{itemize}
  \item Temperature scaling reduces each client’s effective gradient by $1/T$, slowing convergence by a factor $1/\sqrt{T}$ in the dominant $O(1/\sqrt{T\,t})$ term.
  \item \textsc{Adam}’s normalization by $\sqrt{\hat v_{k}^{(t)}}$ mitigates the noise variance term, yielding a faster $\tilde O(1/\sqrt{t})$ decay even when gradients are highly heterogeneous across clients.
  \item Define $\tilde\alpha = \eta/T$. Then the leading bound scales as $O\bigl(1/(\sqrt{\tilde\alpha\,t})\bigr)$, suggesting one should choose $\eta\propto T$ and possibly decay $\eta_t\propto T/\sqrt{t}$ to balance convergence and steady‐state variance.
\end{itemize}

\noindent{\textbf{Summary.}}
\begin{itemize}
  \item Incorporating \textsc{Adam} yields an $O(1/\sqrt{T\,t})$ convergence rate under temperature‐scaled gradients.
  \item Temperature $T$ still slows down the convergence rate via a $1/\sqrt{T}$ factor.
  \item A practical schedule is $\eta_{t} = \Theta\bigl(T/\sqrt{t}\bigr)$ to achieve the best trade‐off between bias and variance under adaptive optimization.
\end{itemize}

\subsection*{\rev{Adaptive Aggregation: FedAdam}}
\label{subsec:fedadam}

FedAdam \cite{reddi2020adaptive} is a variant of FedAvg that incorporates the adaptive moment estimates of Adam at the server side.  In FedAdam, each client still performs local SGD (or optionally local Adam) updates, but the global aggregation step applies an Adam‐style update to the server’s model state.  Concretely, let
\[
    g^{(t)} \;=\; \sum_{k=1}^{K} p_{k}\,\bigl(w_{k}^{(t+1)} - w^{(t)}\bigr)
\]
be the aggregated model increment. FedAdam maintains server‐side moment estimates
\begin{align*}
    m^{(t)} &= \beta_{1}\,m^{(t-1)} + (1-\beta_{1})\,g^{(t)},\\
    v^{(t)} &= \beta_{2}\,v^{(t-1)} + (1-\beta_{2})\,\bigl(g^{(t)}\bigr)^{2},\\
    \hat m^{(t)} &= \frac{m^{(t)}}{1-\beta_{1}^{t}},\quad
    \hat v^{(t)} = \frac{v^{(t)}}{1-\beta_{2}^{t}},
\end{align*}
and updates the global model by
\[
    w^{(t+1)} = w^{(t)} - \eta_s\;\frac{\hat m^{(t)}}{\sqrt{\hat v^{(t)}} + \epsilon},
\]
where $\eta_s$ is the server‐side stepsize and $(\beta_1,\beta_2,\epsilon)$ are the usual Adam hyperparameters.

\medskip
\noindent{\textbf{Trade‐off between FedAdam and Local Adam $+$ Aggregation}}
\begin{itemize}
  \item \emph{Server‐side adaptivity vs.\ client‐side noise:}  
    FedAdam smooths the noisy aggregated gradients $g^{(t)}$ using $m^{(t)},v^{(t)}$, reducing variance at the server.  In contrast, applying Adam locally on each client before averaging retains some heterogeneity in second‐moment estimates across clients.
  \item \emph{Effective learning rate scaling:}  
    With temperature scaling $T$, each client’s model increment scales as $g^{(t)}\propto\eta/T$.  The server‐side Adam then applies another $\eta_s/T$ factor to the update, so the combined effective stepsize is $\alpha_{\mathrm{FedAdam}} \approx (\eta\,\eta_s)/T^2$.
  \item \emph{Convergence bound:}  
    Under similar bounded‐gradient assumptions, one can show (extending the analysis of \cite{reddi2019convergence}) that the server‐averaged gradient norm satisfies
    \begin{align*}
      \frac{1}{T_{\max}}\sum_{t=1}^{T_{\max}}
      &\mathbb{E}\|\nabla F(w^{(t)})\|_{1}
      \;\le\;\\
      &O\!\Bigl(\tfrac{1}{\sqrt{T^2\,T_{\max}}}\Bigr)
      + O\!\Bigl(\tfrac{1}{T^2\,T_{\max}}\Bigr),
    \end{align*}
    highlighting the $1/T^2$ slowdown due to two stages of scaling (client and server).
  \item \emph{Practical scheduling:}  
    To balance convergence speed and variance, one may choose
    \[
      \eta_t = \Theta\!\bigl(\tfrac{T}{\sqrt{t}}\bigr),
      \quad
      \eta_{s,t} = \Theta\!\bigl(\tfrac{T}{\sqrt{t}}\bigr),
    \]
    yielding an overall decay $O\bigl(1/\sqrt{t}\bigr)$ despite the double scaling.
\end{itemize}

\medskip
\noindent{\textbf{Summary.}}
FedAdam augments federated learning with server‐side adaptivity, further damping heterogeneity and noise but introducing an additional temperature scaling factor.  Its convergence rate inherits a $1/\sqrt{T^2\,t}$ dependence, suggesting careful tuning of both client and server learning rates in proportion to $T$.

\section*{$\bullet$ Details on Centered Kernel Alignment (CKA)}
\label{sec:CKA}

\rev{\subsection*{CKA as a feature-space alignment metric in our study}}
\rev{In this work, we use Centered Kernel Alignment (CKA) as our primary metric to quantify feature-space alignment between neural networks. CKA is widely used to compare internal representations across models because it is architecture-agnostic and can be computed layer-wise. In our experiments (Section~\ref{sec:featuresim}), we mainly measure CKA between a global model trained under non-i.i.d. federated settings and a reference model trained on an i.i.d. configuration. This allows us to assess how closely the federated model recovers the feature space of an ideal centralized learner.}

\rev{Empirically, we observe that lower training temperatures lead to faster and stronger feature-space alignment with the i.i.d. reference model, especially in deeper layers. As shown in Figure~\ref{fig:cka_reference} and Table~\ref{tab:cka_reference}, models trained with small temperatures such as $T=0.25$ and $T=0.05$ achieve higher CKA values in the penultimate and classification layers and reach these levels in fewer rounds than $T=1.0$ or higher. This behavior closely follows the accuracy convergence trends: configurations that exhibit faster CKA growth also converge more rapidly and reach higher final accuracy. In other words, CKA in our study is not used in isolation but is interpreted jointly with test accuracy and convergence speed, and together they provide evidence that low-temperature training accelerates both representation alignment and optimization.}

\rev{We also study feature-space similarity at the client level (Figure~\ref{fig:CKA_simmat}), where we compare each client model with the initial global model, the aggregated model, and the i.i.d. reference model. For low-temperature configurations, clients quickly learn representations that are both diverse across clients and better aligned with the reference model, and these representations are effectively aggregated at the server. This pattern is consistent with the observation that \name{} promotes faster global convergence while preserving useful client-specific information.}

\begin{figure}[h!]
    \centering
    \subfigure[Case 1]{
    \includegraphics[width=0.17\linewidth]{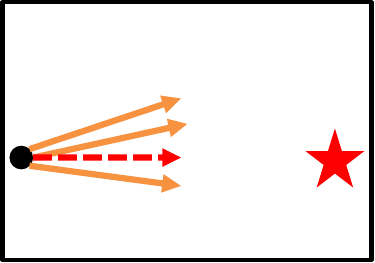}
    }
    \subfigure[Case 2]{
    \includegraphics[width=0.17\linewidth]{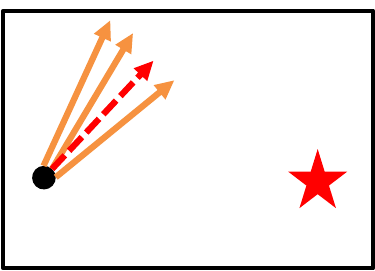}
    }
    \subfigure[Case 3]{
    \includegraphics[width=0.17\linewidth]{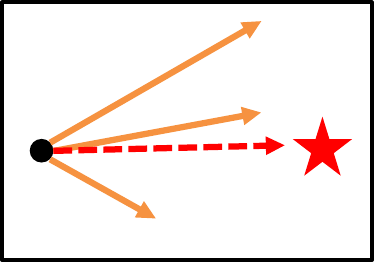}
    }
    \subfigure[Case 4]{
    \includegraphics[width=0.17\linewidth]{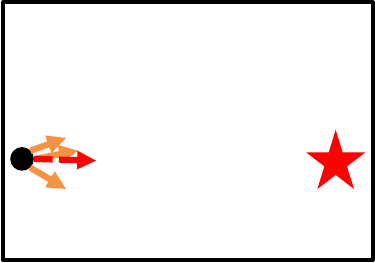}
    }
    \subfigure[Case 5]{
    \includegraphics[width=0.17\linewidth]{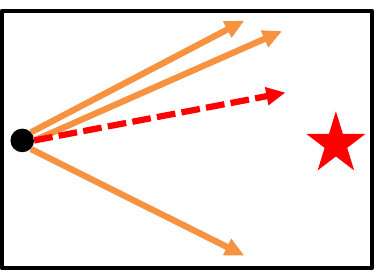}
    }
    \caption{Illustration of possible cases in a federated learning scenario. The solid orange line and dotted red line correspond to the local update direction and aggregated model update direction, respectively. The black circle and the red star depict the initial point and an ideal high-accuracy model in the representation space.}
    \label{fig:cases}
\end{figure}

\rev{\subsection*{Limitation and interpretation of CKA in federated learning}}
\rev{Despite its usefulness, CKA alone does not guarantee high model quality. This subsection explains how we interpret CKA in the context of federated learning and why we additionally compare against a reference model. Figure~\ref{fig:cases} illustrates several possible cases in an abstract representation space. The orange line denotes local update directions for client models starting from a shared global model (black circle), and the red dotted line represents the effective aggregated update at the server. The red star indicates an ideal model with high accuracy.}

\noindent\rev{\textbf{$\bullet$ High similarity between local models does not guarantee a better model (Case 1, 2):} In Case~1, local models produce highly similar updates because their data distributions are well aligned. In such a near-i.i.d. scenario, the aggregated global model can approach the ideal centralized solution. However, this case is relatively rare in realistic non-i.i.d. settings. In Case~2, local model updates are also highly similar but are aligned in a suboptimal direction. Even though the feature spaces of the local models are similar to each other, the global update may move away from the desirable region, resulting in poor accuracy. These cases highlight that high similarity among local models or their updates does not, by itself, ensure good performance.}

\noindent\rev{\textbf{$\bullet$ A superior model can be obtained without similar local models (Case 3):} Case~3 demonstrates that a strong global model can emerge even when local models exhibit relatively low pairwise similarity. Under heterogeneous data distributions, each client may learn distinct but useful features. If these diverse updates are properly aggregated, the resulting global model can achieve high accuracy and robust generalization. This scenario aligns with our empirical findings where low-temperature configurations encourage more aggressive and diverse local updates that still aggregate into a better-aligned and better-performing global model. In this sense, \name{} tends to behave closer to Case~3 than Case~2.}

\noindent\rev{\textbf{$\bullet$ Relationship between model similarity and the magnitude of updates (Case 4, 5):} Cases~4 and 5 emphasize that the magnitude of local updates also affects measured similarity. When update magnitudes are very small, models can appear highly similar even if they have not moved meaningfully toward the optimal solution. Conversely, large but beneficial updates may temporarily reduce similarity while still improving accuracy. This observation suggests that focusing solely on absolute similarity values, without considering update magnitude and direction, can be misleading.}

\rev{Based on these considerations, we do not treat CKA as a standalone proxy for performance. Instead, we measure CKA with respect to a reference model trained in an i.i.d. configuration and analyze its evolution together with accuracy. This reference-based comparison provides a more informative baseline for assessing how closely federated models recover an ideal feature space. In our experiments, federated models trained with lower temperatures consistently achieve higher CKA values relative to this reference and do so more quickly, which correlates well with faster convergence and higher final accuracy.}

\rev{Regarding extreme non-i.i.d. scenarios (for example, very small Dirichlet concentration parameters or nearly disjoint class partitions), our results indicate that absolute CKA values can indeed be lower due to stronger representation drift, which is consistent with prior observations in federated learning. However, within the same non-i.i.d. setting,\name{} still yields higher CKA values and better accuracy than training with $T=1.0$, indicating that low-temperature training partially compensates for drift by amplifying informative gradients and promoting more effective feature-space alignment. We therefore view CKA, when interpreted relative to a reference model and in conjunction with accuracy, as a meaningful but not sufficient indicator for understanding how Logit Chilling shapes representations in heterogeneous federated environments.}

\vspace{1ex}
\section*{$\bullet$ Experiment on Hyperparameters}
\label{sec:ablation}
\begin{figure}[h!]
    \centering
    \subfigure[Number of participants 5]{
    \includegraphics[width=.45\linewidth]{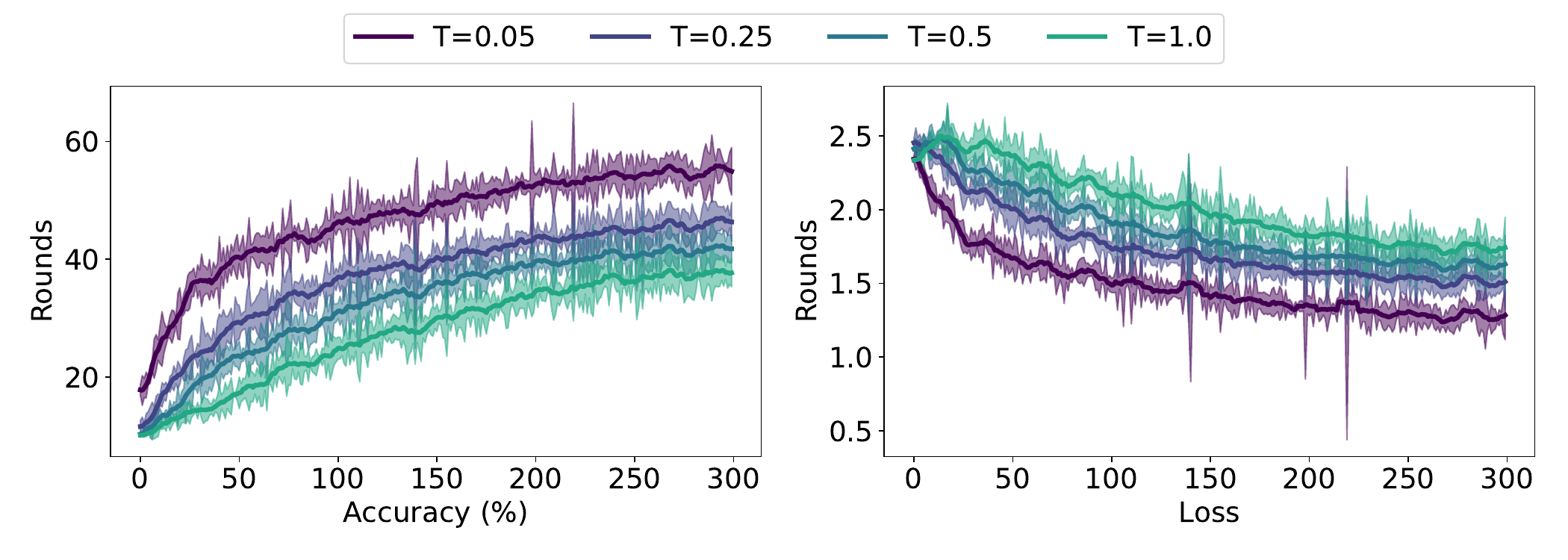}
    }
    \subfigure[Number of participants 10]{
    \includegraphics[width=.45\linewidth]{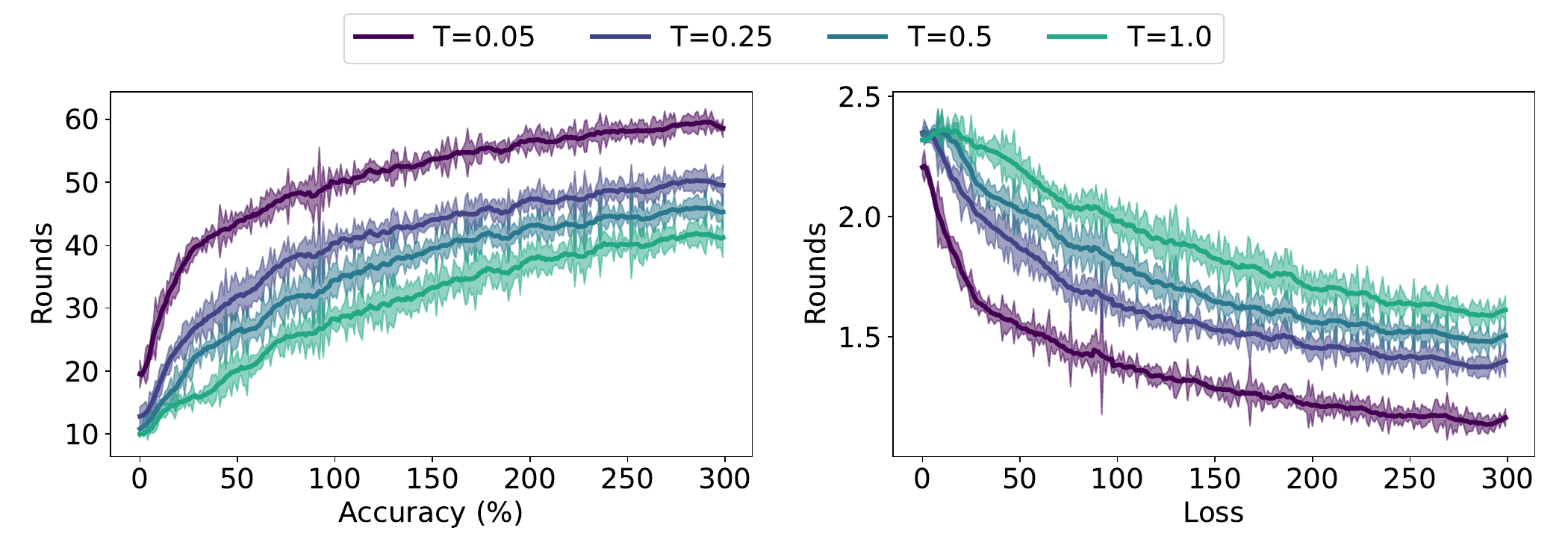}
    }
    \subfigure[Number of participants 15]{
    \includegraphics[width=.45\linewidth]{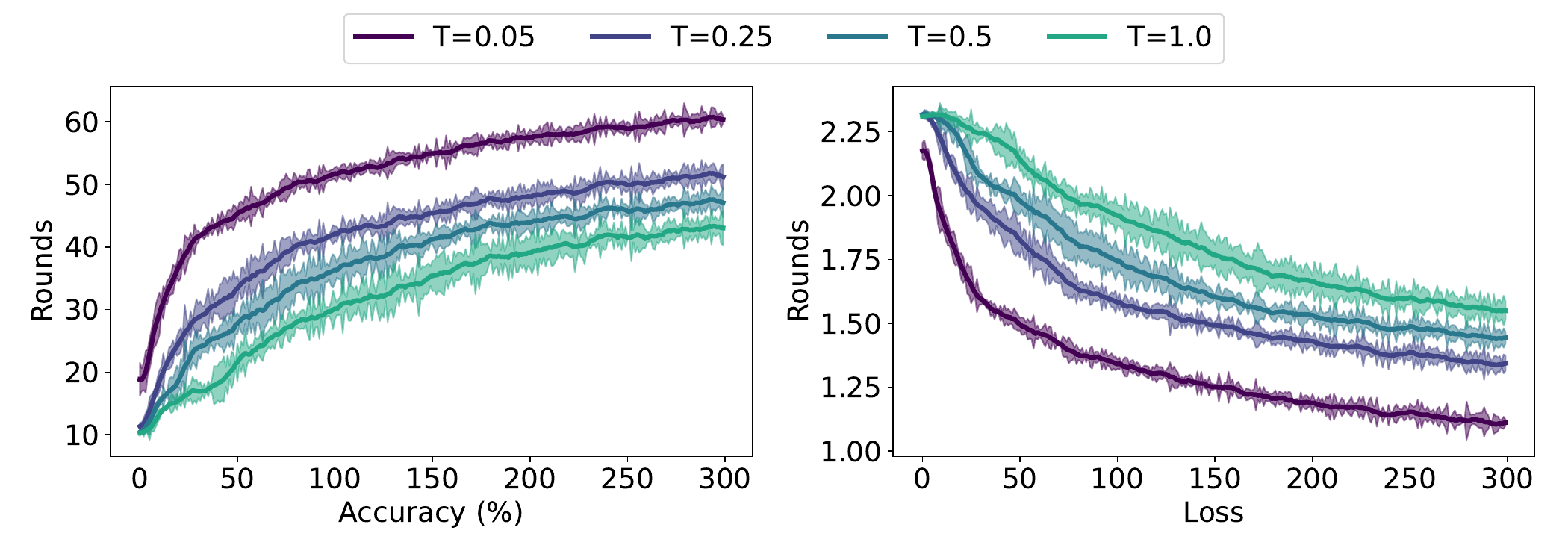}
    }
    \caption{Test accuracy and test loss over federated learning rounds with varying the number of participants and the training temperature on the CIFAR10 dataset. The shaded areas indicate the standard deviation across different runs. When relatively many clients participate in the training, it shows better training quality with stability.}
    \label{fig:np}
\end{figure}

\begin{figure}[h!]
    \centering
    \subfigure[Batch size 8]{
    \includegraphics[width=.45\linewidth]{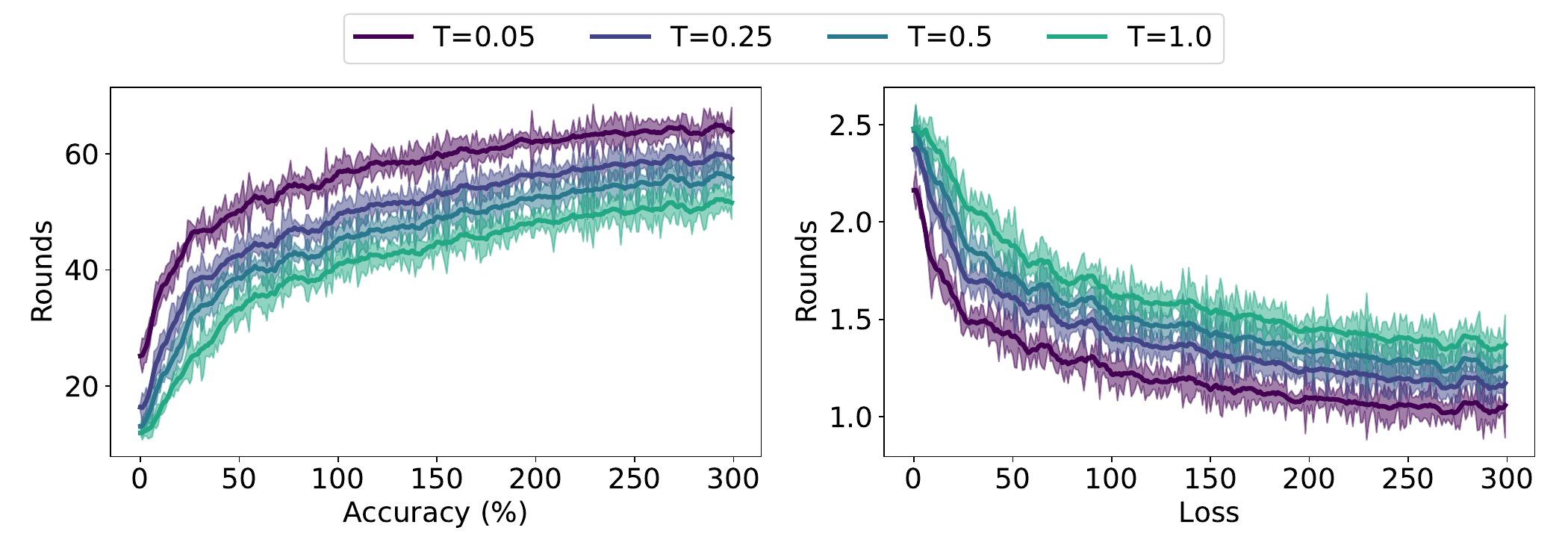}
    }
    \subfigure[Batch size 16]{
    \includegraphics[width=.45\linewidth]{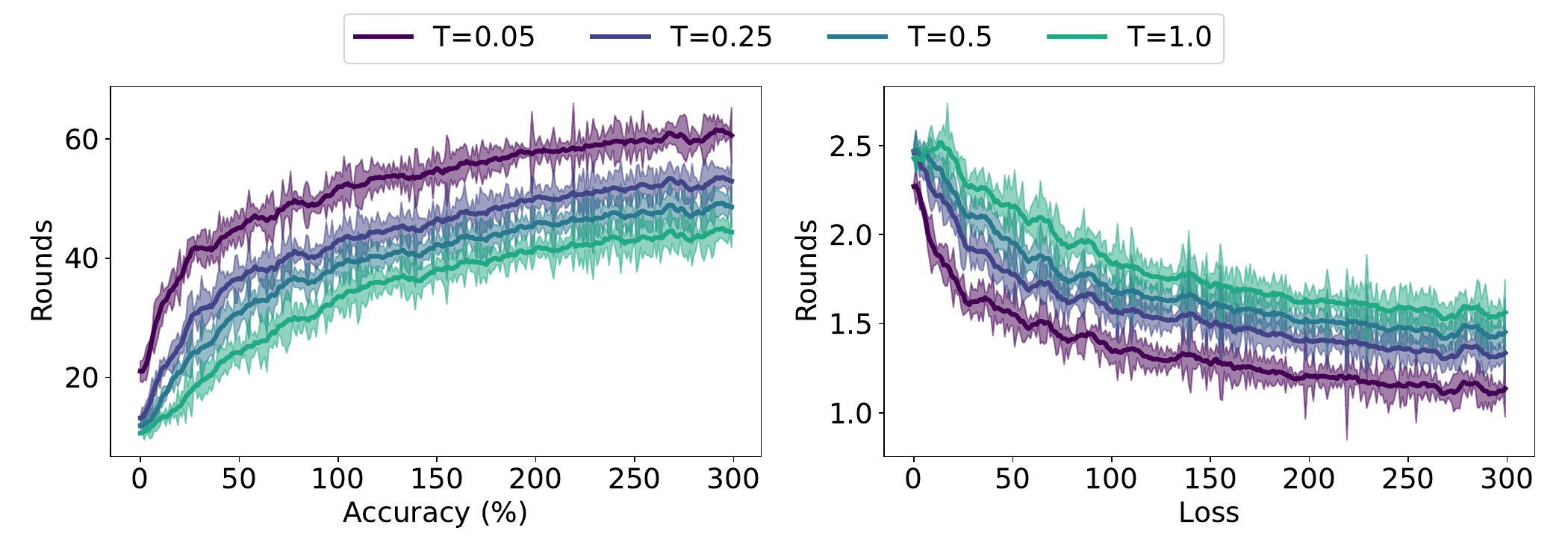}
    }
    \subfigure[Batch size 32]{
    \includegraphics[width=.45\linewidth]{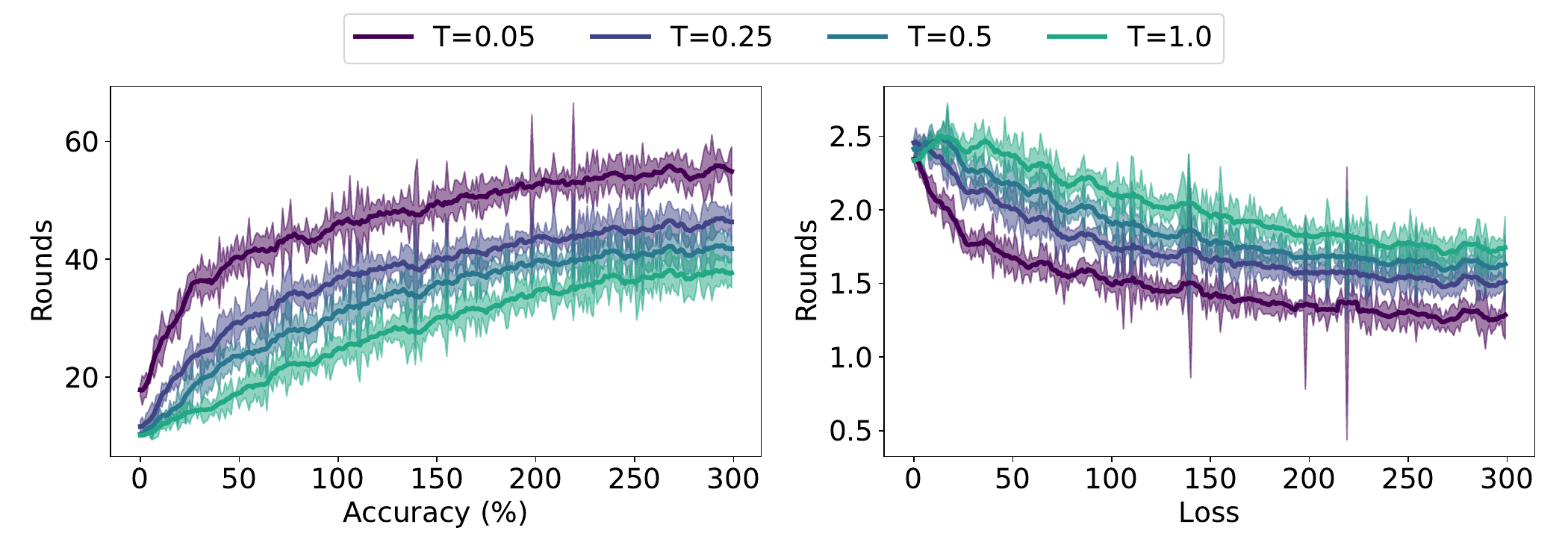}
    }
    \subfigure[Batch size 64]{
    \includegraphics[width=.45\linewidth]{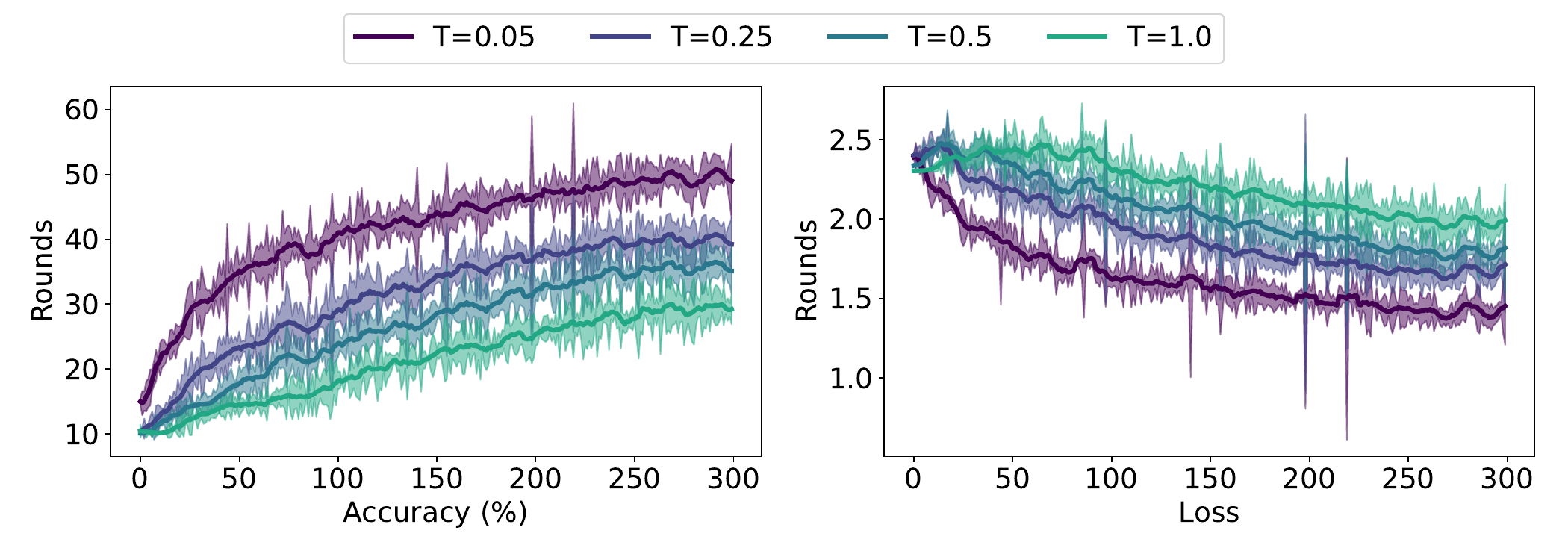}
    }
    \subfigure[Batch size 128]{
    \includegraphics[width=.45\linewidth]{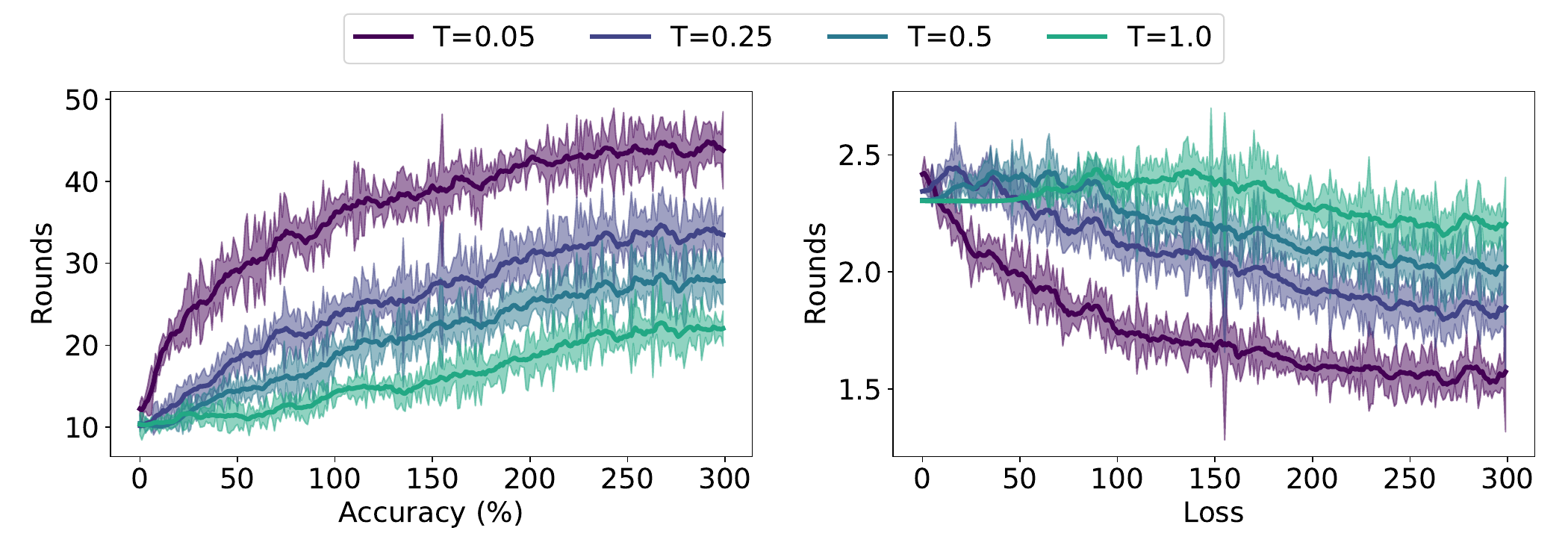}
    }
    \caption{Test accuracy and test loss over federated learning rounds with varying the batch size and the training temperature on the CIFAR10 dataset. The shaded areas indicate the standard deviation across different runs. A relatively small batch size shows better training quality with stability.}
    \label{fig:bs}
\end{figure}

\begin{figure}[h!]
    \centering
    \subfigure[Number of local epoch 1]{
    \includegraphics[width=.45\linewidth]{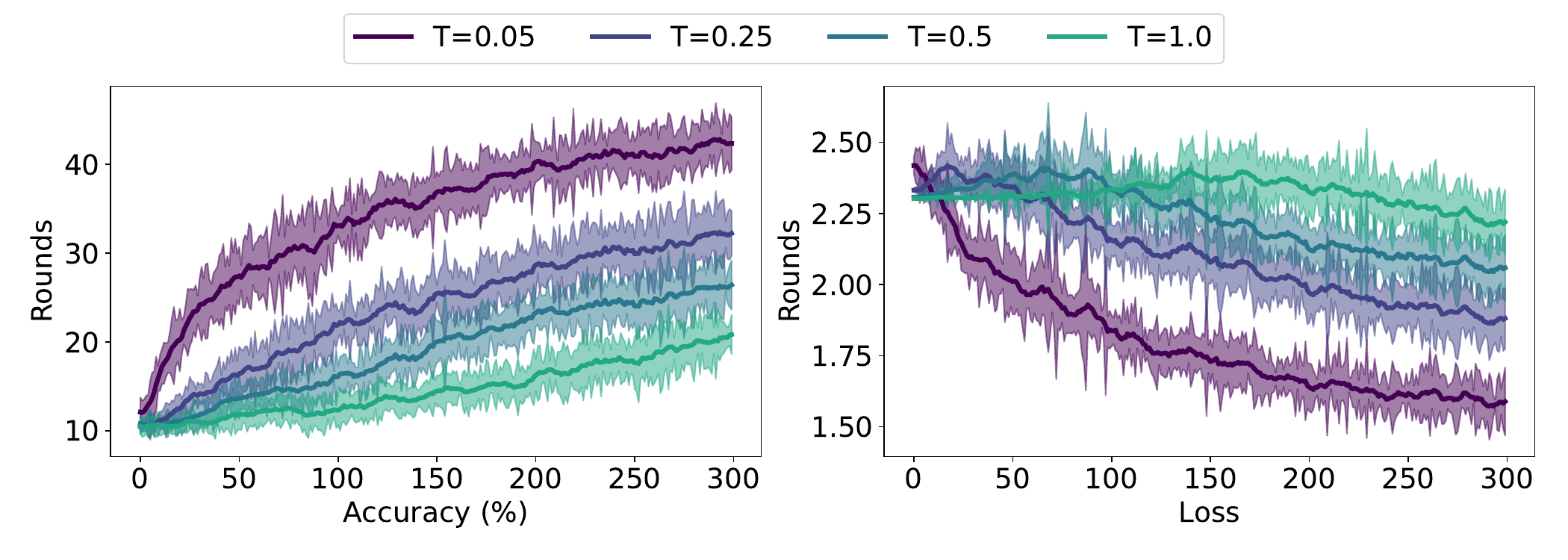}
    }
    \subfigure[Number of local epoch 5]{
    \includegraphics[width=.45\linewidth]{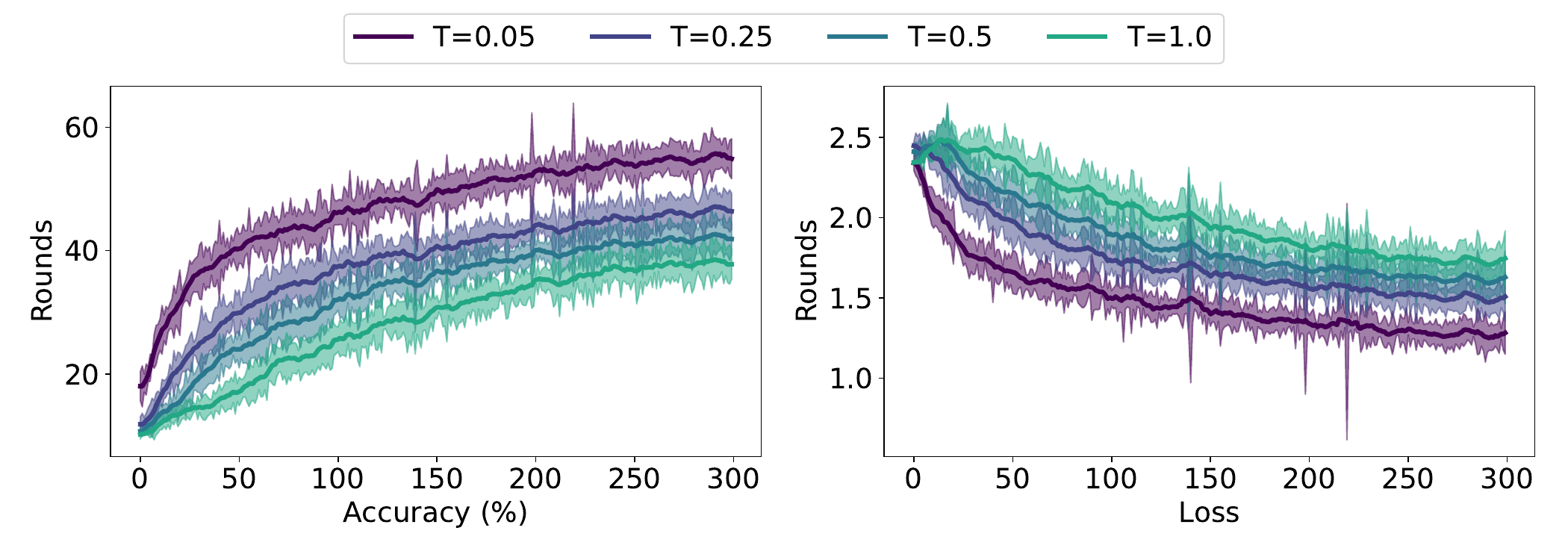}
    }
    \subfigure[Number of local epoch 10]{
    \includegraphics[width=.45\linewidth]{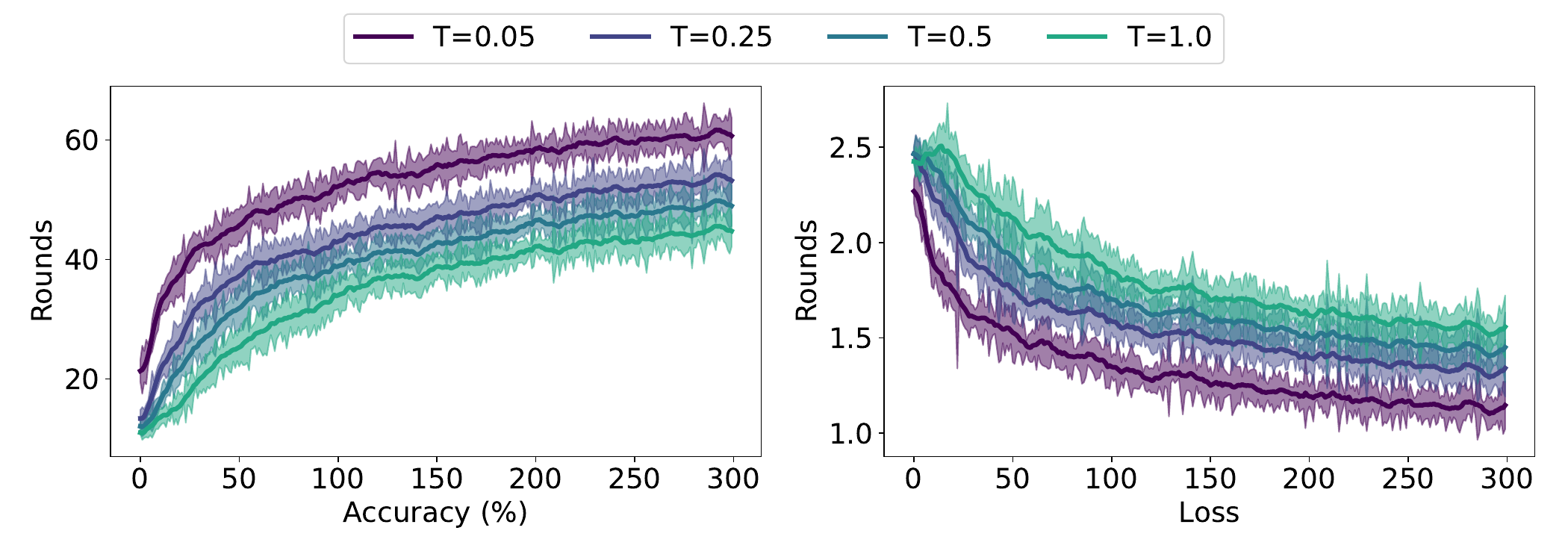}
    }
    \subfigure[Number of local epoch 15]{
    \includegraphics[width=.45\linewidth]{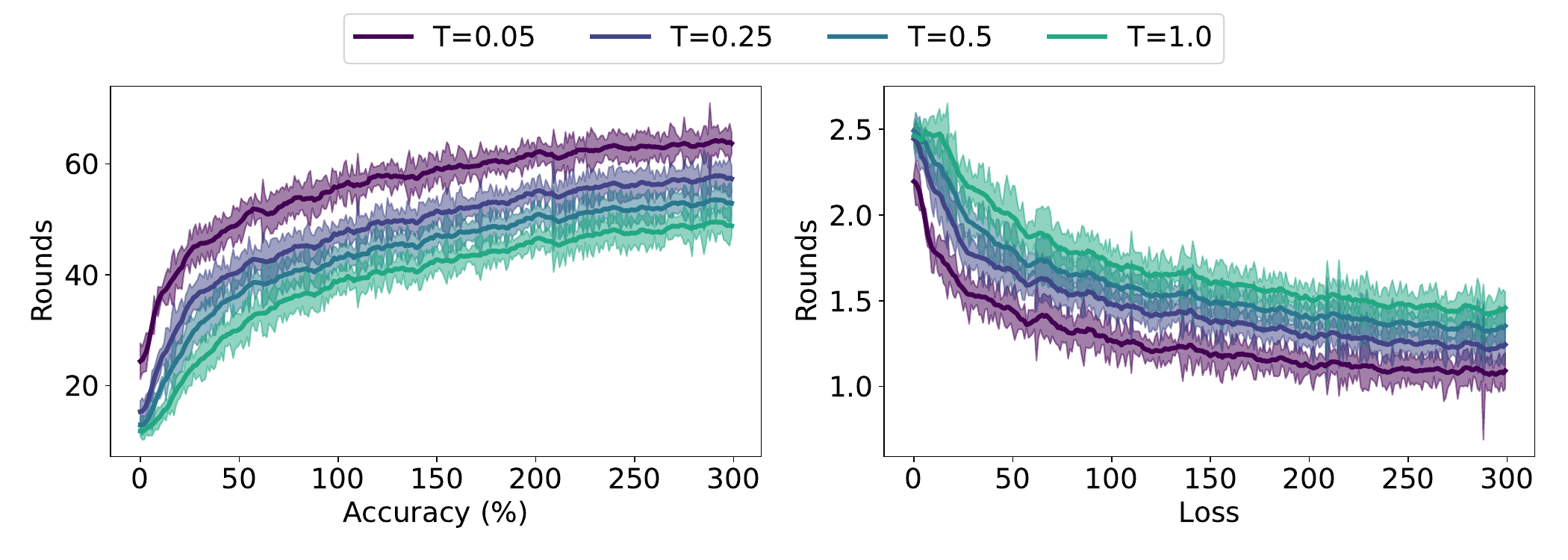}
    }
    \caption{Test accuracy and test loss over federated learning rounds with varying the number of local epochs and the training temperature on the CIFAR10 dataset. The shaded areas indicate the standard deviation across different runs. Relatively many updates in the local training phase show better training quality with stability.}
    \label{fig:ep}
\end{figure}

In this section, we explore the effect of other federated learning hyperparameters (i.e., number of participants (Figure~\ref{fig:np}), batch size (Figure~\ref{fig:bs}), and number of local epochs (Figure~\ref{fig:ep})). Through this set of experiments, we observed that applying low temperatures during local training improves test accuracy and convergence speed, being independent of hyperparameters. We note that the experiments were done by applying the default setting except for the control factors.

One can view participant count, batch size, and local epochs through the lens of gradient variance. Sampling more clients per round increases the chance that the aggregated update better approximates the true gradient, reducing variance. Likewise, using smaller batch sizes with more local epochs yields more frequent updates that capture finer-grained information from each client’s data, which also dampens variance. Because low temperatures can sometimes introduce training instability, tuning these variance-reducing parameters, more clients, smaller batches, or additional local epochs may help stabilize convergence at low temperatures.

% \subsection{Batch size}
% \label{sec:batchsize}

% \subsection{Number of participants}
% \label{sec:participants}

% \subsection{Number local epochs}
% \label{sec:epochs}
\clearpage

\section*{\rev{$\bullet$ Extended Stability Analysis Under Long-Horizon Training}}
\label{appendix:stability}
\rev{To further assess the stability of low-temperature training beyond the main experimental horizon, we conducted an extended study with 1,000 federated training rounds. This experiment aims to evaluate whether the aggressive updates induced by Logit Chilling remain stable under prolonged optimization.}

\begin{figure}[t!]
    \centering
    \includegraphics[width=0.95\linewidth]{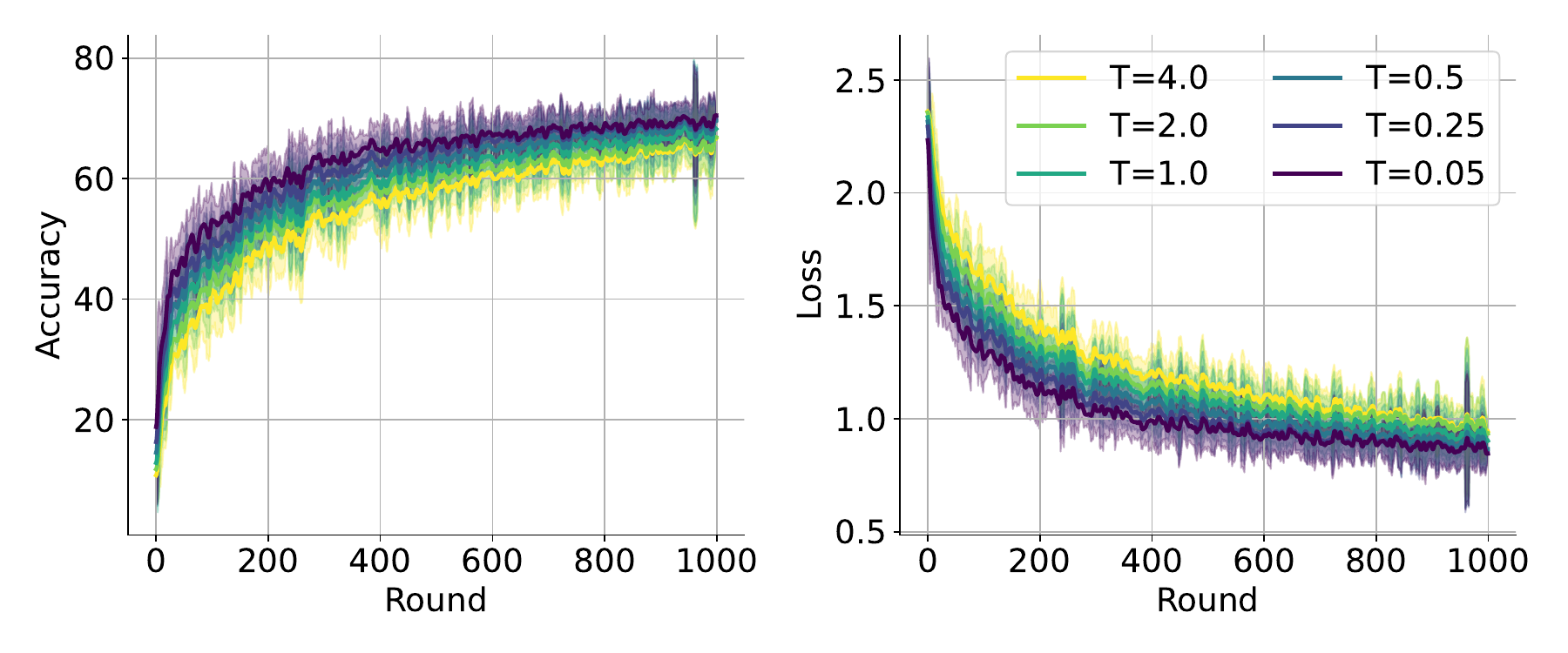}
    \caption{\rev{Average global test accuracy and test loss for 1,000 federated training rounds across different temperature settings. Lower temperatures maintain stable convergence and continue to outperform the standard/high temperature ($T\in\{1.0, 2.0, 4.0\}$).}}
    \label{fig:long_horizon_stability}
\end{figure}

\rev{Figure~\ref{fig:long_horizon_stability} presents the convergence trajectories for multiple temperature settings. Across all configurations, models trained with lower temperatures continue to exhibit stable convergence behavior. Notably, temperatures below one consistently outperform the standard/high temperature setting ($T\in\{1.0,2.0,4.0\}$) throughout the extended training horizon, reaffirming the robustness and long-term benefits of applying temperature scaling during local training. These results demonstrate that the accelerated convergence observed in the main experiments does not introduce long-term instability, even when the training duration is substantially increased.}

\section*{\rev{$\bullet$ Analysis on Model Calibration}}
\label{sec:calibration}
\begin{figure}
    \centering
    \includegraphics[width=0.95\linewidth]{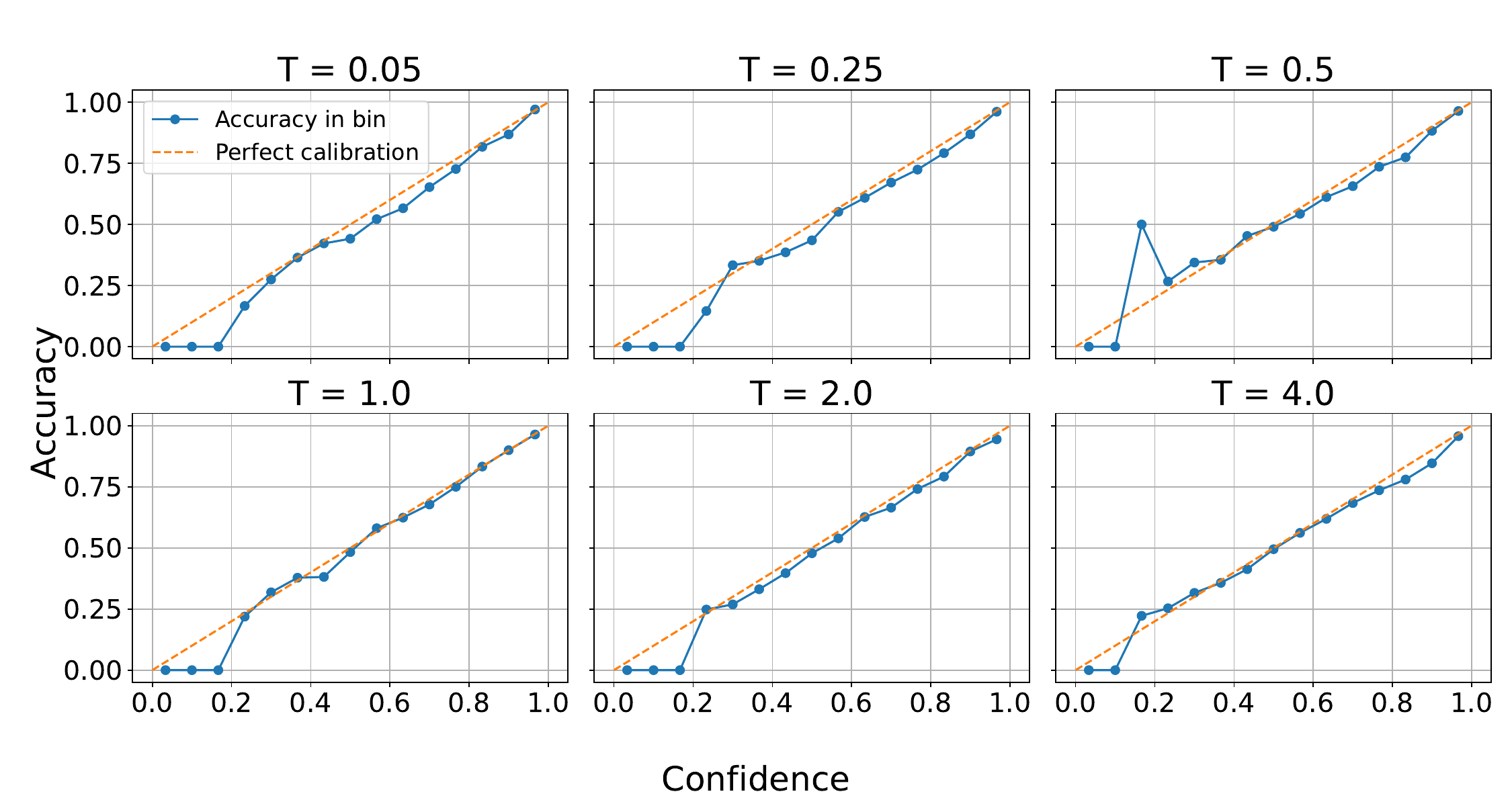}
    \caption{\rev{Reliability diagrams illustrating the calibration performance of models trained with different temperatures. Each diagram compares predicted confidence against empirical accuracy, showing that temperature scaling during training does not introduce notable miscalibration.}}
    \label{fig:calibration}
\end{figure}

\begin{figure}
    \centering
    \includegraphics[width=0.95\linewidth]{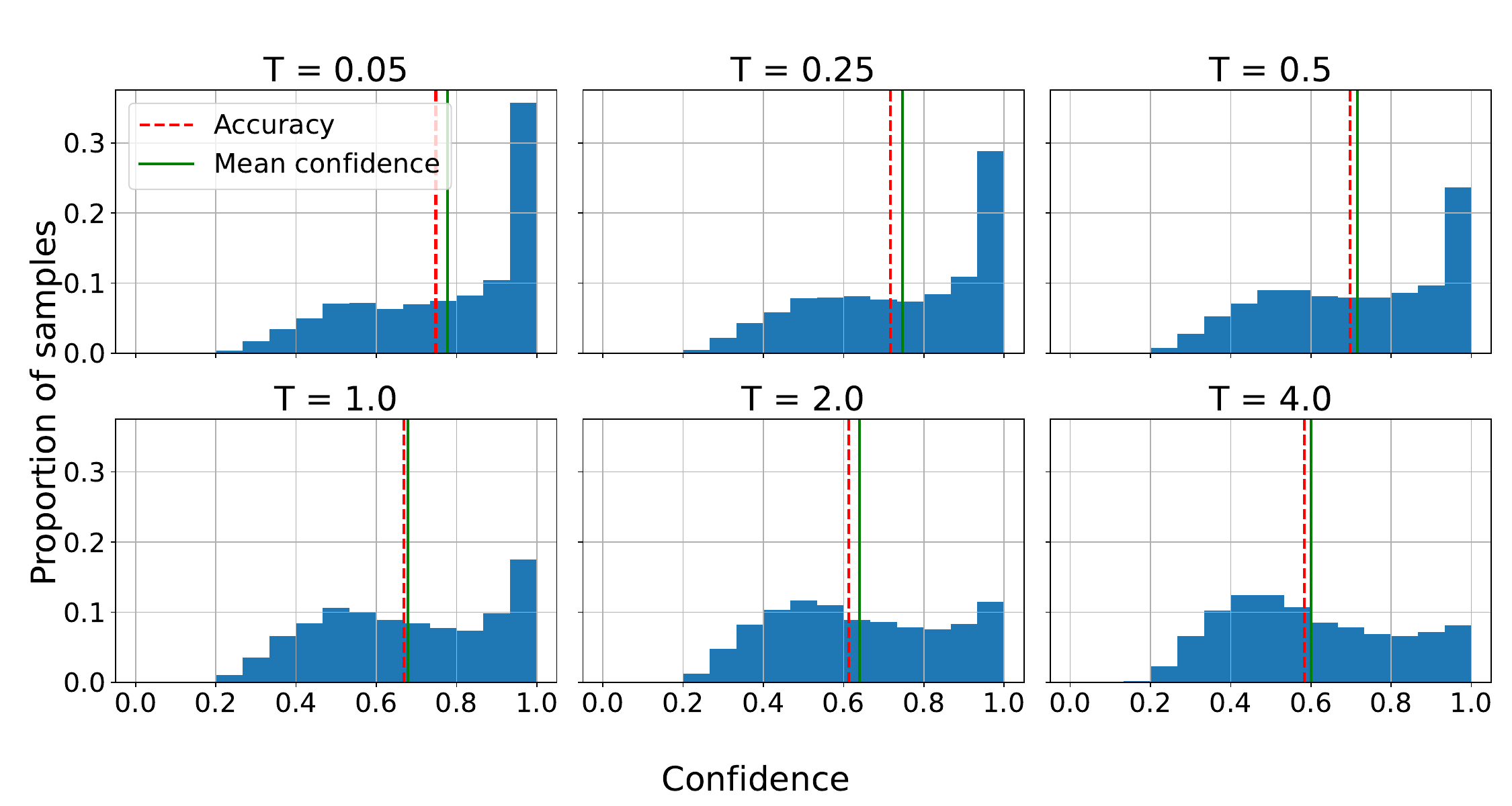}
    \caption{\rev{Prediction confidence distributions of models trained with varying temperatures. The histograms show how the confidence levels are distributed across samples, providing insight into potential over- or under-confidence behaviors under different temperature settings.}}
    \label{fig:confidence}
\end{figure}

\begin{figure}
    \centering
    \includegraphics[width=0.95\linewidth]{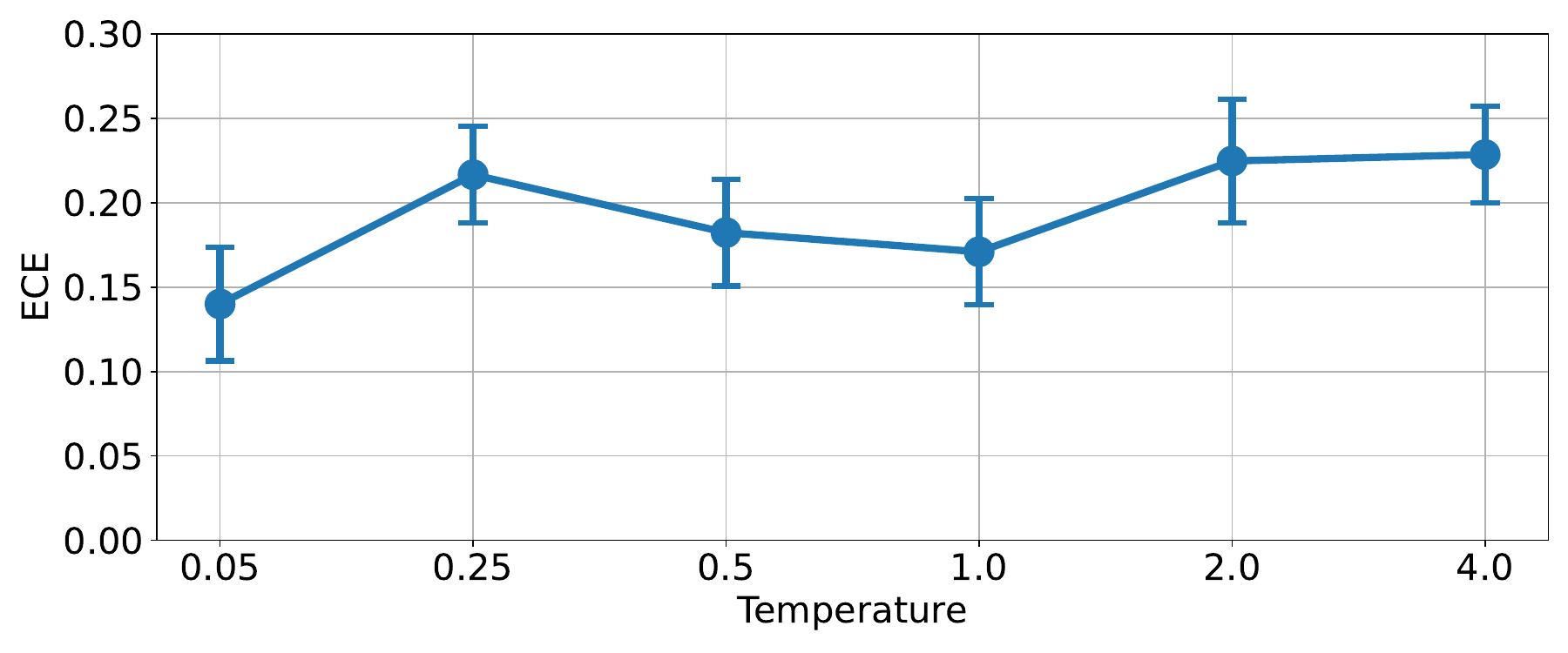}
    \caption{\rev{Expected Calibration Error (ECE) for models trained with different temperatures. Error bars indicate standard deviation across runs. The results demonstrate minimal differences in calibration quality across temperature settings.}}
    \label{fig:ece}
\end{figure}

\rev{Since temperature scaling was originally introduced for model calibration, one might be concerned that training with low temperatures could lead to over- or under-confidence. Model calibration measures how well a model’s predicted probabilities correspond to the true likelihood of correctness. A well-calibrated model outputs high confidence only when its predictions are likely to be correct, and expresses appropriate uncertainty when the outcome is less certain. Thus, calibration evaluates not only \emph{what} the model predicts but also \emph{how confident} it is in those predictions. For example, a model that frequently produces high-confidence predictions but achieves only moderate accuracy is considered over-confident, whereas a model that predicts with low confidence despite high accuracy is under-confident.}

\rev{A standard quantitative metric for evaluating calibration is the Expected Calibration Error (ECE), defined as
\[
\mathrm{ECE} = 
\sum_{m=1}^{M} 
\frac{|B_m|}{n}
\left|
\mathrm{acc}(B_m) - \mathrm{conf}(B_m)
\right|,
\]
where \(B_m\) denotes the set of samples whose predicted confidences fall into the \(m\)-th bin, \(\mathrm{acc}(B_m)\) is the empirical accuracy for that bin, \(\mathrm{conf}(B_m)\) is the average predicted confidence, and \(n\) is the total number of samples. Lower ECE values indicate better calibration.}

\rev{To assess calibration quality in our setting, we examine the prediction probability distributions, the model’s average confidence, and its accuracy. Concretely, we present the calibration results (Figure~\ref{fig:calibration}), the prediction confidence distributions (Figure~\ref{fig:confidence}), and the corresponding ECE values (Figure~\ref{fig:ece}) for models trained with different temperatures (0.05, 0.25, 0.5, 1.0, 2.0, 4.0). We note that we use the same temperature for testing as for training. Figure~\ref{fig:confidence} shows that models trained with higher temperatures produce fewer samples in the high-confidence region, which is consistent with their lower performance at the same training round and indicates that they are less certain about the given inputs. In contrast, models trained with lower temperatures assign higher confidence to a larger portion of samples, and Figure~\ref{fig:calibration} shows that these high-confidence regions remain well calibrated: the empirical accuracy in the highest confidence bins does not deteriorate, and the reliability curves stay close to the diagonal. Finally, as shown in Figure~\ref{fig:ece}, across our experimental settings, we do not observe meaningful differences in ECE among models trained with different temperatures. Taken together, the ECE values, confidence distributions, and reliability diagrams indicate that applying temperature scaling during training does not introduce notable calibration degradation, nor systematic over-confidence or under-confidence, in our setting.}

\rev{\section*{$\bullet$ Comparison to Knowledge Distillation}}
\rev{In this section, we clarify how our use of temperature scaling in \name{} differs fundamentally from conventional knowledge distillation techniques. Although both rely on a temperature parameter applied to the softmax function, their goals, where the temperature is applied, and the corresponding training regimes are distinct.}

\rev{In classical knowledge distillation, a high-capacity teacher model is first trained in a centralized manner. A temperature value larger than 1 is then applied to the teacher’s logits to produce softened probability distributions. These softened probabilities are used as soft labels to train a student model, typically by minimizing a distillation loss such as the Kullback--Leibler divergence between the teacher and student outputs, often combined with a standard cross-entropy loss on hard labels. The key role of temperature in this setting is to smooth the teacher’s output distribution so that relative relationships among non-argmax classes become more informative to the student. Importantly, in conventional knowledge distillation, the temperature is introduced primarily to generate softened targets and to define the distillation loss: both the teacher and the student outputs are typically divided by the same temperature $T>1$ inside the softmax when computing the distillation term, while the hard-label loss and inference-time predictions of the student still use temperature $T=1$. In other words, the temperature is mostly confined to the soft-label matching objective, rather than being used as a global training knob that systematically controls all aspects of the student’s optimization dynamics.}

\rev{In contrast, our work does not involve a teacher--student architecture or soft-label transfer. \name{} applies temperature scaling directly to the softmax of the model that is being trained on each client, and this is done throughout the federated training process rather than as a post-training calibration or label-smoothing step. Moreover, we operate in the low-temperature regime where the temperature lies between 0 and 1, which sharpens the output distribution and, as analyzed in our convergence theory, amplifies gradient magnitudes and accelerates optimization in the presence of non-i.i.d. data. In other words, temperature in our setting is used as an optimization control knob that shapes gradient flow and feature-space alignment across heterogeneous clients, not as a mechanism for compressing or transferring knowledge from a pre-trained teacher network. These differences imply that \name{} is complementary to, rather than a variant of, conventional knowledge distillation.}

%% file: reference.bib
@article{lee2025mind,
  title={Mind your indices! Index hijacking attacks on collaborative unpooling autoencoder systems},
  author={Lee, Kichang and Yun, Jonghyuk and Jin, Jaeho and Han, Jun and Ko, JeongGil},
  journal={Internet of Things},
  volume={29},
  pages={101462},
  year={2025},
  publisher={Elsevier}
}

@inproceedings{lee2023exploiting,
  title={Exploiting Indices for Man-in-the-Middle Attacks on Collaborative Unpooling Autoencoders},
  author={Lee, Kichang and Yun, Jonghyuk and Han, Jun and Ko, Jeonggil},
  booktitle={Proceedings of the 21st Annual International Conference on Mobile Systems, Applications and Services},
  pages={598--599},
  year={2023}
}

@article{reddi2020adaptive,
  title={Adaptive federated optimization},
  author={Reddi, Sashank and Charles, Zachary and Zaheer, Manzil and Garrett, Zachary and Rush, Keith and Kone{\v{c}}n{\`y}, Jakub and Kumar, Sanjiv and McMahan, H Brendan},
  journal={arXiv preprint arXiv:2003.00295},
  year={2020}
}

@article{reddi2019convergence,
  title={On the convergence of adam and beyond},
  author={Reddi, Sashank J and Kale, Satyen and Kumar, Sanjiv},
  journal={arXiv preprint arXiv:1904.09237},
  year={2019}
}

@article{dean2012large,
  title={Large scale distributed deep networks},
  author={Dean, Jeffrey and Corrado, Greg and Monga, Rajat and Chen, Kai and Devin, Matthieu and Mao, Mark and Ranzato, Marc'aurelio and Senior, Andrew and Tucker, Paul and Yang, Ke and others},
  journal={Advances in neural information processing systems},
  volume={25},
  year={2012}
}

@article{kingma2014adam,
  title={Adam: A method for stochastic optimization},
  author={Kingma, Diederik P and Ba, Jimmy},
  journal={arXiv preprint arXiv:1412.6980},
  year={2014}
}

@article{lee2024tazza,
  title={Tazza: Shuffling Neural Network Parameters for Secure and Private Federated Learning},
  author={Lee, Kichang and Jin, Jaeho and Park, JaeYeon and Ko, JeongGil},
  journal={arXiv preprint arXiv:2412.07454},
  year={2024}
}

@article{lee2024detrigger,
  title={DeTrigger: A Gradient-Centric Approach to Backdoor Attack Mitigation in Federated Learning},
  author={Lee, Kichang and Shin, Yujin and Yun, Jonghyuk and Han, Jun and Ko, JeongGil},
  journal={arXiv preprint arXiv:2411.12220},
  year={2024}
}

@article{PARK23fedhm,
title = {FedHM: Practical federated learning for heterogeneous model deployments},
journal = {ICT Express},
year = {2023},
issn = {2405-9595},
doi = {https://doi.org/10.1016/j.icte.2023.07.013},
author = {JaeYeon Park and JeongGil Ko},
}

@inproceedings{mcmahan2017communication,
  title={Communication-efficient learning of deep networks from decentralized data},
  author={McMahan, Brendan and Moore, Eider and Ramage, Daniel and Hampson, Seth and y Arcas, Blaise Aguera},
  booktitle={Artificial intelligence and statistics},
  pages={1273--1282},
  year={2017},
  organization={PMLR}
}

@article{park2023attfl,
  title={AttFL: A Personalized Federated Learning Framework for Time-series Mobile and Embedded Sensor Data Processing},
  author={Park, JaeYeon and Lee, Kichang and Lee, Sungmin and Zhang, Mi and Ko, JeongGil},
  journal={Proceedings of the ACM on Interactive, Mobile, Wearable and Ubiquitous Technologies},
  volume={7},
  number={3},
  pages={1--31},
  year={2023},
  publisher={ACM New York, NY, USA}
}

@article{zhao2018federated,
  title={Federated learning with non-iid data},
  author={Zhao, Yue and Li, Meng and Lai, Liangzhen and Suda, Naveen and Civin, Damon and Chandra, Vikas},
  journal={arXiv preprint arXiv:1806.00582},
  year={2018}
}

@article{hinton2015distilling,
  title={Distilling the knowledge in a neural network},
  author={Hinton, Geoffrey and Vinyals, Oriol and Dean, Jeff},
  journal={arXiv preprint arXiv:1503.02531},
  year={2015}
}

@inproceedings{touvron2021training,
  title={Training data-efficient image transformers \& distillation through attention},
  author={Touvron, Hugo and Cord, Matthieu and Douze, Matthijs and Massa, Francisco and Sablayrolles, Alexandre and J{\'e}gou, Herv{\'e}},
  booktitle={International conference on machine learning},
  pages={10347--10357},
  year={2021},
  organization={PMLR}
}

@inproceedings{guo2017calibration,
  title={On calibration of modern neural networks},
  author={Guo, Chuan and Pleiss, Geoff and Sun, Yu and Weinberger, Kilian Q},
  booktitle={International conference on machine learning},
  pages={1321--1330},
  year={2017},
  organization={PMLR}
}

@inproceedings{evci2022gradient,
  title={Gradient flow in sparse neural networks and how lottery tickets win},
  author={Evci, Utku and Ioannou, Yani and Keskin, Cem and Dauphin, Yann},
  booktitle={Proceedings of the AAAI conference on artificial intelligence},
  volume={36},
  number={6},
  pages={6577--6586},
  year={2022}
}

@article{kim2022curved,
  title={Curved Representation Space of Vision Transformers},
  author={Kim, Juyeop and Park, Junha and Kim, Songkuk and Lee, Jong-Seok},
  journal={arXiv preprint arXiv:2210.05742},
  year={2022}
}

@article{pearce2021understanding,
  title={Understanding softmax confidence and uncertainty},
  author={Pearce, Tim and Brintrup, Alexandra and Zhu, Jun},
  journal={arXiv preprint arXiv:2106.04972},
  year={2021}
}

@article{park2023self,
  title={Self-Attention LSTM-FCN model for arrhythmia classification and uncertainty assessment},
  author={Park, JaeYeon and Lee, Kichang and Park, Noseong and You, Seng Chan and Ko, JeongGil},
  journal={Artificial Intelligence in Medicine},
  volume={142},
  pages={102570},
  year={2023},
  publisher={Elsevier}
}

@article{jang2016categorical,
  title={Categorical reparameterization with gumbel-softmax},
  author={Jang, Eric and Gu, Shixiang and Poole, Ben},
  journal={arXiv preprint arXiv:1611.01144},
  year={2016}
}

@article{wang2020contextual,
  title={Contextual temperature for language modeling},
  author={Wang, Pei-Hsin and Hsieh, Sheng-Iou and Chang, Shih-Chieh and Chen, Yu-Ting and Pan, Jia-Yu and Wei, Wei and Juan, Da-Chang},
  journal={arXiv preprint arXiv:2012.13575},
  year={2020}
}

@article{shih2023long,
  title={Long Horizon Temperature Scaling},
  author={Shih, Andy and Sadigh, Dorsa and Ermon, Stefano},
  journal={arXiv preprint arXiv:2302.03686},
  year={2023}
}

@article{krizhevsky2009learning,
  title={Learning multiple layers of features from tiny images},
  author={Krizhevsky, Alex and Hinton, Geoffrey and others},
  year={2009},
  publisher={Toronto, ON, Canada}
}

@article{goodfellow2014explaining,
  title={Explaining and harnessing adversarial examples},
  author={Goodfellow, Ian J and Shlens, Jonathon and Szegedy, Christian},
  journal={arXiv preprint arXiv:1412.6572},
  year={2014}
}

@inproceedings{shin2022fedbalancer,
  title={FedBalancer: data and pace control for efficient federated learning on heterogeneous clients},
  author={Shin, Jaemin and Li, Yuanchun and Liu, Yunxin and Lee, Sung-Ju},
  booktitle={Proceedings of the 20th Annual International Conference on Mobile Systems, Applications and Services},
  pages={436--449},
  year={2022}
}

@inproceedings{zeng2021mercury,
  title={Mercury: Efficient on-device distributed dnn training via stochastic importance sampling},
  author={Zeng, Xiao and Yan, Ming and Zhang, Mi},
  booktitle={Proceedings of the 19th ACM Conference on Embedded Networked Sensor Systems},
  pages={29--41},
  year={2021}
}

@article{caldas2018leaf,
  title={Leaf: A benchmark for federated settings},
  author={Caldas, Sebastian and Duddu, Sai Meher Karthik and Wu, Peter and Li, Tian and Kone{\v{c}}n{\`y}, Jakub and McMahan, H Brendan and Smith, Virginia and Talwalkar, Ameet},
  journal={arXiv preprint arXiv:1812.01097},
  year={2018}
}

@inproceedings{cohen2017emnist,
  title={EMNIST: Extending MNIST to handwritten letters},
  author={Cohen, Gregory and Afshar, Saeed and Tapson, Jonathan and Van Schaik, Andre},
  booktitle={2017 international joint conference on neural networks (IJCNN)},
  pages={2921--2926},
  year={2017},
  organization={IEEE}
}

@inproceedings{karimireddy2020scaffold,
  title={Scaffold: Stochastic controlled averaging for federated learning},
  author={Karimireddy, Sai Praneeth and Kale, Satyen and Mohri, Mehryar and Reddi, Sashank and Stich, Sebastian and Suresh, Ananda Theertha},
  booktitle={International conference on machine learning},
  pages={5132--5143},
  year={2020},
  organization={PMLR}
}

@article{li2020federated,
  title={Federated optimization in heterogeneous networks},
  author={Li, Tian and Sahu, Anit Kumar and Zaheer, Manzil and Sanjabi, Maziar and Talwalkar, Ameet and Smith, Virginia},
  journal={Proceedings of Machine learning and systems},
  volume={2},
  pages={429--450},
  year={2020}
}

@inproceedings{he2016deep,
  title={Deep residual learning for image recognition},
  author={He, Kaiming and Zhang, Xiangyu and Ren, Shaoqing and Sun, Jian},
  booktitle={Proceedings of the IEEE conference on computer vision and pattern recognition},
  pages={770--778},
  year={2016}
}

@article{ruder2016overview,
  title={An overview of gradient descent optimization algorithms},
  author={Ruder, Sebastian},
  journal={arXiv preprint arXiv:1609.04747},
  year={2016}
}

@inproceedings{yu2021fed2,
  title={Fed2: Feature-aligned federated learning},
  author={Yu, Fuxun and Zhang, Weishan and Qin, Zhuwei and Xu, Zirui and Wang, Di and Liu, Chenchen and Tian, Zhi and Chen, Xiang},
  booktitle={Proceedings of the 27th ACM SIGKDD conference on knowledge discovery \& data mining},
  pages={2066--2074},
  year={2021}
}

@inproceedings{liu2020client,
  title={Client-edge-cloud hierarchical federated learning},
  author={Liu, Lumin and Zhang, Jun and Song, SH and Letaief, Khaled B},
  booktitle={ICC 2020-2020 IEEE International Conference on Communications (ICC)},
  pages={1--6},
  year={2020},
  organization={IEEE}
}

@inproceedings{kornblith2019similarity,
  title={Similarity of neural network representations revisited},
  author={Kornblith, Simon and Norouzi, Mohammad and Lee, Honglak and Hinton, Geoffrey},
  booktitle={International conference on machine learning},
  pages={3519--3529},
  year={2019},
  organization={PMLR}
}

@article{luo2021no,
  title={No fear of heterogeneity: Classifier calibration for federated learning with non-iid data},
  author={Luo, Mi and Chen, Fei and Hu, Dapeng and Zhang, Yifan and Liang, Jian and Feng, Jiashi},
  journal={Advances in Neural Information Processing Systems},
  volume={34},
  pages={5972--5984},
  year={2021}
}

@article{collins2022fedavg,
  title={Fedavg with fine tuning: Local updates lead to representation learning},
  author={Collins, Liam and Hassani, Hamed and Mokhtari, Aryan and Shakkottai, Sanjay},
  journal={Advances in Neural Information Processing Systems},
  volume={35},
  pages={10572--10586},
  year={2022}
}

@incollection{NEURIPS2019_9015,
title = {PyTorch: An Imperative Style, High-Performance Deep Learning Library},
author = {Paszke, Adam and Gross, Sam and Massa, Francisco and Lerer, Adam and Bradbury, James and Chanan, Gregory and Killeen, Trevor and Lin, Zeming and Gimelshein, Natalia and Antiga, Luca and Desmaison, Alban and Kopf, Andreas and Yang, Edward and DeVito, Zachary and Raison, Martin and Tejani, Alykhan and Chilamkurthy, Sasank and Steiner, Benoit and Fang, Lu and Bai, Junjie and Chintala, Soumith},
booktitle = {Advances in Neural Information Processing Systems 32},
pages = {8024--8035},
year = {2019},
publisher = {Curran Associates, Inc.},
}

@article{hsu2019measuring,
  title={Measuring the effects of non-identical data distribution for federated visual classification},
  author={Hsu, Tzu-Ming Harry and Qi, Hang and Brown, Matthew},
  journal={arXiv preprint arXiv:1909.06335},
  year={2019}
}

@article{
agarwala2020temperature,
title={Temperature check: theory and practice for training models with softmax-cross-entropy losses},
author={Atish Agarwala and Samuel Stern Schoenholz and Jeffrey Pennington and Yann Dauphin},
journal={Transactions on Machine Learning Research},
issn={2835-8856},
year={2023},
note={}
}

@inproceedings{
Li2020On,
title={On the Convergence of FedAvg on Non-IID Data},
author={Xiang Li and Kaixuan Huang and Wenhao Yang and Shusen Wang and Zhihua Zhang},
booktitle={International Conference on Learning Representations},
year={2020},
}

@article{shin2024effective,
  title={Effective Heterogeneous Federated Learning via Efficient Hypernetwork-based Weight Generation},
  author={Shin, Yujin and Lee, Kichang and Lee, Sungmin and Choi, You Rim and Kim, Hyung-Sin and Ko, JeongGil},
  journal={arXiv preprint arXiv:2407.03086},
  year={2024}
}

@misc{ucihar,
  author       = {Reyes-Ortiz Jorge and Anguita Davide and Ghio Alessandro and Oneto Luca and Parra Xavier},
  title        = {{Human Activity Recognition Using Smartphones}},
  year         = {2013},
  howpublished = {UCI Machine Learning Repository},
  note         = {{DOI}: https://doi.org/10.24432/C54S4K}
}

@inproceedings{li2021fedrs,
  title={Fedrs: Federated learning with restricted softmax for label distribution non-iid data},
  author={Li, Xin-Chun and Zhan, De-Chuan},
  booktitle={Proceedings of the 27th ACM SIGKDD conference on knowledge discovery \& data mining},
  pages={995--1005},
  year={2021}
}

@article{li2021fedbn,
  title={Fedbn: Federated learning on non-iid features via local batch normalization},
  author={Li, Xiaoxiao and Jiang, Meirui and Zhang, Xiaofei and Kamp, Michael and Dou, Qi},
  journal={arXiv preprint arXiv:2102.07623},
  year={2021}
}

@article{xing2022efficient,
  title={An efficient federated distillation learning system for multitask time series classification},
  author={Xing, Huanlai and Xiao, Zhiwen and Qu, Rong and Zhu, Zonghai and Zhao, Bowen},
  journal={IEEE Transactions on Instrumentation and Measurement},
  volume={71},
  pages={1--12},
  year={2022},
  publisher={IEEE}
}

@article{xiao2025federated,
  title={Federated contrastive learning with feature-based distillation for human activity recognition},
  author={Xiao, Zhiwen and Tong, Huagang},
  journal={IEEE Transactions on Computational Social Systems},
  year={2025},
  publisher={IEEE}
}

@article{seo2025fedunet,
  title={FedUNet: A Lightweight Additive U-Net Module for Federated Learning with Heterogeneous Models},
  author={Seo, Beomseok and Lee, Kichang and Park, JaeYeon},
  journal={arXiv preprint arXiv:2508.12740},
  year={2025}
}

@article{lee2025gmt,
  title={GMT: Gzip-based Memory-efficient Time-series classification},
  author={Lee, Sungmin and Lee, Kichang and Park, Jaeyeon and Ko, Jeonggil},
  journal={ICT Express},
  volume={11},
  number={2},
  pages={270--274},
  year={2025},
  publisher={Elsevier}
}

@article{dosovitskiy2020image,
  title={An image is worth 16x16 words: Transformers for image recognition at scale},
  author={Dosovitskiy, Alexey},
  journal={arXiv preprint arXiv:2010.11929},
  year={2020}
}
